\documentclass[times,final]{elsarticle}
\usepackage{jcomp}
\usepackage{geometry}
\usepackage{framed,multirow}
\usepackage{amssymb}
\usepackage{latexsym}
\usepackage[utf8]{inputenc} 
\usepackage[ruled]{algorithm2e}
\usepackage[T1]{fontenc}    
\usepackage{hyperref}       
\usepackage{url}            
\usepackage{float}
\usepackage{ulem}
\usepackage{booktabs}       
\usepackage{amsfonts}       
\usepackage{nicefrac}       
\usepackage{microtype}      
\usepackage[T1]{fontenc}    
\usepackage{hyperref}       
\usepackage{url}            
\usepackage{color}   
\usepackage{times}          
\usepackage{xspace}
\usepackage{amsmath, amssymb, amsthm, enumerate, bm, capt-of,mathtools}
\usepackage{amssymb}
\usepackage{wrapfig}
\usepackage{natbib}
\usepackage{graphicx}
\usepackage{booktabs} 

\usepackage{natbib}
\usepackage{cite}	

\usepackage{amssymb,amsthm}
\usepackage{amsmath}
\usepackage{amsfonts}
\usepackage{breqn} 

\usepackage{xcolor}
\usepackage{color}

\newcommand\Alpha[0]{\bm{\alpha}}
\usepackage[small, compact]{titlesec}
\usepackage{enumerate}

\usepackage{epstopdf}
\usepackage{graphicx,epsfig}
\usepackage{caption}
\usepackage{subcaption}
\usepackage{grffile}	

\newcommand{\angstrom}{\mbox{\normalfont\AA}}

\newcommand{\ours}{{DMF}\xspace}

\newcommand{\mf}{multi-fidelity }
\newcommand{\hf}{high-fidelity }
\newcommand{\lf}{low-fidelity }

\newcommand{\h}{{\bf h}}

\renewcommand{\o}{{\bf o}}

\renewcommand{\r}{{\bf r}}

\newcommand{\w}{{\bf w}}
\newcommand{\x}{{\bf x}}
\newcommand{\y}{{\bf y}}
\newcommand{\z}{{\bf z}}

\newcommand{\E}{{\bf E}}

\newcommand{\ben}{\begin{enumerate}}
\newcommand{\een}{\end{enumerate}}

\newcommand{\norm}[1]{\lVert#1\rVert}

\newcommand{\ie}{{i.e.,}\xspace}
\newcommand{\eg}{{e.g.,}\xspace}

\newcommand{\cmt}[1]{}

\date{}
\journal{Journal of Computational Physics}
\begin{document}

\begin{frontmatter}

\title{Differentiable Multi-Fidelity Fusion: Efficient Learning of Physics Simulations with Neural Architecture Search and Transfer Learning}

\author[1]{Yuwen {Deng}}
\author[1]{Wang {Kang}}
\author[1]{Wei W. {Xing} \corref{cor1}}
\cortext[cor1]{Corresponding author; 
Email: wxing@buaa.edu.cn.}
\address[1]{School of Integrated Circuit Science and Engineering, Beihang University, 37 Xueyuan Road, Haidian District, Beijing, 100191, China
}


\begin{abstract}
With rapid progress in deep learning,
neural networks have been widely used in scientific research and engineering applications as surrogate models.
Despite the great success of neural networks in fitting complex systems, two major challenges still remain: i) the lack of generalization on different problems/datasets, and ii) the demand for large amounts of simulation data that are computationally expensive. 
To resolve these challenges, we propose the differentiable \mf (DMF) model, which leverages neural architecture search (NAS) to automatically search the suitable model architecture for different problems, and transfer learning to transfer the learned knowledge from low-fidelity (fast but inaccurate) data to high-fidelity (slow but accurate) model.
Novel and latest machine learning techniques such as hyperparameters search and alternate learning are used to improve the efficiency and robustness of DMF.
As a result, DMF can efficiently learn the physics simulations with only a few high-fidelity training samples, and outperform the state-of-the-art methods with a significant margin (with up to 58$\%$ improvement in RMSE) based on a variety of synthetic and practical benchmark problems.


%
%
%
\end{abstract}
\end{frontmatter}

\section{Introduction}\label{sec:Introduction}
Many physics systems in science and engineering require resource-consuming physics simulation.
Generally, these simulations require to be repeated large amounts of times to deliver a satisfactory analysis, \eg uncertainty quantification, optimization, sensitivity analysis, and inverse parameter estimation. 
Therefore, for these systems, a data-driven surrogate model obtaining predictions is usually adopted to provide a quick estimation and design guidelines for the simulations.
The surrogate model is usually a regression model approximating the relation between the system inputs and outputs, which may come from either the experiment results or a numerical model based on the simulations, also known as training data in this instance.

 In general, there are various mathematical models to describe a given physical system, these models provide different fidelity data according to different levels of physical detail incorporated into the equations, \eg geometry and initial-boundary conditions.
 The traditional surrogate model is usually built based on high-fidelity simulation results, which are accurate but expensive to acquire.
In practice, we may efficiently reduce the computational cost by using low-fidelity simulations. Roughly speaking, low-fidelity computational models are those of low resolution or complexity in terms of the physical laws or adjustable settings of the computer-based approximation. 
In many cases, the less accurate \lf data are cheap and abundant while they 
can supply useful information on the trends for \hf data, hence we can trade the accuracy for the computational cost by using low-fidelity models and make predictions of scarce \hf data based on \lf data.

Multi-fidelity fusion has shown to be an efficient and promising approach to achieving high accuracy in many systems by utilizing both low- and high-fidelity data.
 Auto-regressive based Gaussian process (GP) is an important framework using Bayesian techniques for \mf modeling, which can yield predictions of the mean values and uncertainties. 
 The seminal auto-regressive (AR) fusion model of Kennedy and O'Hagan~\citep{kennedy2000predicting} assumes the linear relationship between different levels of fidelity and introduces a linear transformation for the univariate outputs. 
 To overcome the limitation of the linearity in AR, Perdikaris et al.~\citep{perdikaris2017nonlinear} propose non-linear AR (NAR). They ignore the output distributions and directly use the low-fidelity solution as an input for the high-fidelity GP model, which is essentially a concatenating GP structure, also known as \textit{Deep GP}~\citep{damianou2015deep}.
Other works~\citep{cutajar2019deep,poloczek2017multi,kandasamy2016gaussian,zhang2017information,song2019general} develop more GP-based \mf models to improve the performance. 
Unfortunately, the main drawback of the GP-based method is the significant computational complexity for optimizing the \mf model, which demands large-matrix inversion that scales at $O(N^3)$ ($N$ is the number of training samples). As a result, the GP-based method is limited in multiple output scenarios.
Also, the GP-based method cannot easily extend to high-dimensional simulations, making it difficult to apply to many real-world problems.
Therefore, the \mf approach that can overcome these limitations is urgently needed.

With the rapid development of deep learning, neural networks (NN) have successfully been applied in many areas, including computational science and engineering.
With the incorporation of the neural layers and nonlinear activation function, NN has demonstrated that the composite function model possesses a universal approximation property~\citep{hornik1989multilayer,opschoor2020deep} and can be used to approximate any function.
In the meanwhile, there have been numerous open-source frameworks for building and training 
deep neural networks in recent years, such as Tensorflow~\citep{abadi2016tensorflow}, Theano~\citep{bergstra2010theano}, Pytorch~\citep{paszke2017automatic}. These frameworks supply with clients compact and organized methodology to effectively speed up work processes that build expressive surrogate models. 
In addition, these software packages have a large user base outside of applications in science and engineering. Using specialized hardware architectures that speed up model training is also an advantage of using neural networks to create surrogate models, such as GPUs and TPUs. More
importantly, with the significant investments in constructing specifically tailored high-performance computing resources, neural networks driven by artificial intelligence (AI) applications are becoming more popular. As a result, hardware accelerators' capabilities will expand rapidly.


NN has a long history of being used in surrogate models almost as long as the history of GP surrogates.
We briefly discuss some of the latest advances first.
Kim et al. introduce a knowledge-based neural network (KBNN)~\citep{kim2007hybrid} that utilizes a prior-knowledge response surface model (RSM) as the low-fidelity model.  
Meng, Raissi, Pun et al. introduce physics informed neural network (PINN)~\citep{raissi2019physics,pun2019physically} to build the surrogate model, which combines a physics regularization layer with the normal NN, which is based on the corresponding PDEs. 
Then, they revise the training loss of NN and penalized the solutions that do not satisfy the equations.  However, KBNN and PINN both need prior knowledge of general physical laws, for complex physical simulation, it's infeasible to describe in analytical equations. Moreover, they also need
to design the network manually according to
each specific PDE, which also leads to inconveniences in modeling. 
To make the uncertainty qualification, 
Meng replaces the normal NN with Bayesian neural network (BNN) in ~\citep{meng2021multi} to yield predictions of the mean values as well as estimates of uncertainty; Li et al.~\citep{li2020deep} propose a BNN approach to multi-fidelity fusion and use active learning technique to select the training samples. 
However, BNN is extremely computationally expensive and performs poorly on high dimensional data; moreover, it is hard to interpret and requires Copula functions to separate out effects between different parts of the network. These methods employing neural networks primarily concentrate on designing networks to improve clarity involving the application of physical constraints. However, the main drawback of using the neural network surrogate model is the limitation of the number of training samples or \hf data. The incorporation of data from a variety of lower fidelity models or experiments into the neural network training procedure may offer a potential solution to the problem of a lack of data volume.

To harness the advantage of multi-fidelity data, many multi-fidelity NN methods have been proposed recently.
Many of them root in the idea of transfer learning~\citep{pan2009survey}, which aims to use the knowledge learned from one task (in this case, the low-fidelity mapping) to solve another similar task (in this case, the high-fidelity mapping) where the labeled data are limited or missing.
In the areas of computer vision and natural language processing, transfer learning has been widely used to improve the performance of learning by avoiding many expensive computations~\citep{jiang2017face, zhu2011heterogeneous}.
%
%
It has been demonstrated that transfer learning has the following advantages over traditional learning~\citep{pan2009survey}: (i) faster convergence of the target network, 
 (ii) smaller initial training error, and
(iii) smaller validation error on smaller datasets. 

In the \mf problem, we are only interested the \hf data. However, the \hf simulation tends to require many computational resources
and provides higher levels of accuracy. Thus, the size of \hf data is extremely limited.
At the same time, the \lf data and \hf data are also highly correlated. 
In this case, knowledge transfer, if done successfully, will significantly improve the performance of learning by avoiding many expensive data-simulating efforts. 
 Souvik Chakraborty et al.~\citep{chakraborty2021transfer} train a physics-informed deep neural network (DNN) as the \lf model first and then transfer the parameters to a \hf model, however, in the latter phase, they only transfer the parameters of DNN consisting of a fixed number of layers and nodes, which leads to lack of flexibility.
 Guo et al.~\citep{guo2022multi} propose a multilevel NN model, which is done sequentially by firstly modeling \lf model and then modeling the correlation between two fidelity levels with a single hidden layer, but they fix the \lf model parameters while training on \hf samples instead of fine-tuning, which may lead to an error while there is no strong linear correlation between two levels.
 Song et al.~\citep{song2021transfer} acquire transfer learning with a convolutional neural network-based encoder and decoder architecture~\citep{mo2019deep}, however, the convolutional layer tends to extract the features of 2D data and is suboptimal on other physical system datasets.

Despite the above efforts, the transfer learning-based NN methods still have some limitations.
First, the transfer learning-based NN methods implicitly assume that the \lf model and \hf model share the same architecture, which imposes restrictions on the \hf model and may lead to suboptimal performance.
Second, the transfer learning-based NN methods are not flexible enough to adapt to different \mf structures and different \hf models.
%
Most importantly, the NN structure requires a careful design for each specific problem and relies on domain knowledge, which is not always available.
To address the above issues and further improve the performance of NN-based multi-fidelity fusion, we first propose a novel method, {differentiable neural network}, based on binary optimization that can learn the optimal NN architecture automatically for any given problem.
To enable multi-fidelity fusion, we then use fine-tuning methods to transfer the learned parameters on low-fidelity data to the high-fidelity model, which can significantly reduce the computational cost and improve the performance of NN-based multi-fidelity fusion by putting the \lf data and \hf data on the same footing.
To further improve the efficiency and robustness of the proposed method, we also propose a novel hyperparameters search method to select the best hyperparameters settings efficiently and an alternate learning method to replace the joint optimization without the introduction of extra error.
%
The proposed method is assessed on various datasets including synthetic data and real applications.
It is shown that the proposed method can significantly improve the performance of NN-based multi-fidelity fusion and is more robust than the SOTA multi-fidelity fusion methods.
The main contributions of this paper are summarized as follows:
%
%
\begin{enumerate}
\item To be able to build surrogates for a wide spectrum of physics simulations without the need for domain knowledge and expensive computational cost, 
we propose a novel differentiable neural network (DNN) based on binary optimization that can learn the optimal NN architecture automatically form cheap-to-gain \lf data and transfer such knowledge to deliver a high-fidelity model.
\item we introduce an alternate optimization scheme to replace the traditional joint optimization in our automatic machine learning pipeline to improve the efficiency and robustness of the proposed method without introducing extra error.
\item To reduce the influence of the selection of hyperparameters, we introduce search space pruning methods to select the best hyperparameters settings efficiently.
%
%
\item The proposed method achieves remarkable performance improvement on various datasets, including standard benchmarks, high-dimensional benchmarks, and real applications, and outperform most state-of-the-art multi-fidelity fusion methods consistently.
    
\end{enumerate}

\section{Statement of The Problem}\label{sec:Statement}
Without loss of generality, we denote the solution of PDE by scalar variable $u(\z,t,\x)$, where $\z$ and $t$ are the spatial and temporal coordinates, respectively, and $\x$ the parameters corresponding to the system of equations. Consider a general nonlinear system with steady-state PDEs,
\begin{equation}
    \left\{
    \begin{aligned}
        &\frac{\partial}{\partial t}u(\z,t,\x)+\mathcal{F}(u(\z,t,\x))=\mathcal{S}(\z,t,\x),&\Omega\times (0,T]\times\mathcal{X}\\
        &\mathcal{B}(u,\x)=0,&\Omega\times (0,T]\times\mathcal{X}\\
        &u(\z,0,\x)=u_0(\z,\x),&\Omega\times \{t=0\}\times\mathcal{X}
    \end{aligned}
    \right.
\end{equation}
where $\mathcal{S}$ is the source function, 
$\mathcal{F}$ is a nonlinear function that may contain parameters indicated by $\x \in \mathbb{R}^l$, $\mathcal{B}$ is the boundary condition, and $u_0(\z,\x)$ denotes the initial condition; $\Omega$ and $T$ are the spatial and temporal domains of interest.
As a result, $\z \in\Omega\subset \mathbb{R}^p $ and $t \in[0, T] $ parameterizes the temporal domain, the system is solved for $u(\z,t,\x) \in \Omega\times (0, T]\times\mathcal{X}$ using a numerical solver. 
As referenced in the introduction,
the immediate execution of the mathematical solver can be computationally restrictive. Experts propose a data-driven surrogate model to estimate $u(\z,t,\x)$ effectively given the boundary condition in the scope of interest for the spatial-temporal domain of $\Omega\times (0, T]\times\mathcal{X}$.

For the surrogate model, we will generate $N_\z \times N_t \times N_\x$ training samples on the domain of interest, where $N_\z$, $N_t$, and $N_\x$ are the quantities of spatial, temporal, and parameter samples, respectively.
To cover the response surface for such a high-dimensional input space problem, we discretize the continuous inputs space on specified spatial locations $\z_1,...,\z_{N_\z}$ and temporal locations $t_1,...,t_{N_t}$ to provide
an approximation of $u(\z,t, \x)$. 
With the given coordinates, our quantity of interest (QoI) becomes a vectorial function of the PDE parameters.

\begin{equation}
\y(\x)=\left(u(\z_1,t_1,\x),\cdots,u(\z_{N_\z},t_1,\x),u(\z_{1},t_2,\x),\cdots,u(\z_{N_\z},t_{N_t },\x)\right)\in \mathbb{R}^d,
\end{equation}
where $\z_i, t_i $ are the spatial and temporal coordinates of a regular/irregular grid, and $d = N_\z \times N_t$ is the total number of spatial-temporal grid
points. Because the number of coordinates $N_\z$ and $N_t$ needs to be 
sufficiently large to provide an accurate/\hf
solution, the dimension of vector $\y(\x)$ also becomes extremely large. 
Notice that the transformation from learning $u(\z,t,\x)$ to learning $\y(\x)$ is equivalent.

To acquire a high-fidelity solution for $\y(\x)$, we generally need to use a high-order basis expansion, high-order stencil, or tight iteration bounds, which are costly to process. On the contrary, a lower-order basis expansion, a lower-order stencil, or a looser iteration bound.
Moreover, we can use a simpler physical model, e.g., linearizing, spatial averaging, considering a 2-d slice, or others. The computational model with options for adjusting settings to generate various fidelity solutions and variable input parameters is required. 

Without loss of generality, We denote the \mf {training set} by $D_m=\{(\x_m^i,\y_m^i)\}^{N_m}$, $m=1,2,\cdots, M$. $\x\in \mathcal{X}$ represents an l-dimensional input, $\y_m\in \mathbb{R}^{d_m}$ represents the corresponding response, where $d_m$ denotes the dimension of $\y_m$, $N_m$ represents the number of samples with fidelity $m$, and $M$ is the total number of {fidelity}. 
In general, $\{d_m\}$ are not necessarily identical, and we assume that $d_1\leqslant d_2\leqslant \cdots\leqslant d_M$, and $N_1\geqslant N_2\geqslant \cdots \geqslant N_M$ to match the practical situation that high-fidelity simulations show more details but are expensive to obtain. 
Also notice that in many multi-fidelity fusion methods, the high-fidelity inputs are required to form a subset of the low-fidelity input, \ie $\{\x_m^i\}_{i=1}^{N_m} \subset \{\x_{m'}^i\}_{i=1}^{N_{m'}}$ for all $ m > m'$. This requirement significantly hinders the application of multi-fidelity fusion methods in many real-world problems. In this work, we do not impose such a restriction, and thus the proposed method is more general for broader applications.
Our goal is to estimate mapping $f_M(\x): \x \rightarrow \y_M$ as accurately as possible using dataset $\{D_m\}_{m=1}^{M}$.

\section{Differentiable Multi-fidelity Fusion}\label{dmf}

\subsection{Adaptive Surrogate Model Using Differentiable Architecture Search} 

Recent studies in surrogate model~\citep{meng2020composite,meng2021multi,guo2022multi,islam2021extraction} make much effort to design a NN model to improve its performance.
They introduce the composition of linear NN and non-linear NN to capture the complex relationships between inputs and outputs. 
However, the design of the NN and the transformation normally rely on field knowledge, which is not always available in practice and may not be general enough to be applied to different problems.


In our model, to better utilize the information of low-fidelity data, we introduce a differentiable architecture search (DARTS)~\citep{liu2018darts} to automatically search for the best NN structure as the foundation model.
More specifically, the mapping from $\x$ to $\y_L$ is considered as a composition of multiple NN operations $ o$, and the weights of each operation are optimized by DARTS.
In this work, our operation candidates set $ \mathcal{O}$ consists of five different operations, namely, shallow operation, wide operation, linear operation, and zero operation.
The deep operation has four layers with 20 neurons per layer; the shallow has two layers each with 20 neurons; the wide operation has two layers with 40 neurons per layer. 
We use the rectified linear unit (ReLU) as the activation function for the above operations.
The linear operation has one layer with 20 neurons and does not have any activation function; the zero operation means no layer and only outputs zero regardless of the inputs.
Each operation is associated with a weight coefficient $\alpha_l$.
The composition of multiple operations is defined as:
%
%
%
\begin{equation}\label{eq2}
     o=\sum_{l=1}^{|\mathcal{O}|}\frac{\exp{(\alpha_l)}}{\sum_{l=1}^{|\mathcal{O}|}\exp{(\alpha_l)}}o_l,
\end{equation}
where {$|\mathcal{O}|$} is the number of operation in $\mathcal{O}$, e.g., five different NN {operations} in our model; $o_l$ is from the operation candidate set $\mathcal{O}$. Thus the operation $o$ is a mixing operation parameterized by a vector $\Alpha$ of dimension $|\mathcal{O}|$.
Different from the existing models, $\Alpha$ will be decided automatically in the training process to automatically form the best NN structure.

\begin{figure}[h]
    \centering
    \includegraphics[width=14cm]{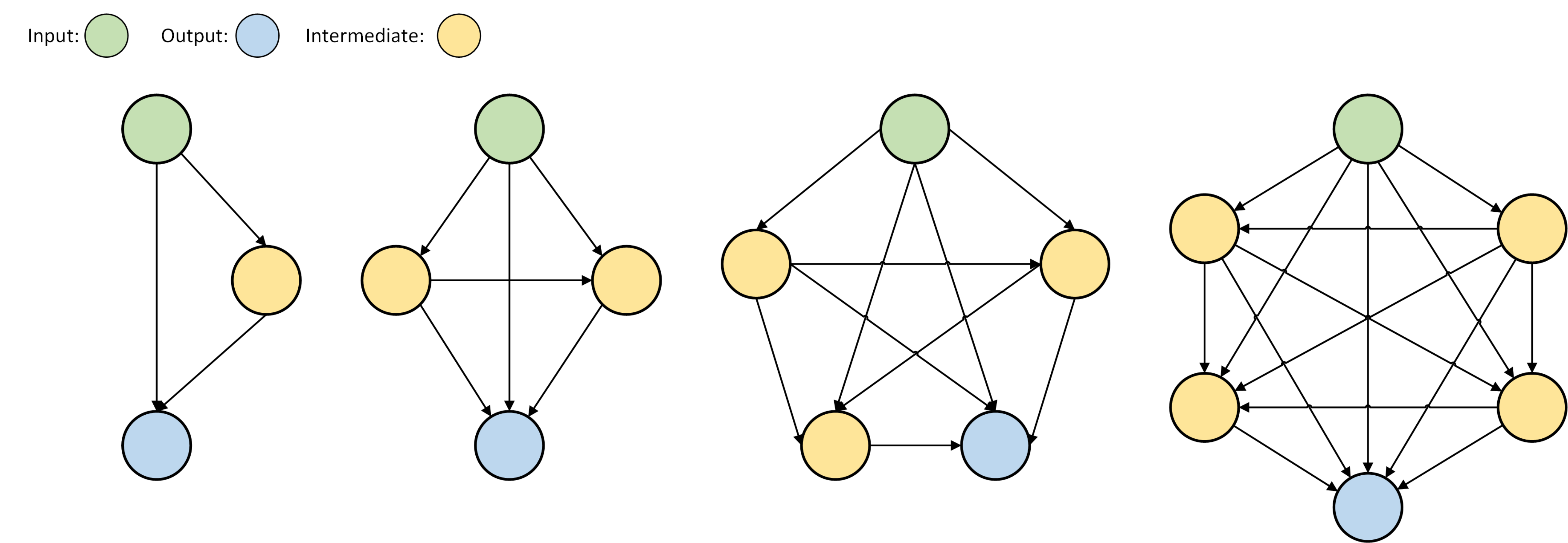}
    \caption{Scheme of DMF with three cells, four cells, five cells, and six cells from left to right. Each arrow represents a composition operation.}
    \label{fig0}   
\end{figure}

We can now directly learn the mapping from $\x$ to $\y_L$ by optimizing the network parameters $\w$ and the operation weights $\Alpha$ jointly to get the optimal NN structure.
However, with only five different operations, the resulting model is simply a sum of several networks, which is not powerful enough to capture the complex relationship between $\x$ and $\y_L$ unless we dramatically increase the number of operations, which requires careful hand-crafting and is not scalable. To resolve this issue, we expand the search space by introducing the intermediate nodes between $\x$ and $\y_L$.
More specifically, the final model is represented as a Directed Acyclic Graph (DAG) denoted by $G$. 
It consists of three types of nodes: input node, output node, and intermediate node.
We show some possible DAGs in Fig.~\ref{fig0} with different numbers of intermediate nodes for illustration.

For the sake of clarity, we consider only one intermediate node and denote all nodes as $\{\h_{0},\h_{1},\h_{2} \}$, respectively. Here, $\h_{0}=\x$, $\h_{2}=\y_L$ are the input and low-fidelity nodes.
$o^{(i,j)}$ represents the composite operation from $i$ to $j$, and the zero operation represents no connection between $i$ and $j$. Therefore, each node $j$ can be written as:
\begin{equation}
    \h_{j}=\sum_{i<j}o^{(i,j)}(\h_{i}).
\end{equation}
%
%
%

With the model being fully defined with $G$ and the composite operation, we can now train the model to learn the mapping from $\x$ to $\y_L$.
Our objective is to optimize a binary optimization problem with the parameters $\w$ in layers and the weights $\Alpha$ of each operation given $G$ and the low-fidelity data.
Denote the cost function by the $\mathcal{L}_{m}(\w,\Alpha)$ for the fidelity $m$ , which is written as:
\begin{equation}
    \mathcal{L}_{m}(\w,\Alpha)=\frac{1}{N_m}\sum_{i=1}^{N_m}\norm{\y^i_m-f(\x^i,\w),\Alpha)}_2^2+\lambda_1\norm{\w}_2^2+\lambda_2\norm{\Alpha}_2^2,
    \label{eq10}
\end{equation}
where the first term is the mean square error (MSE) for the observation data, and the rest are regularization terms with $\lambda_i (i=1,2)\geqslant0$ being the penalty coefficients or weight decays,
Optimizing Equation \eqref{eq10} directly may be of low efficiency because their high correlation during the optimization can cause a high variance of the gradient.
Thus, we propose an alternate optimization:
%
\begin{equation}\label{eq3}
\begin{aligned}
       \w^{t+1} &=  \arg\min\limits_{\w}\mathcal{L}_m(\w | \Alpha^{t}) \\ 
       \Alpha^{t+1} &=  \arg\min\limits_{\Alpha}\mathcal{L}_m(\Alpha | \w^{t+1} ),
\end{aligned}
\end{equation}
where the superscript indicates the iteration.
There are two immediate advantages to such an optimization approach:
i) simplification of the code implementation and reduction of the memory cost during the computation of the backpropagation of the gradient 
and ii) potential improvement of the convergence by reducing the variance of the computed gradient during the optimization by limiting the number of variables involved in one optimization step. 
In practice, updating of $\Alpha$ will introduce a significant perturbation to the gradient of $\w$. Thus, instead of updating $\w$ and $\Alpha$ jointly totally, $\w$ was updated more frequently than $\Alpha$. The ratio is controlled by the parameter $k$.
A schematic illustration of the training of DMF is shown in Algorithm~\ref{algo0}.
\begin{algorithm}[]
\caption{Alternate Training for DMF}
\KwIn{Initial dataset $\mathcal{D}$, iterations $\mathcal{N}$, $\w$ training epochs $\mathcal{W}$, step $k$, $\w$ learning rate  $r_1$, $\Alpha$ learning rate $r_2$}
\For{\ $t$ \ from \ 1 \ to \ $\mathcal{N}$ \ }{
\For{$i$ from 1 to $\mathcal{W}$}{
$\qquad \w_t^{(i+1)}=\w_t^{(t)}-r_1 \nabla_\w \mathcal{L}_m(\w,\Alpha) \;$

\If{$i\mod k=0$}{
$\qquad \Alpha_t^{(i+1)}=\Alpha_t^{(i)}-r_2 \nabla_\Alpha \mathcal{L}_m(\w,\Alpha)\;$
}
}
}
\label{algo0}
\end{algorithm}
Furthermore, we show that such an alternate optimization will not introduce extra error to the convergence.

\newtheorem{proposition}{Proposition}
\begin{proposition}
    If the product $r_{\w} \times r_{\Alpha}$ is small enough, $r$ represents the corresponding learning rate, optimizing $\Alpha$ and $\w$ alternately is equivalent to optimizing then simultaneously, \ie the optimization will lead to the same asymptotic convergence and convergence rate. 
\end{proposition}
The proof of {Proposition 1} is preserved in Appendix ~\ref{ap} for clarity.

%
 



\subsection{Transfer Learning For Multi-Fidelity Fusion}\label{trans}
We first use the method discussed in the previous section to train the \lf model $f_L(\x)$.
To explore the cross-correlation between $\y_L$ and $\y_H$,  we consider two different models in this section.

%

\begin{figure}[H]
    \centering
    \includegraphics[width=12cm]{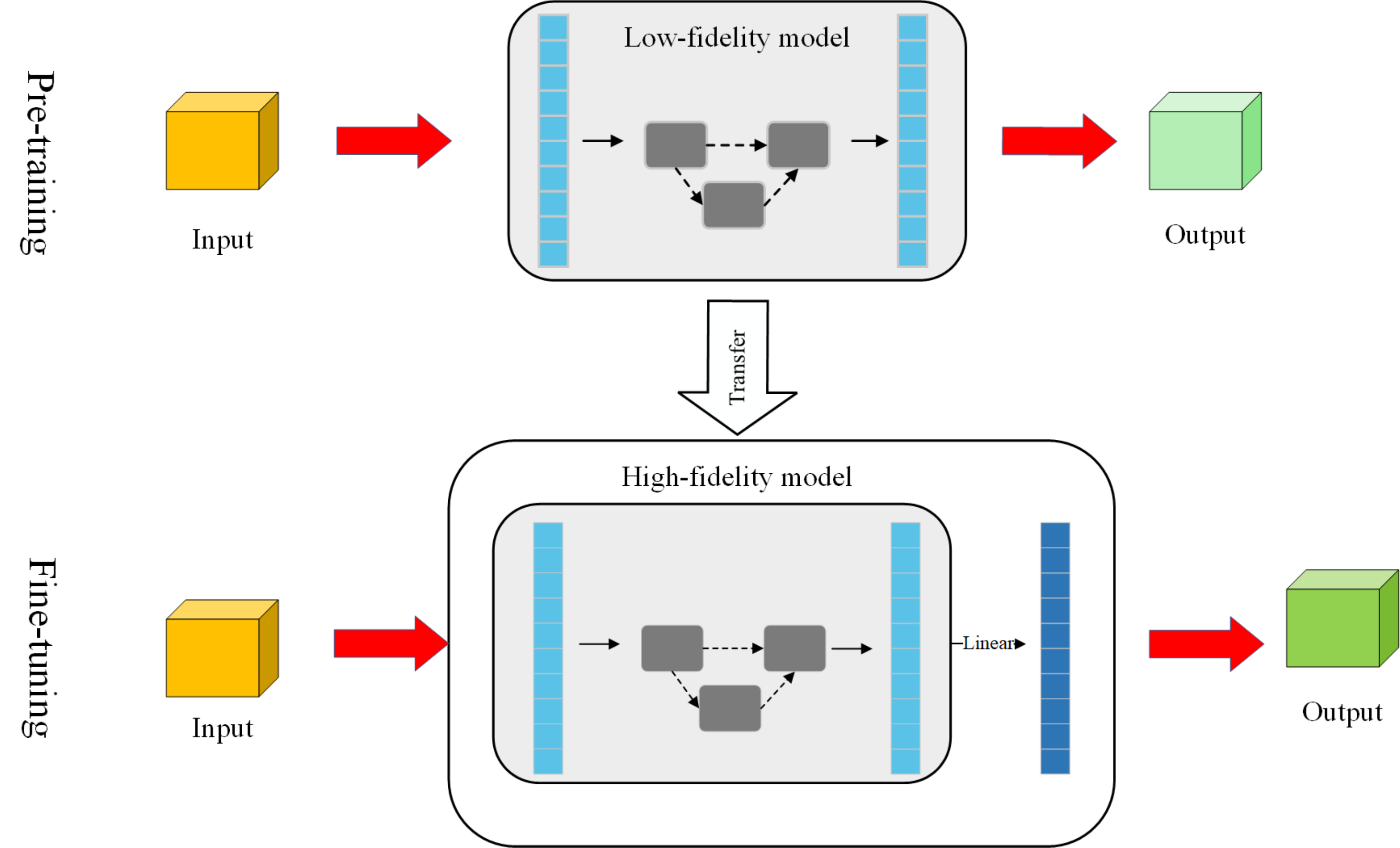}
    \caption{Overall procedures for DMF. The \lf model ${\rm DMF}_L$ is taken as pre-training, and the \hf model DMF-trans consists of two components, ${\rm DMF}_{L}$ and $\mathcal{NN}_l$.}
    \label{fig2}
\end{figure}

The first one uses the \lf model to predict the corresponding low-fidelity response $\y_L$, 
then the mapping between { $\y_L$} and $\y_H$ is again learned using the automatic learning method discussed in the previous section for the \lf model.
We denote this model by the DMF-2 network. The second approach builds a high-fidelity model based on the \lf model with transfer learning.
The underlying assumption is that the \lf data and the \hf data are highly correlated, and the \lf model already contains information to make accurate predictions for the \hf data given any input $\x$.
To allow some more flexibility in the \hf model, we add a single linear layer $\mathcal{NN}_l$ to the \lf model, which is trained to predict $\y_H$ based on the input data $(\mathcal{X}_{H},f_L(\mathcal{X}_{H}))=\{(\x_H^{i},f_L(\x_H^{i})) \}^{N_H}$, and the available high-fidelity output data $\y_H=\{y_H^{i}\}^{N_H}$.
We call this model DMF-trans and the overview of our model is shown in Fig.~\ref{fig2}.


We denote the pre-training \lf model by ${\rm DMF}_{L}$, which is concatenated with a simple network $\mathcal{NN}_l$ approximating the high-fidelity function.
%
The parameter of DMF-trans includes ${\rm DMF}_L$'s parameters $\w_L$ and $\Alpha_L$, and $\mathcal{NN}_l$'s parameters $\w_{\mathcal{NN}_l}$.
%
After the model $ {\rm DMF}_L$ is finished, we train DMF-trans on the \hf data. Notably, we use different learning rates for different parameters. In general, we train $\w_L, \Alpha_L$ with a tiny learning rate while training $\w_{\mathcal{NN}_l}$ with a relatively large learning rate. We will analyze the effect of different learning rates in Section~\ref{SH}. 


\begin{figure}[htbp]
    \begin{minipage}[t]{0.33\linewidth}
    \centering
    \includegraphics[width=5cm]{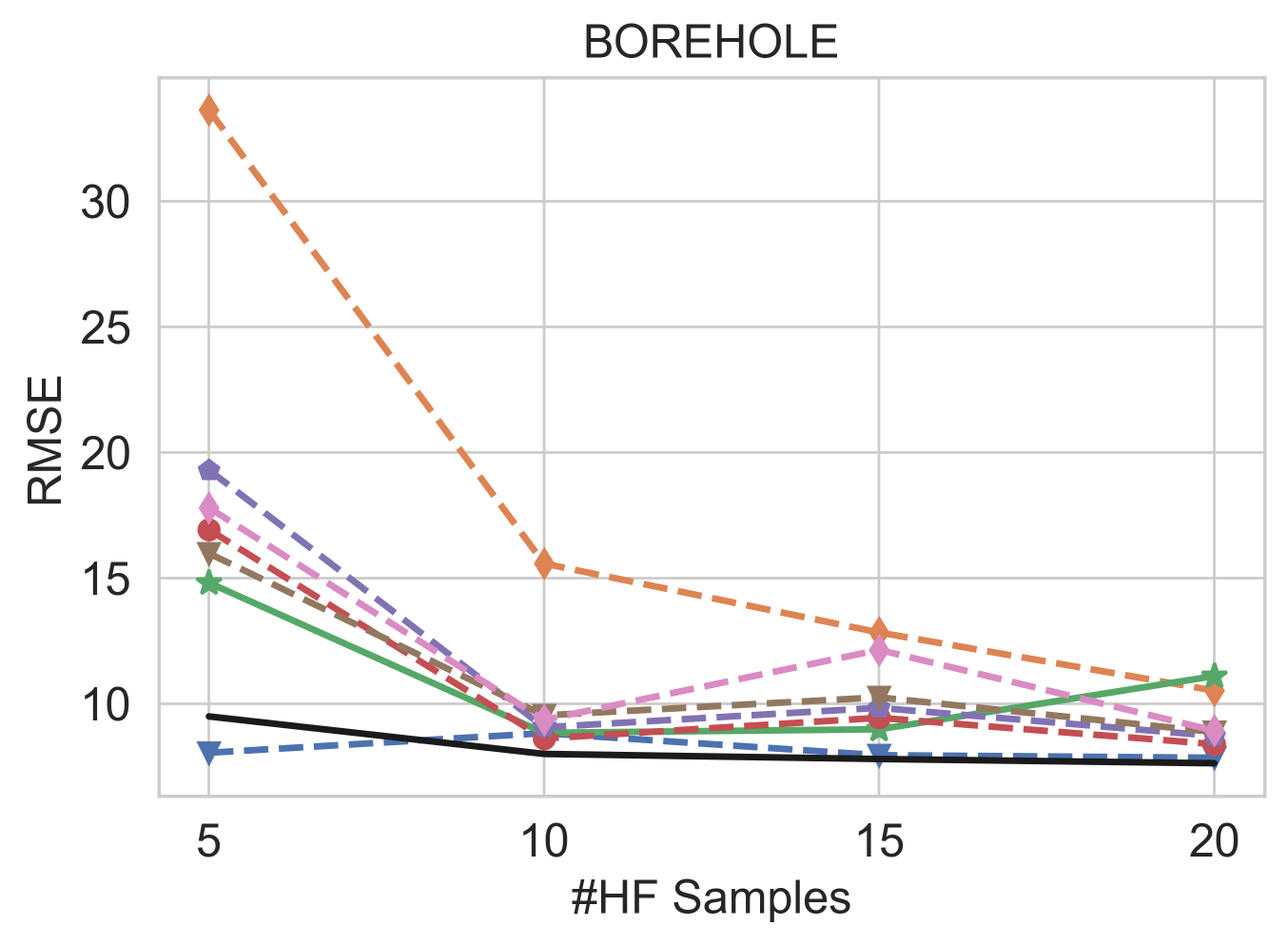}
    \end{minipage}%
    \hfill
    \begin{minipage}[t]{0.33\linewidth}
    \centering
    \includegraphics[width=5cm]{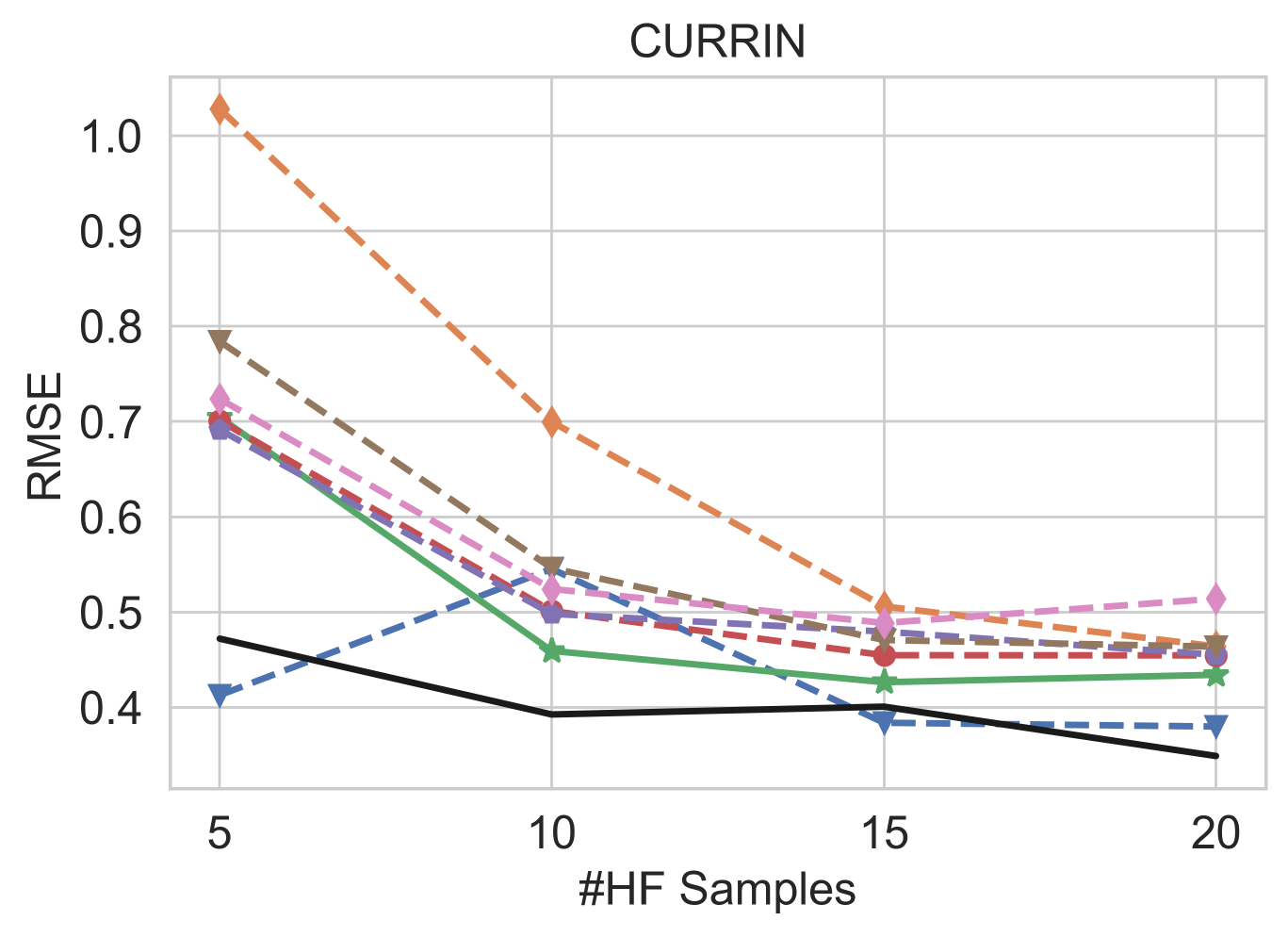}
    \end{minipage}
    \begin{minipage}[t]{0.33\linewidth}
    \centering
    \includegraphics[width=5cm]{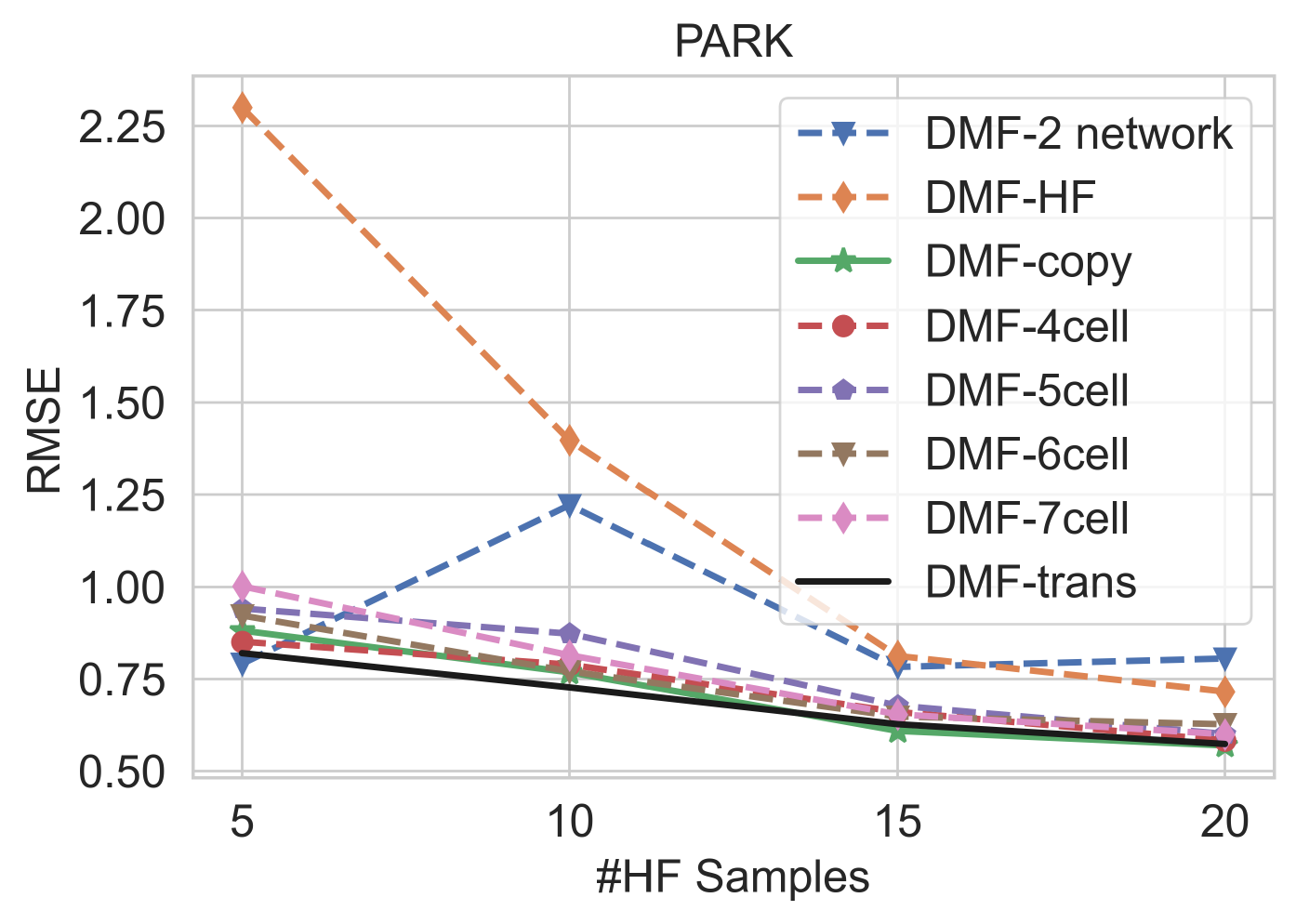}
    \end{minipage}
    \caption{Comparison of the different knowledge transfer frameworks}
    \label{fig10}
\end{figure}
\subsection{Comparison of Different Model Structures}
Except for the proposed transfer learning method, there are also many other frameworks based on DMF. 
To compare the performance of different methods, we try many different possible derived frameworks, including the proposed DMF-2 network and DMF-trans. 
We compare them on three classic benchmark problems, namely, the Borehole, the Currin, and the Park problems (details given in the experimental section).
Fig.~\ref{fig10} shows the root mean square error (RMSE) with the fixed 20 \lf data with an increasing number of \hf training samples. The DMF-HF means only a single DMF is used on \hf data; DMF-copy means a deep copy of \lf model directly without combining the linear layer $\mathcal{NN}_l$ with fine-tuning to the ${\rm DMF}_L$ on the high-fidelity data.
The DMF-$n$cells ($n = 3, 4,\cdots,7$) means ${\rm DMF}_L$ is formed by $n$ cells described in Fig.~\ref{fig0}. 

As can be seen, the performance decreases as the number of cells increases possibly due to the overfitting of the model or the complexity of the model requiring a significant number of training samples.
Therefore, we use the DMF-3cells as the baseline model for DMF-trans, which is the best model among the DMF-$n$cells models and other models including DMF-HF and DMF-copy, and DMF-2 network.

\begin{figure}[htbp]
    \begin{minipage}[t]{0.33\linewidth}
    \centering
    \includegraphics[width=5cm]{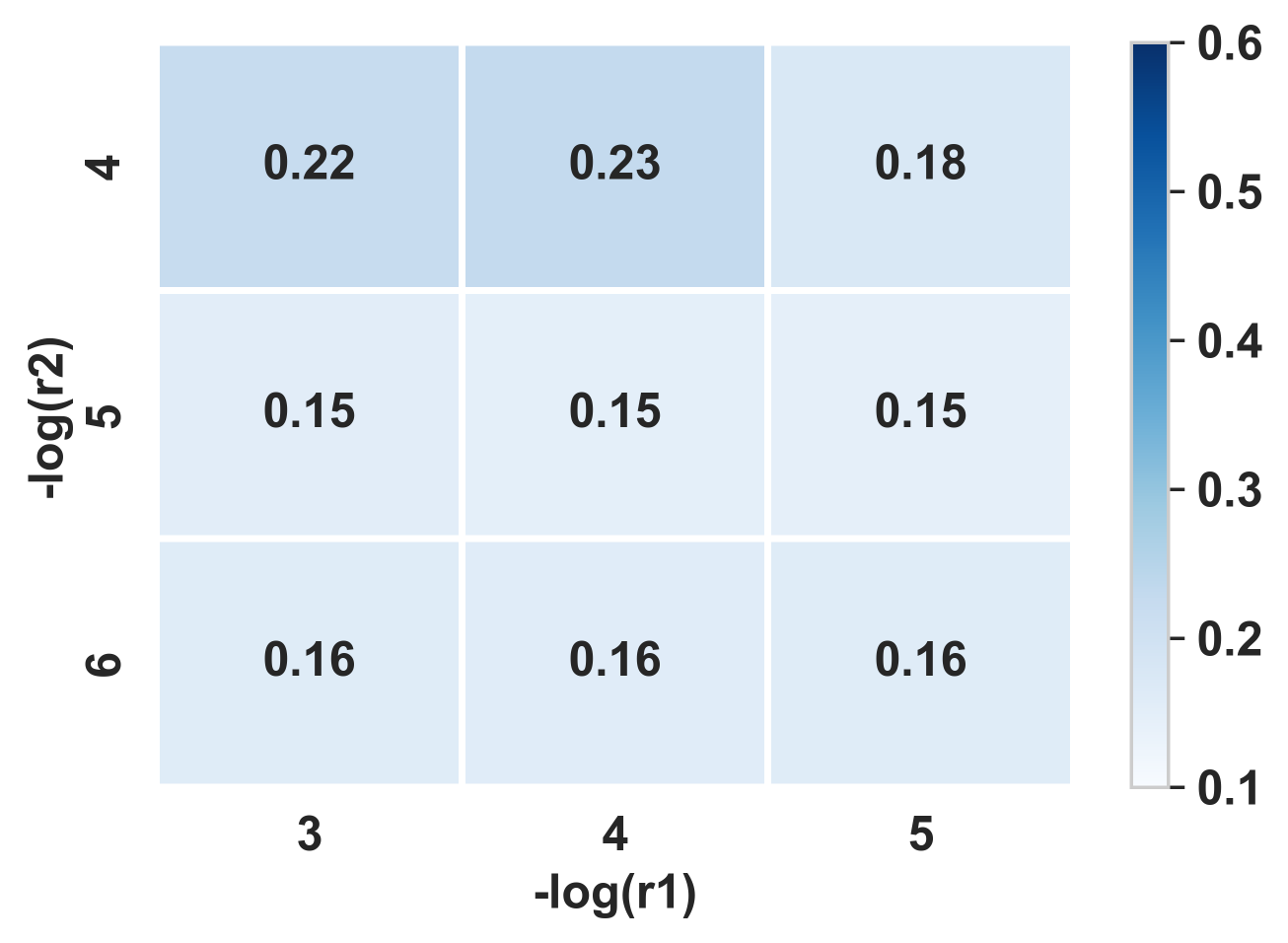}
    \end{minipage}%
    \hfill
    \begin{minipage}[t]{0.33\linewidth}
    \centering
    \includegraphics[width=5cm]{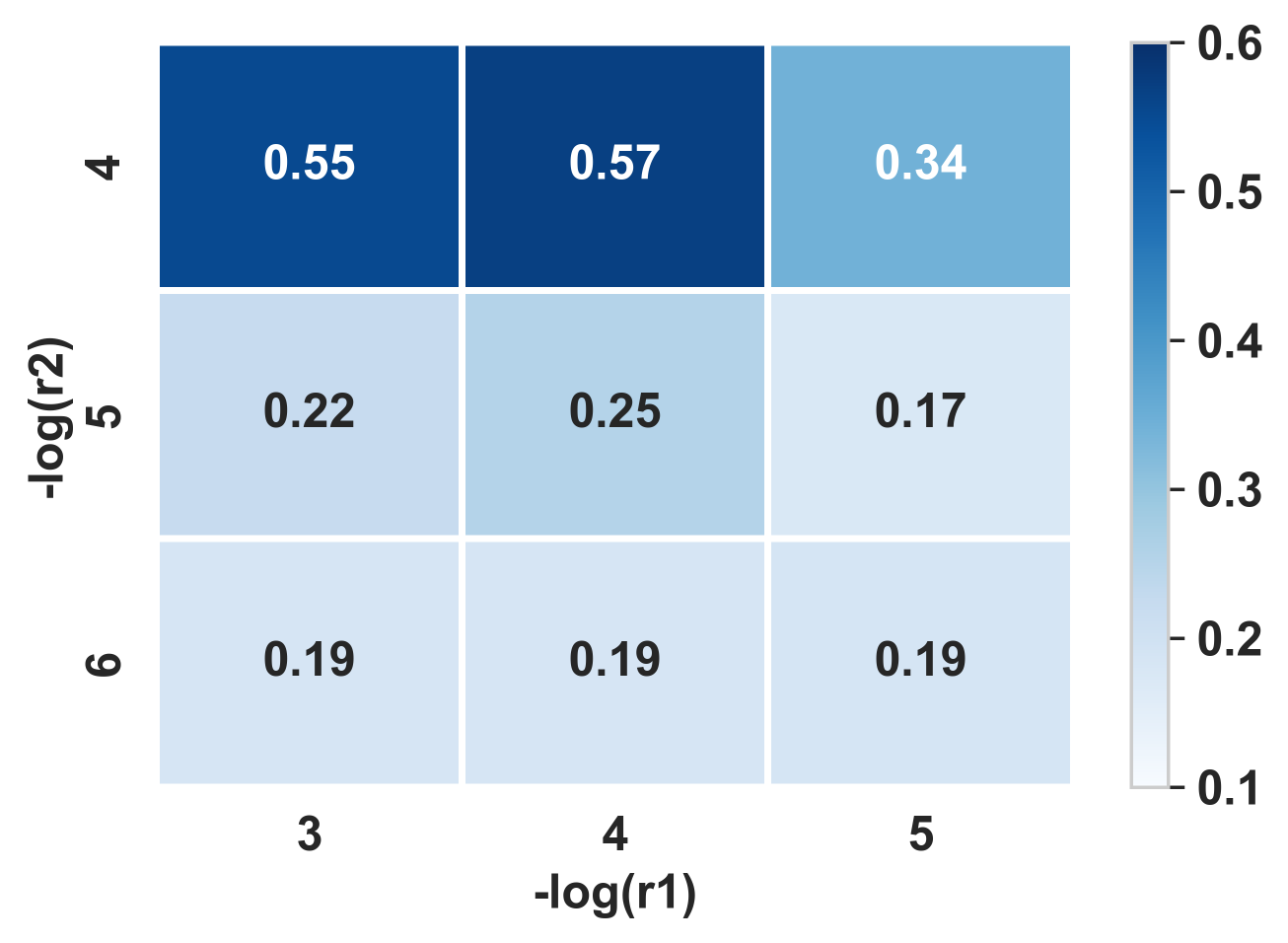}
    \end{minipage}
    \begin{minipage}[t]{0.33\linewidth}
    \centering
    \includegraphics[width=5cm]{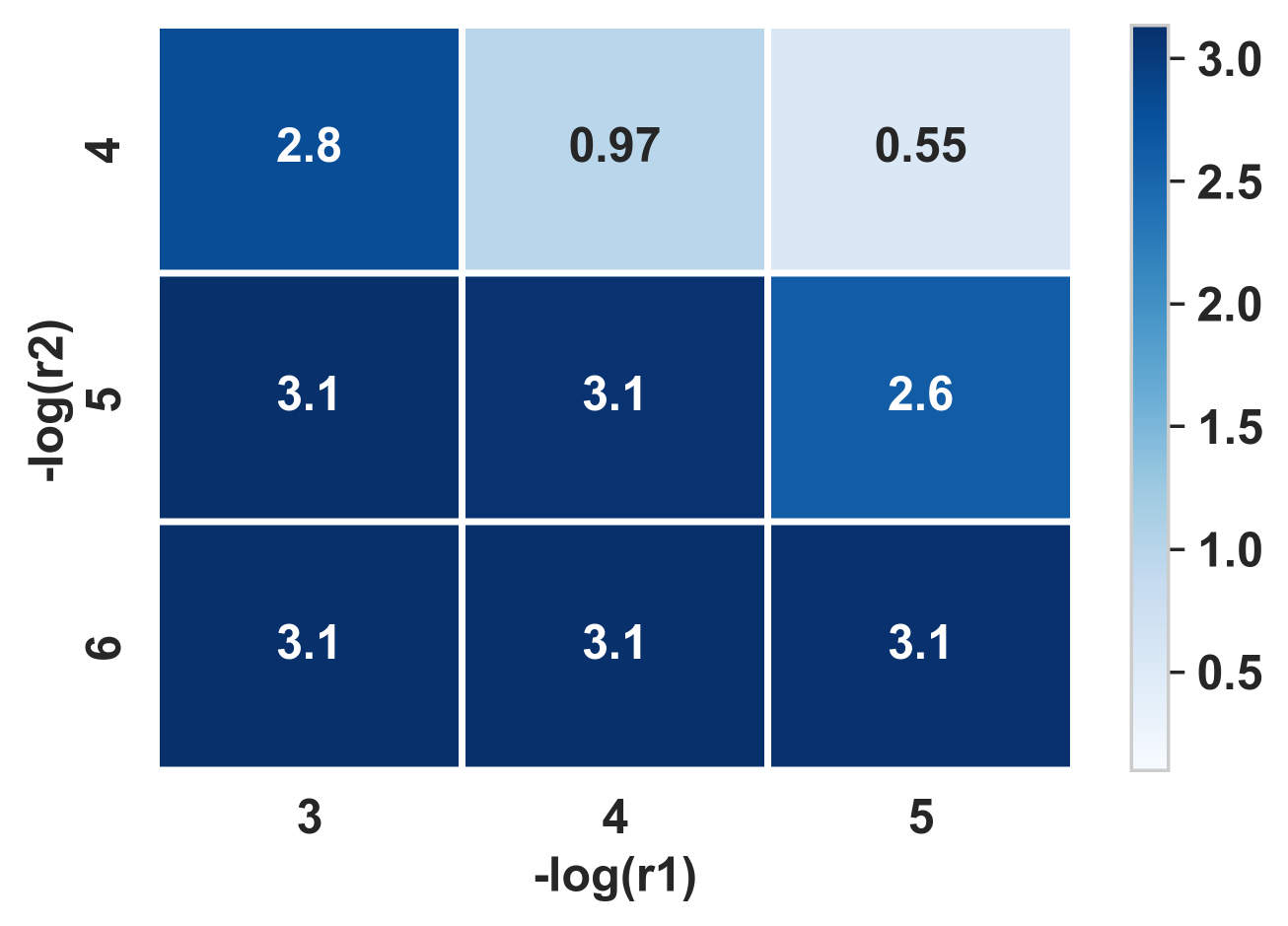}
    \end{minipage}
    \caption{RMSE with different hyperparameters bundle, $r_3=0.01$ (Left), $r_3=0.001$ (Middle), $r_3=0.0001$ (Right)}
    \label{fig_hyper1}
\end{figure}
\subsection{Hyperparameter Search Space Pruning}\label{SH}
Despite the efforts we put into an automatic machine learning surrogate model for multi-fidelity fusion, there are still some hyperparameters that need to be tuned manually. 
These hyperparameters, such as learning rate $\lambda_1$ and weight decay $\lambda_2$ described in Equation~\ref{eq10}, can influence the performance of the proposed method significantly.
To see this, we conduct a hyperparameter grid search experiment on the {Plasmonic simulation} problem (see the experimental section for details). 
We consider three key hyperparameters $r_1$ (the learning rate of $\w_L$), $r_2$ (the learning rate of $\Alpha_L$), and $r_3$ (the learning rate of $\w_{\mathcal{NN}_l}$).
Intuitively, $r_1$ and  $r_2$ decide how much the knowledge is transferred to the \hf model from the low-fidelity model.
The results are shown in Fig.~\ref{fig_hyper1} where the RMSE with different hyperparameter settings are plotted. 
From the figure, we can see clearly that the performance of the model is highly influenced by the hyperparameter settings.

Instead, to deliver a consistent and labour-free performance, we equip the proposed method with a hyperparameter search space pruning technique to automatically prune the hyperparameter search space to keep only a small set of hyperparameter settings that are likely to perform well.

\begin{figure}[]
    \centering
    \includegraphics[width=10cm]{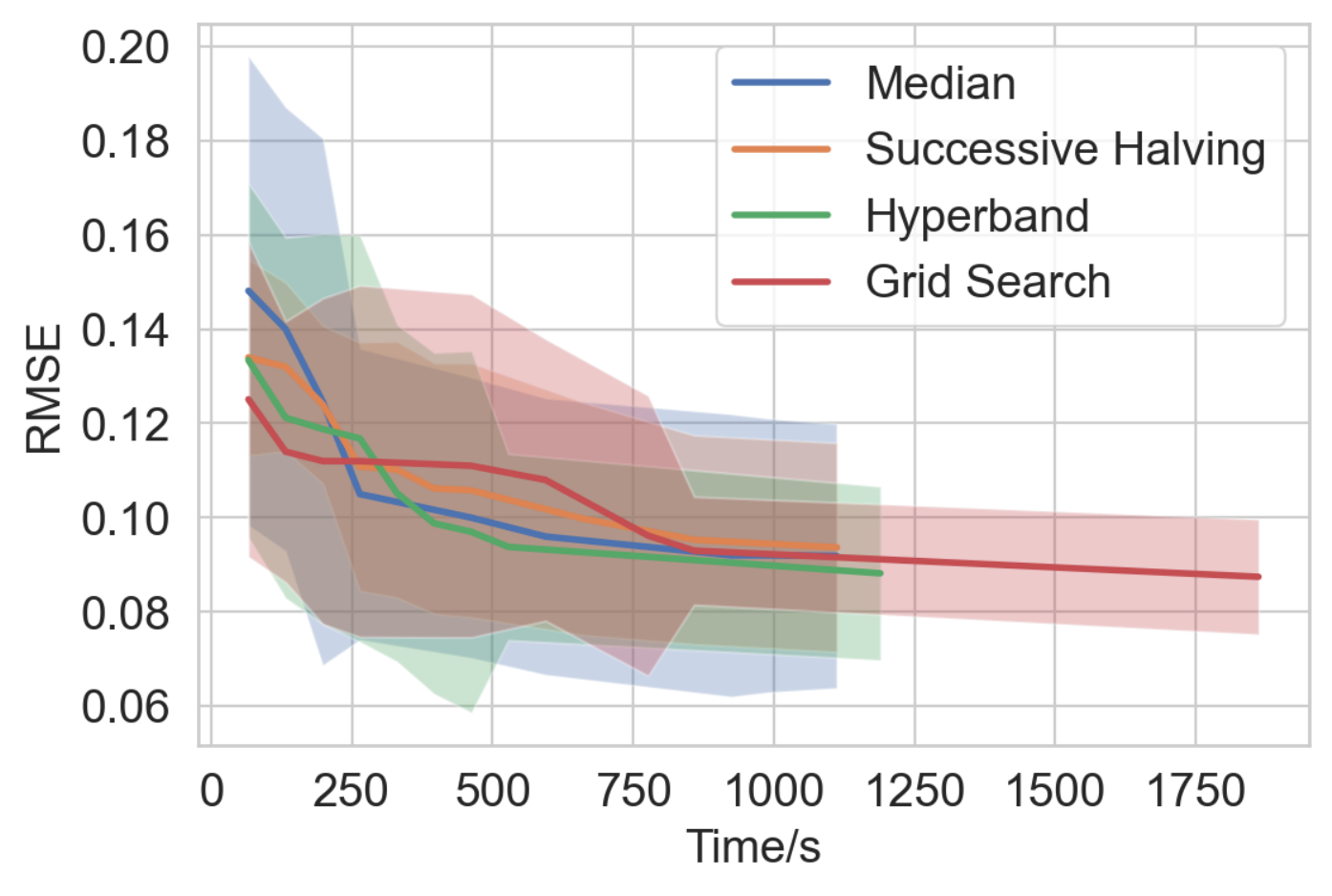}
    \caption{RMSE with different pruning methods. }
    \label{fig_hyper2}
\end{figure}

The most simple and intuitive methods for selecting hyperparameters include grid search and random search. However, the former is costly for our problem, whereas the latter can fall into suboptimal solutions easily without enough computational resources. 
Instead of allocating the computational budget to every hyperparameter setting evenly like in the grid search, pruning techniques try to retain the potentially good settings and discard the ones that produce unsatisfied performance.
The pioneering work median pruning~\citep{he2019filter} runs models with a different setting in parallel and computes their statistical result at each iteration; only settings with loss values higher than the median of the total settings are kept for the next iterations.
The median pruning is prone to discard potentially good candidates at the early stage due to the lack of control over the process.
To resolve this issue, the successive halving algorithm (SHA)~\citep{kumar2018parallel} provides more flexibility by (i) allocating budgets to all configurations in a given rung (usually a small value) at the beginning stage and (ii) keeping only $1/\eta$ of the current configurations for the next iteration. The process is repeated $\eta$ times, where $\eta$ is the rate of elimination.
When the number of hyperparameter configurations is large, the initial budget SHA allocated to each configuration is extremely limited, which leads to inaccurate estimation of performance and bad overall performance.  
As an efficient remedy, Hyperband~\citep{li2017hyperband}, 
extends SHA by repeating it many times with different $\eta$s and budgets, which is called a bracket.
It begins with the most aggressive bracket (\ie large $\eta$), and it then decreases $\eta$ gradually in the following iterations,
which enables the hyperband to perform well in situations where more conservative allocations are required while taking advantage of adaptive allocations. 
In DMF, the number of configurations is large whereas each trial requires adequate training iterations. Thus, we use hyperband as our prime method over others.
We compare their performances on the Plasmonic simulation with the results shown in Fig.~\ref{fig_hyper2}. As can be seen, the pruning technique can reach the optimal point more rapidly with approximately $60\%$  total time  of the grid search. {Hyperband outperforms the other pruning techniques in terms of performance and produces a configuration that is very close to the optimal (founded by grid search with a large budget), which is also consistent with the literature on pruning.}



\section{Experiment}\label{sec: exp}
We assess the performance of the proposed DMF on a series of benchmark problems and real-world applications.
For the simple univariate problems, we compare with the state-of-the-art methods, including GP-HF~\citep{li2020multi}, LAR~\citep{kennedy2000predicting}, and NAR~\citep{perdikaris2017nonlinear}.
For the high-dimensional canonical emulation benchmarks and real-world applications, we use the enhanced LAR and NAR as suggested in generalized AR (GAR)~\citep{wang2023gar} for high-dimension problems.
GAR, deep coregionalization (DC cigp~\citep{xing2015reduced}), and the SOTA deep learning-based method, dmfal~\citep{li2020multi} are also included as the baseline.
All experiments are repeated five times with random seeds from 0 to 4 to deliver a robust and fair comparison. 
The reported performance, \eg RMSE, is the average of the five runs if not stated otherwise.
The experiments are run on Intel CPU i7-8550U and 16GB RAM.


\subsection{Univariate-output Multi-fidelity Benchmark Evaluation}

{Firstly, we make a comparison with the SOTA methods on the three univariate-output benchmarks, including Borehole, Currin, and Park functions.
The Borehole function models water flow through a borehole and is a common function for testing a wide range of methods in computer experiments due to its simplicity and speed of evaluation. The Currin function is a two-dimensional function that occurs multiple times in the computer experiment literature. The Park function is a four-dimensional test function. For the purpose of tuning, Cox et al. ~\citep{cox2001statistical} add a small noise term to the response to simulate the collection of experimental data. } 
The detailed equations of the three functions with explanation and parameter range are explained as follows:

\noindent Borehole:
\begin{align}
    f_H(\mathbf{x})&=\frac{2 \pi T_u\left(H_u-H_l\right)}{\ln \left(r / r_w\right)\left(1+\frac{2 L T_u}{\ln \left(r / r_w\right) r_w^2 K_w}+\frac{T_u}{T_l}\right)},\\
    f_L(\mathbf{x})&=\frac{5 T_u\left(H_u-H_l\right)}{\ln \left(r / r_w\right)\left(1.5+\frac{2 L T_u}{\ln \left(r / r_w\right) r_w^2 K_w}+\frac{T_u}{T_l}\right)}
\end{align}
with the inputs variables listed in Table ~\ref{tab0}:
\begin{table}[H]
\centering
\begin{tabular}{|l|l|}
\hline$r_w \in[0.05,0.15]$ & radius of borehole $(\mathrm{m})$ \\
\hline$r \in[100,50000]$ & radius of influence $(\mathrm{m})$ \\
\hline $T_u \in[63070,115600]$ & transmissivity of upper aquifer $\left(\mathrm{m}^2 / \mathrm{yr}\right)$ \\
\hline $H_u \in[990,1110]$ & potentiometric head of upper aquifer $(\mathrm{m})$ \\
\hline $T_l \in[63.1,116]$ & transmissivity of lower aquifer $\left(\mathrm{m}^2 / \mathrm{yr}\right)$ \\
\hline $H_l \in[700,820]$ & potentiometric head of lower aquifer $(\mathrm{m})$ \\
\hline $L \in[1120,1680]$ & length of borehole (m) \\
\hline $K_w \in[9855,12045]$ & hydraulic conductivity of borehole $(\mathrm{m} / \mathrm{yr})$ \\
\hline
\end{tabular}
\caption{Inputs variables and domain of Borehole function}
\label{tab0}
\end{table}
\noindent Currin:
\begin{align}
    f_H(\mathbf{x})&=\left[1-\exp \left(-\frac{1}{2 x_2}\right)\right] \frac{2300 x_1^3+1900 x_1^2+2092 x_1+60}{100 x_1^3+500 x_1^2+4 x_1+20},  \\
     f_L(\mathbf{x}) & =\frac{1}{4}\left[f\left(x_1+0.05, x_2+0.05\right)+f\left(x_1+0.05, \max \left(0, x_2-0.05\right)\right)\right] \\ & +\frac{1}{4}\left[f\left(x_1-0.05, x_2+0.05\right)+f\left(x_1-0.05, \max \left(0, x_2-0.05\right)\right)\right], \quad x_i  \in [0, 1], i = 1, 2.
\end{align}
and Park:
\begin{align}
    f_H(\mathbf{x})&=\frac{x_1}{2}\left[\sqrt{1+\left(x_2+x_3^2\right) \frac{x_4}{x_1^2}}-1\right]+\left(x_1+3 x_4\right) \exp \left[1+\sin \left(x_3\right)\right],  \\
    f_L(\mathbf{x})&=\left[1+\frac{\sin \left(x_1\right)}{10}\right] f(\mathbf{x})-2 x_1+x_2^2+x_3^2+0.5, \quad x_i \in [0, 1), i = 1, 2, 3, 4.
\end{align}

\begin{figure}[]
    \begin{minipage}[t]{0.33\linewidth}
    \centering
    \includegraphics[width=5cm]{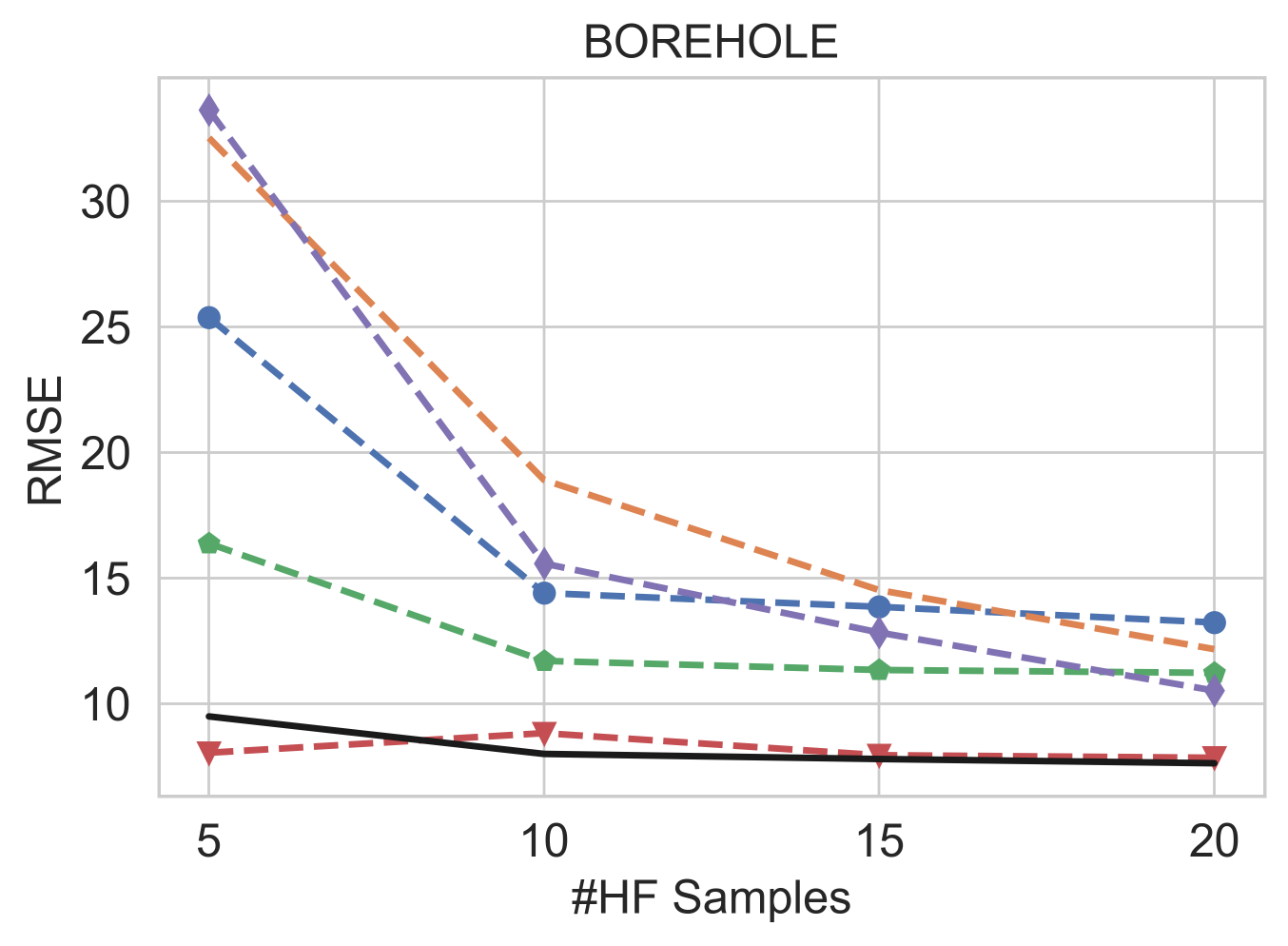}
    \end{minipage}%
    \hfill
    \begin{minipage}[t]{0.33\linewidth}
    \centering
    \includegraphics[width=5cm]{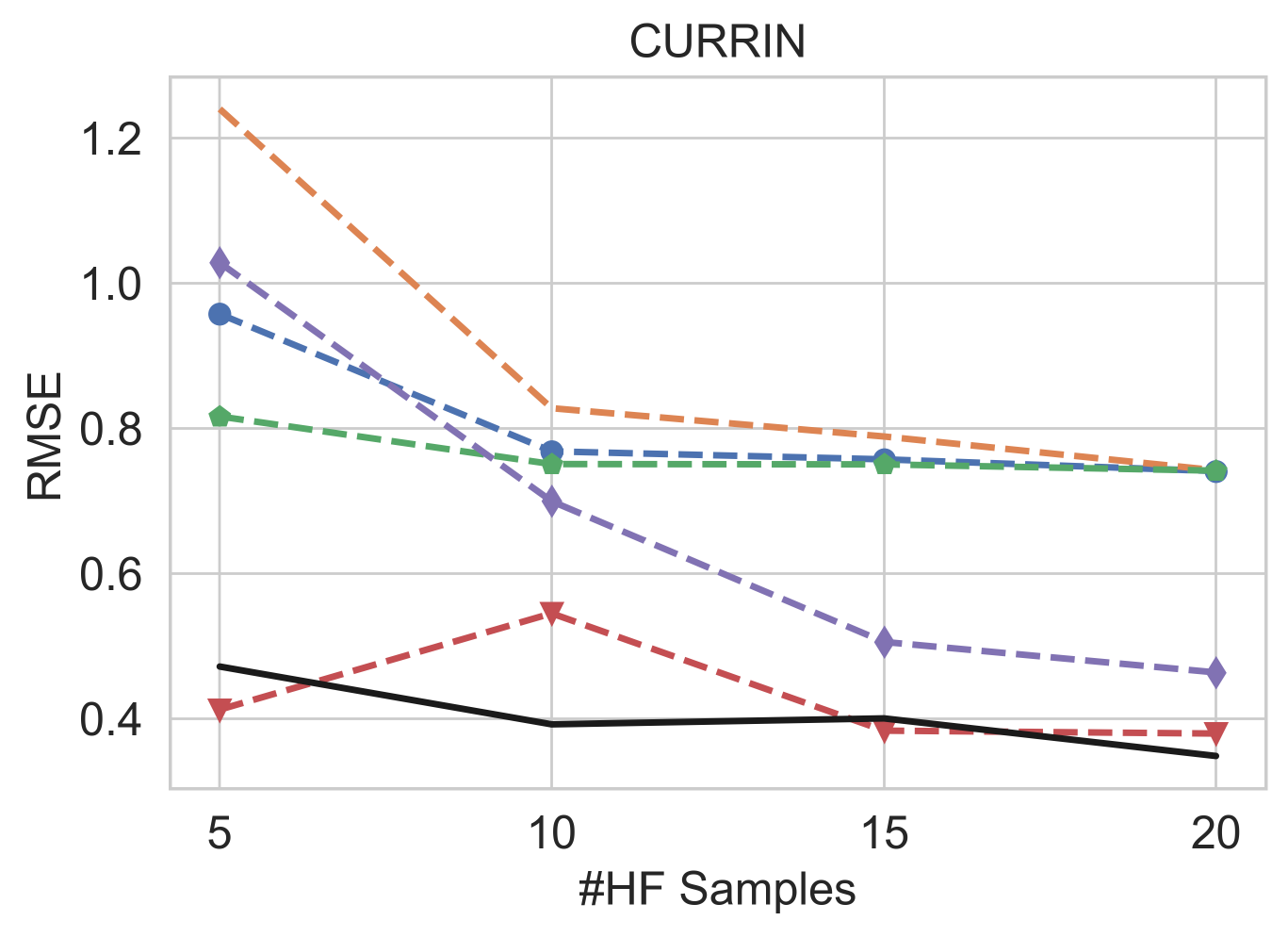}
    \end{minipage}
    \begin{minipage}[t]{0.33\linewidth}
    \centering
    \includegraphics[width=5cm]{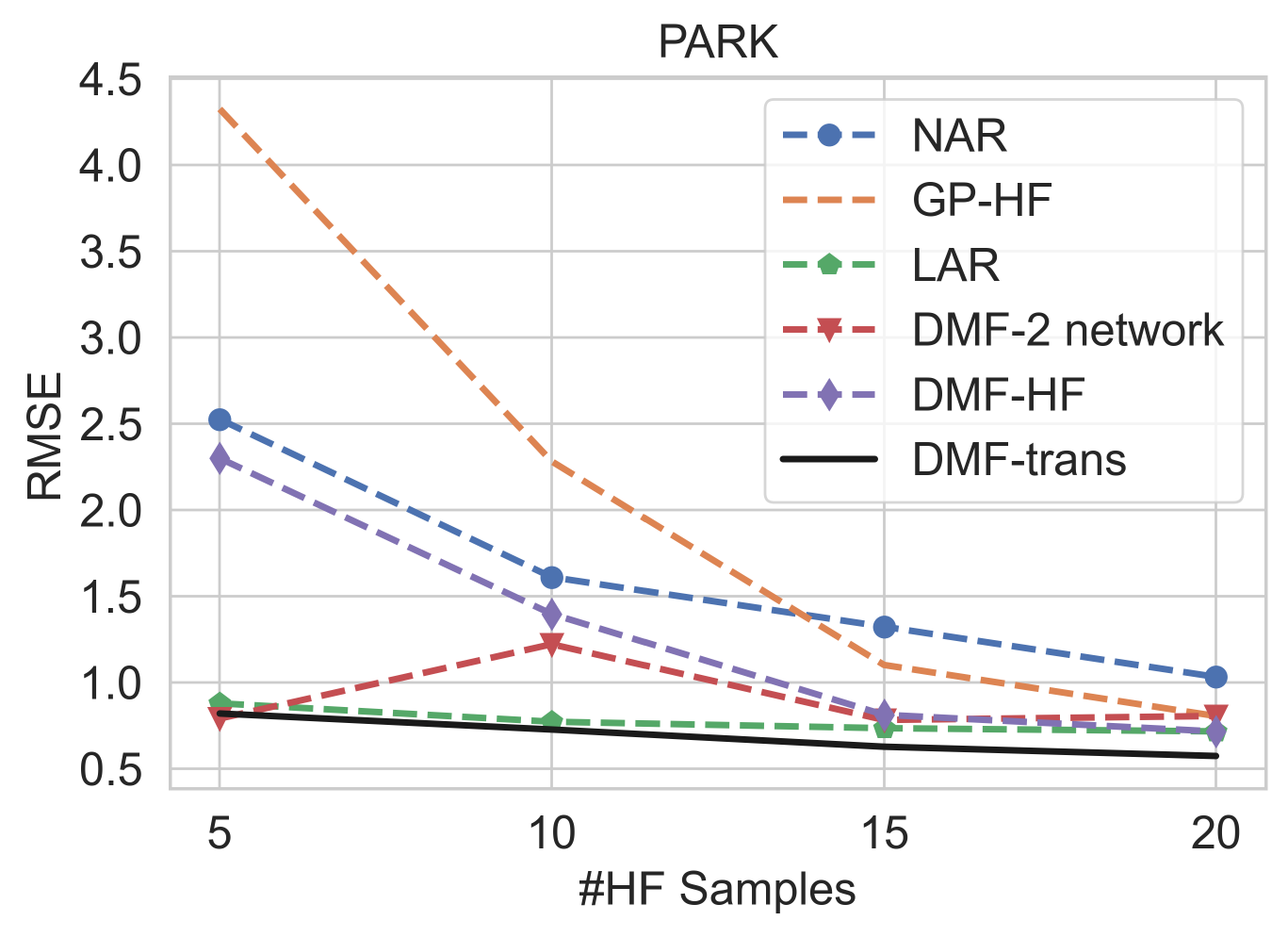}
    \end{minipage}
    \caption{RMSE against an Increasing number of high-fidelity training samples for uni-variate benchmarks. }
    \label{exp1}
\end{figure}

We randomly generate 20 low-fidelity samples and increase the number of high-fidelity samples for each dataset. Then, we evaluate each model on 50 preserved test data points.
The results are shown in Fig.~\ref{exp1}. 
We can see that the proposed DMF-trans almost always achieves the best performance in most cases.
DMF-2 network also obtains similar performance for the Borehole function but slightly worse accuracy for Currin and Park functions.
Popular and classic multi-fidelity fusion methods, such as LAR and NAR, are also included in the comparison. Their performance varies according to the problem, which is consistent with the results found in the literature due to their different assumptions and limitations.

\begin{figure}[htbp]
    \begin{minipage}[t]{0.33\linewidth}
    \centering
    \includegraphics[width=5cm]{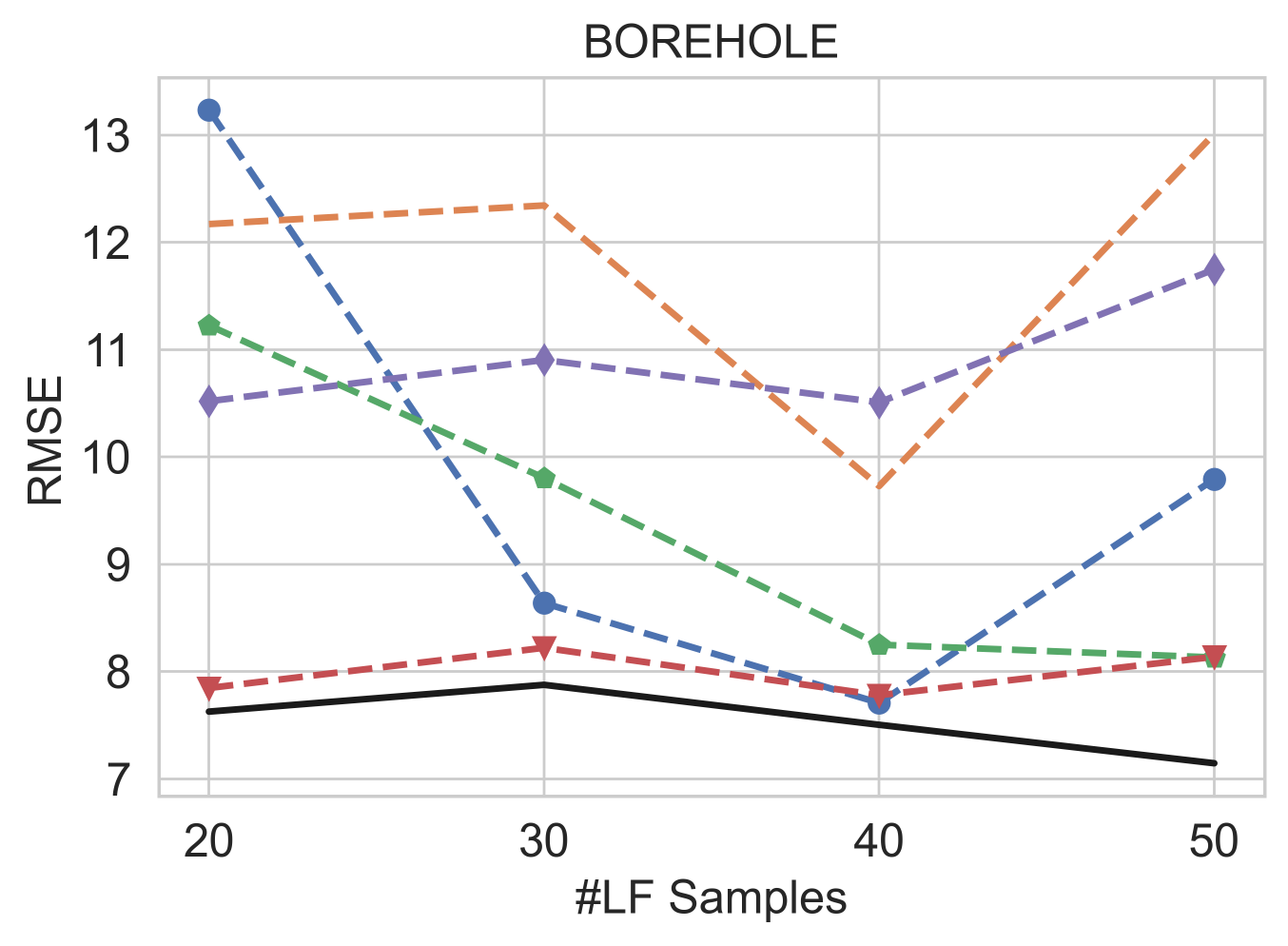}
    \end{minipage}%
    \hfill
    \begin{minipage}[t]{0.33\linewidth}
    \centering
    \includegraphics[width=5cm]{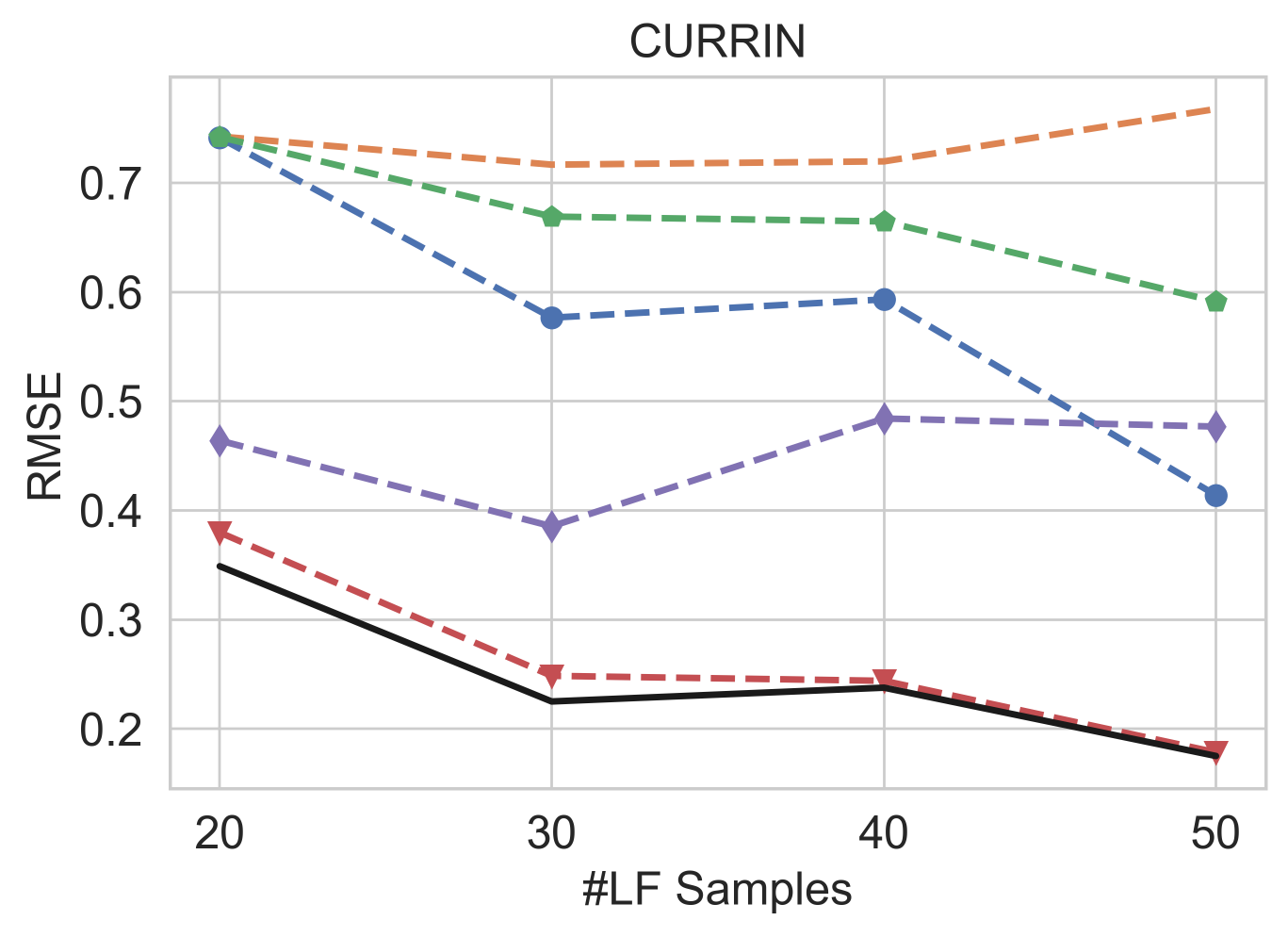}
    \end{minipage}
    \begin{minipage}[t]{0.33\linewidth}
    \centering
    \includegraphics[width=5cm]{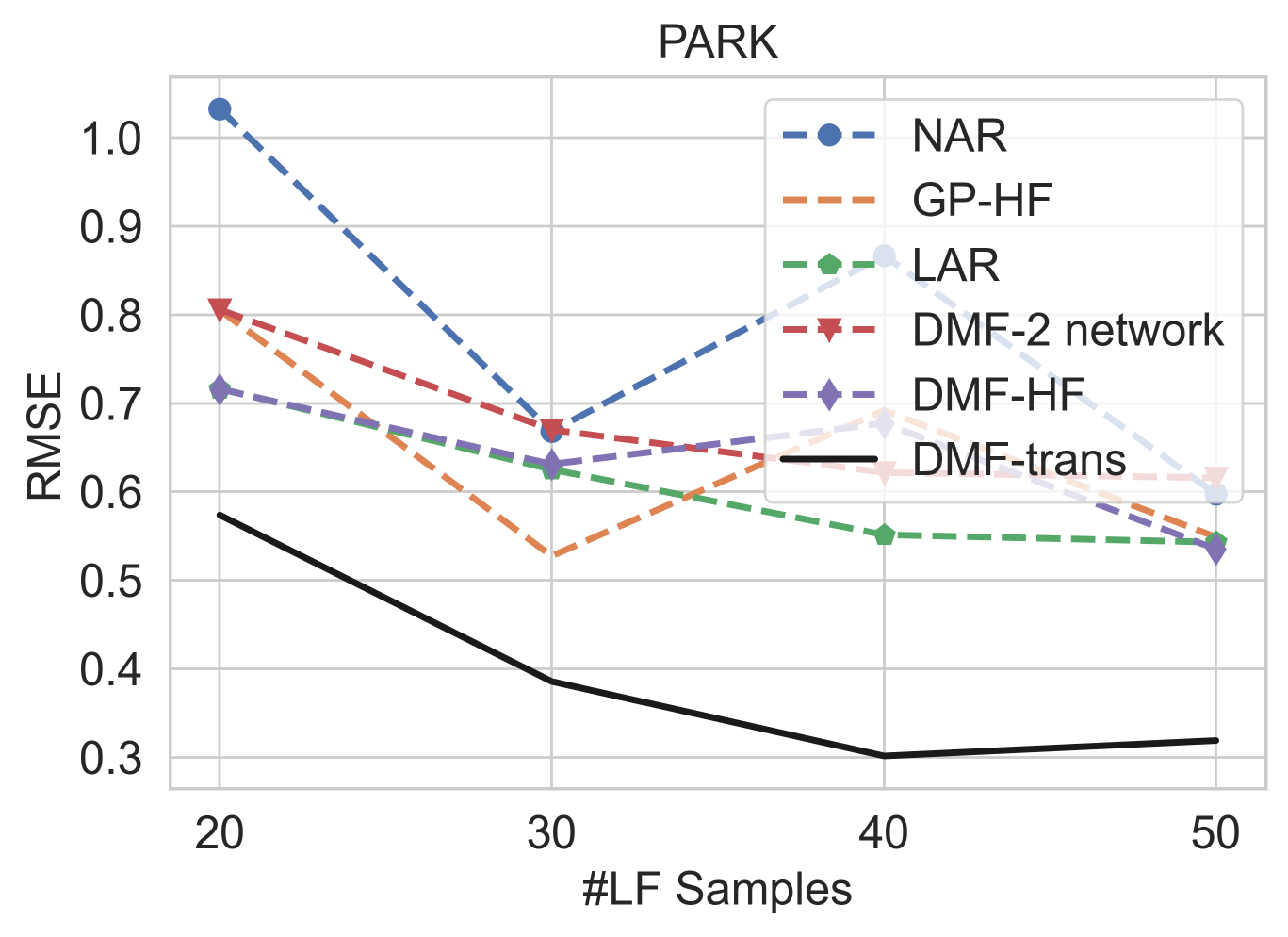}
    \end{minipage}
    \caption{RMSE against an Increasing number of low-fidelity training samples for uni-variate benchmarks. }
    \label{exp1-2}
\end{figure}

To investigate the influences of the number of low-fidelity data, we conduct the same experiment with fixed 20 high-fidelity samples and gradually increase the number of low-fidelity samples from 20 to 50. The results are shown in Fig.~\ref{exp1-2}.
As can be seen in these figures, DMFs (DMF-2 network and DMF-trans) outperform other SOTA methods in three datasets. 
With further analysis, when the low-fidelity data are inadequate, \eg 20 samples or less, the performance of DMFs does not change considerably with the increment of high-fidelity data. If the high-fidelity data is extremely scarce (less than five samples), the DMF-2 network is more effective than DMF-trans. However, when the low-fidelity data increase, the validation RMSE of DMFs reduces notably. Moreover, DMF-trans becomes the most accurate model, especially for the Park function. It shows the efficiency of transfer learning, which successfully applies knowledge learning on low-fidelity data to building the high-fidelity model. Increasing the number of low-fidelity data, which is easy to acquire, will notably improve DMF's performance.

\subsection{High-Dimensional and Spatial-Temporal Canonical Physics Simulation Benchmarks}
To assess the effectiveness of \ours for the complex but practical high-dimensional physics simulations, we conduct experiments on approximating two canonical PDE simulations, which produce high-dimensional spatial-temporal fields. 
Specifically, we test on Burger's equation and Poisson's equation as in many previous works~\citep{tuo2014surrogate,raissi2017machine,xing2021deep}. These two PDEs play crucial roles in many physical and engineering problems.
 Besides, they result in common simulation challenges, such as high-dimensional spatial-temporal field outputs, nonlinearities, and discontinuities. 
 Thus, they are commonly taken as the benchmarks for the multi-fidelity fusion~\citep{xing2021deep,tuo2014surrogate,raissi2017machine}. 
 For clarity, we follow the convention and let $x$ and $y$ denote the spatial coordinates and $t$ denote the time coordinate, which slightly abuses the notation in the main test where $\x$ indicates the model inputs.



\begin{figure}[htbp]
    \centering
    \includegraphics[width=6cm]{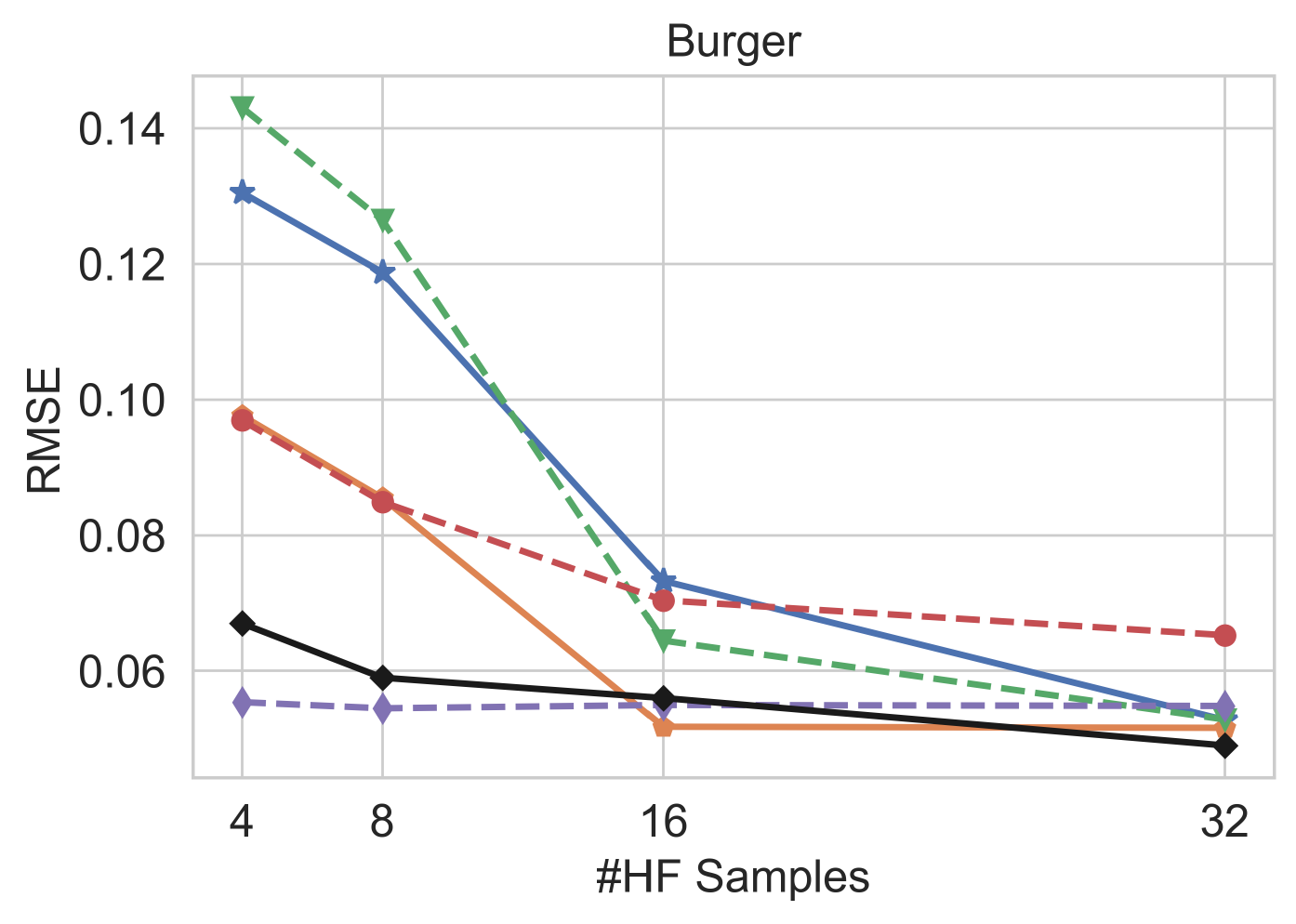}
    \includegraphics[width=6cm]{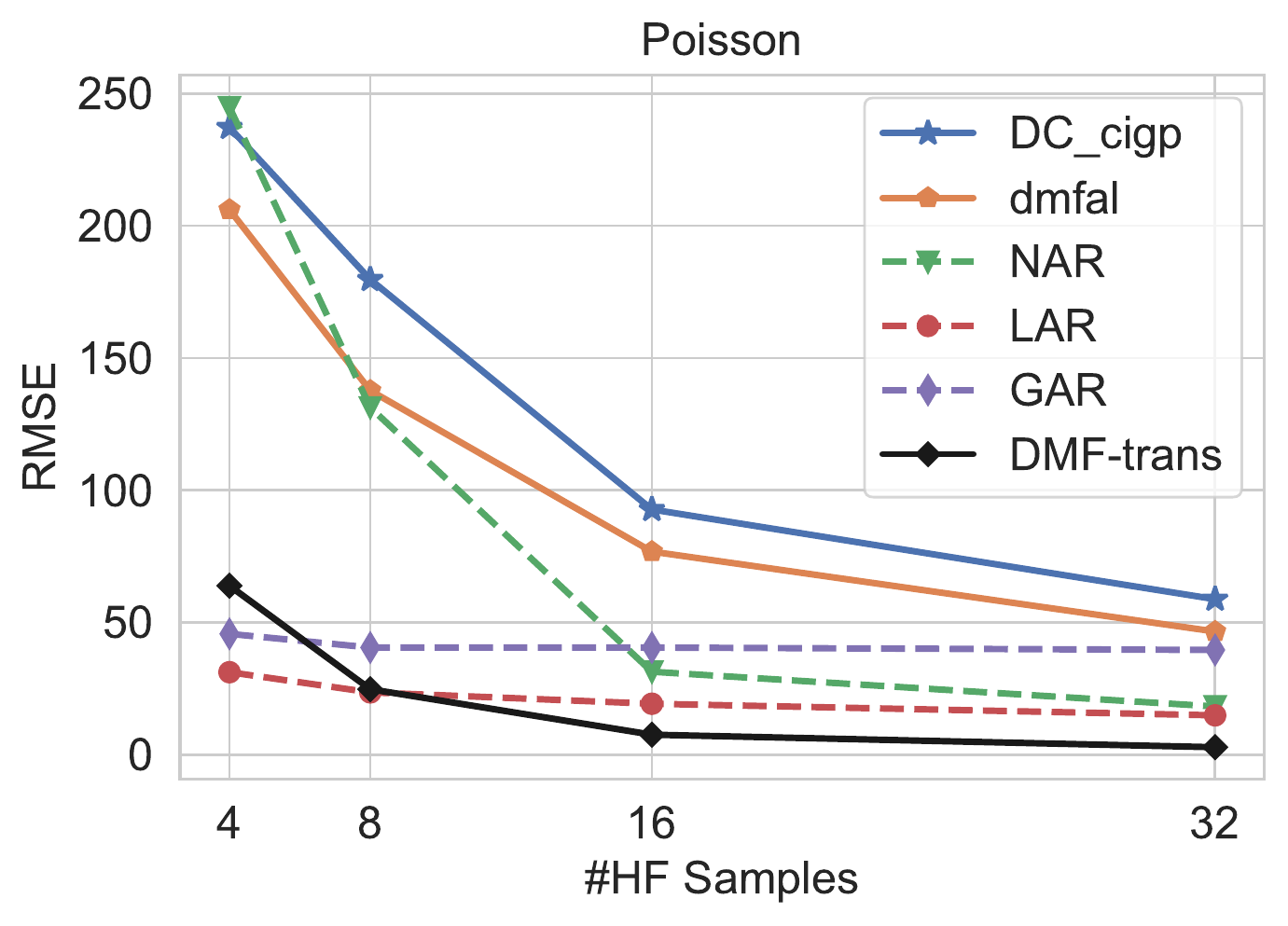}
    \caption{RMSE against an Increasing number of high-fidelity training samples for the Burger's and Poisson's equations.}
    \label{exp22}
\end{figure}
\noindent \textbf{Burgers' equation} is regarded as a standard nonlinear hyperbolic PDE; it is widely used to represent a variety of physical phenomena, including fluid dynamics~\citep{chung2002computational}, nonlinear acoustics~\citep{sugimoto1991burgers}, and traffic flows~\citep{nagel1996particle}. The viscous version of this equation is given by 
\begin{equation}
     \frac{\partial u}{\partial t} + u \frac{\partial u}{\partial x} = v \frac{\partial^2 u}{\partial x^2},
\end{equation}
  where $u$ stands for the volume, $x$ represents a spatial location, $t$ indicates the time, and $v$ denotes the viscosity. With homogeneous Dirichlet boundary conditions, we set $x\in[0,1]$\cmt{(in $\si{\meter}$)}, $t \in [0,3]$\cmt{ (in $\si{\second}$)}, and $u(x,0)=\sin(x\pi/2)$.  We uniformly sampled viscosities $v \in [0.001,0.1]$\cmt{ (in $ \si[inter-unit-product = \ensuremath{{}\cdot{}}] {\milli\pascal\second}$)} as the surrogate input parameter to generate the solution fields. 

The PDE is solved using the backward Euler and finite element methods in the space and time domains, respectively. For the first (low-fidelity) solution, the spatial-temporal domain is discretized into $16\times16$ regular rectangular mesh. Higher-fidelity solvers double the number of nodes in each dimension of the mesh, \eg $32\times32$ for the high fidelity.
In our experiment, the result fields (\ie outputs) are recorded using a $64\times64$ regular spatial-temporal mesh. {Due to the variation of dimensions of different fidelity data,  we utilize interpolation to upscale the \lf and \hf fields to fields with $100\times 100$ regular nodes}.
At the beginning of the experiment, we flatten the outputs as vectors and perform the experiment with a bundle of hyperparameters, including learning rates mentioned in Section~\ref{SH}. We perform the pruning algorithm to choose the best hyperparameters with the lowest RMSE.  
Similar to the previous experiments, we gradually increase the number of high-fidelity training samples to 32 with the number of low-fidelity training samples being fixed to 32.
The results are shown in Fig.~\ref{exp22}.
We can see that \ours gives robust and good performance across the range. More specifically, \ours achieves the second lowest error with a low number of training data (\ie 4 and 8); the RMSE continues to drop as more training data becomes available. 
With 32 high-fidelity training points, \ours obtains the lowest RMSE, which is expected as our method generally requires more data to well construct a proper model from scratch.
In contrast, GAR shows consistently good performance but fails to improve significantly with more training data, which is consistent with its original work. Other baselines, such as LAR and NAR, are significantly outperformed by \ours for almost all settings.

\begin{figure}[htbp]
    \begin{minipage}[t]{0.2\linewidth}
    \centering
    \includegraphics[width=3.25cm]{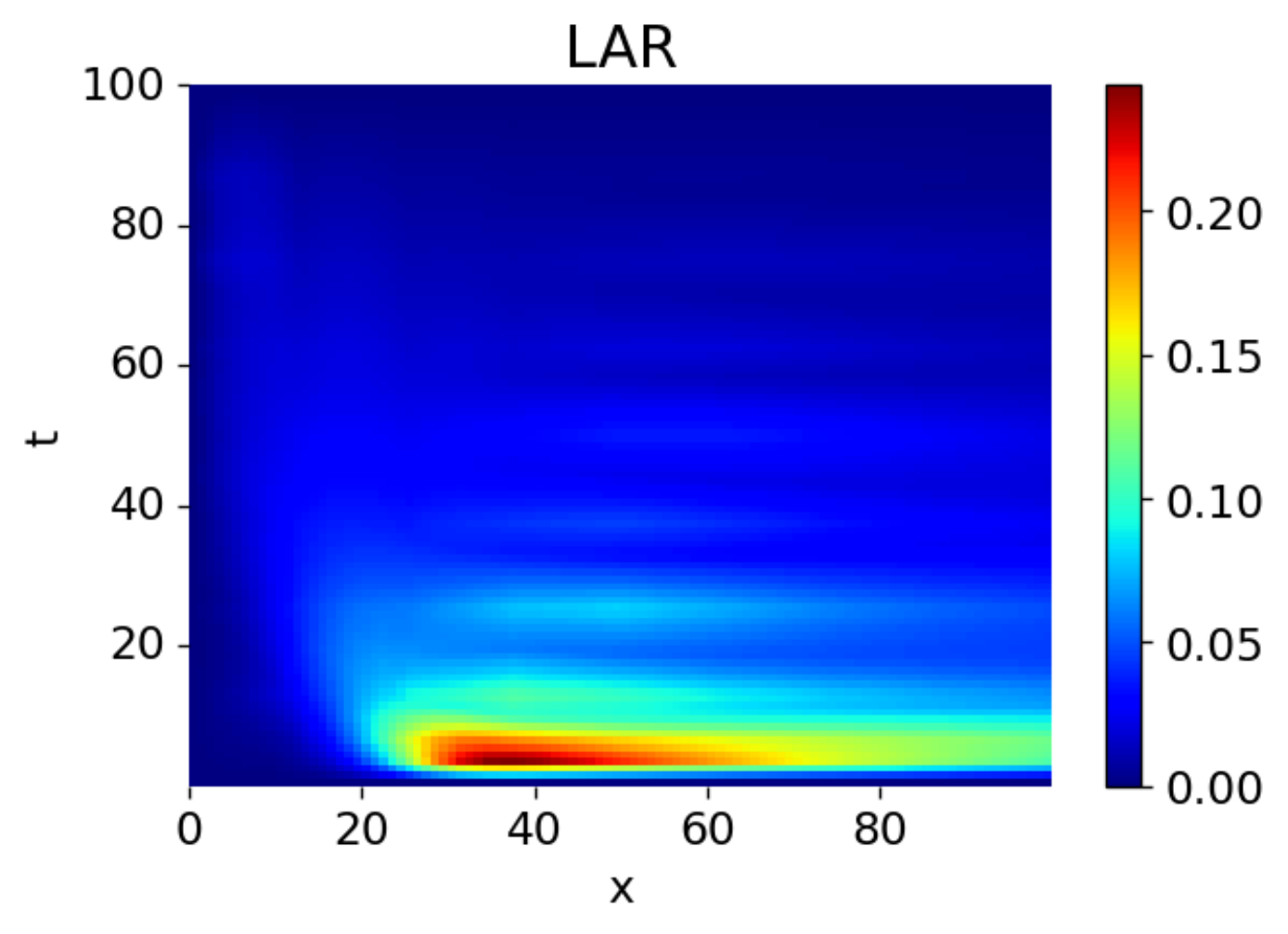}
    \end{minipage}%
    \hfill
    \begin{minipage}[t]{0.2\linewidth}
    \centering
    \includegraphics[width=3.25cm]{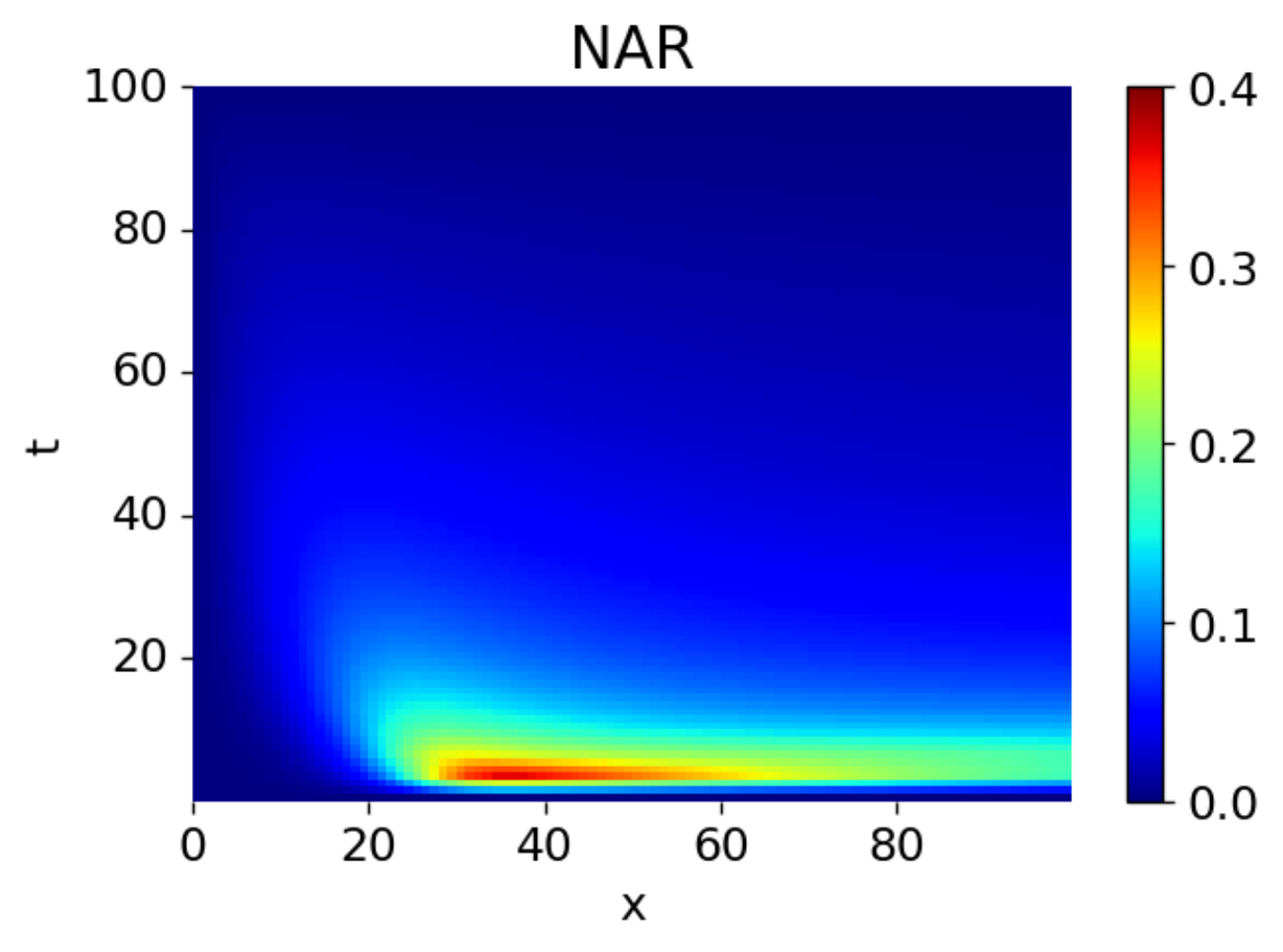}
    \end{minipage}%
    \begin{minipage}[t]{0.2\linewidth}
    \centering
    \includegraphics[width=3.25cm]{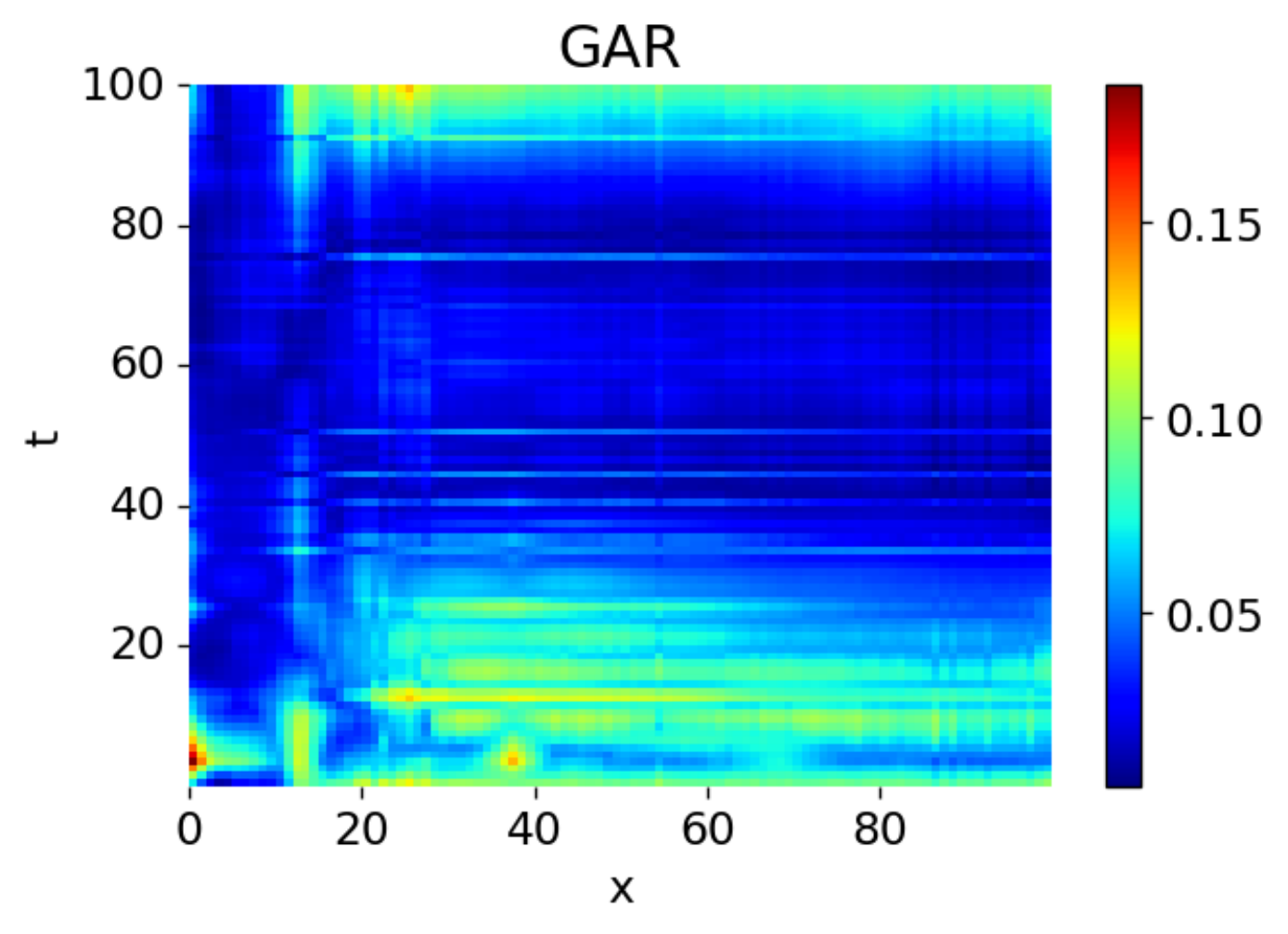}
    \end{minipage}%
    \hfill
    \begin{minipage}[t]{0.2\linewidth}
    \centering
    \includegraphics[width=3.25cm]{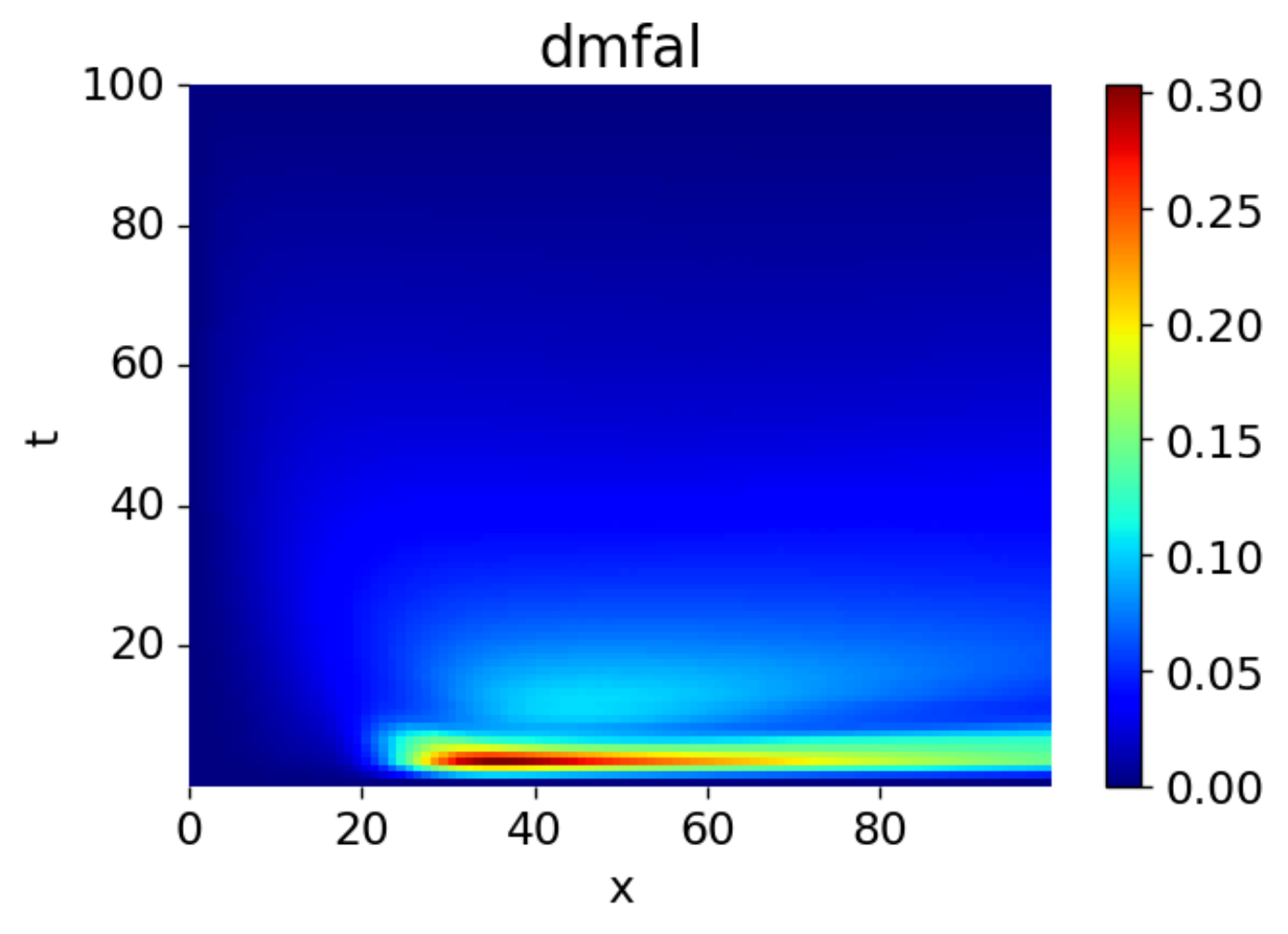}
    \end{minipage}%
    \hfill
    \begin{minipage}[t]{0.2\linewidth}
    \centering
    \includegraphics[width=3.25cm]{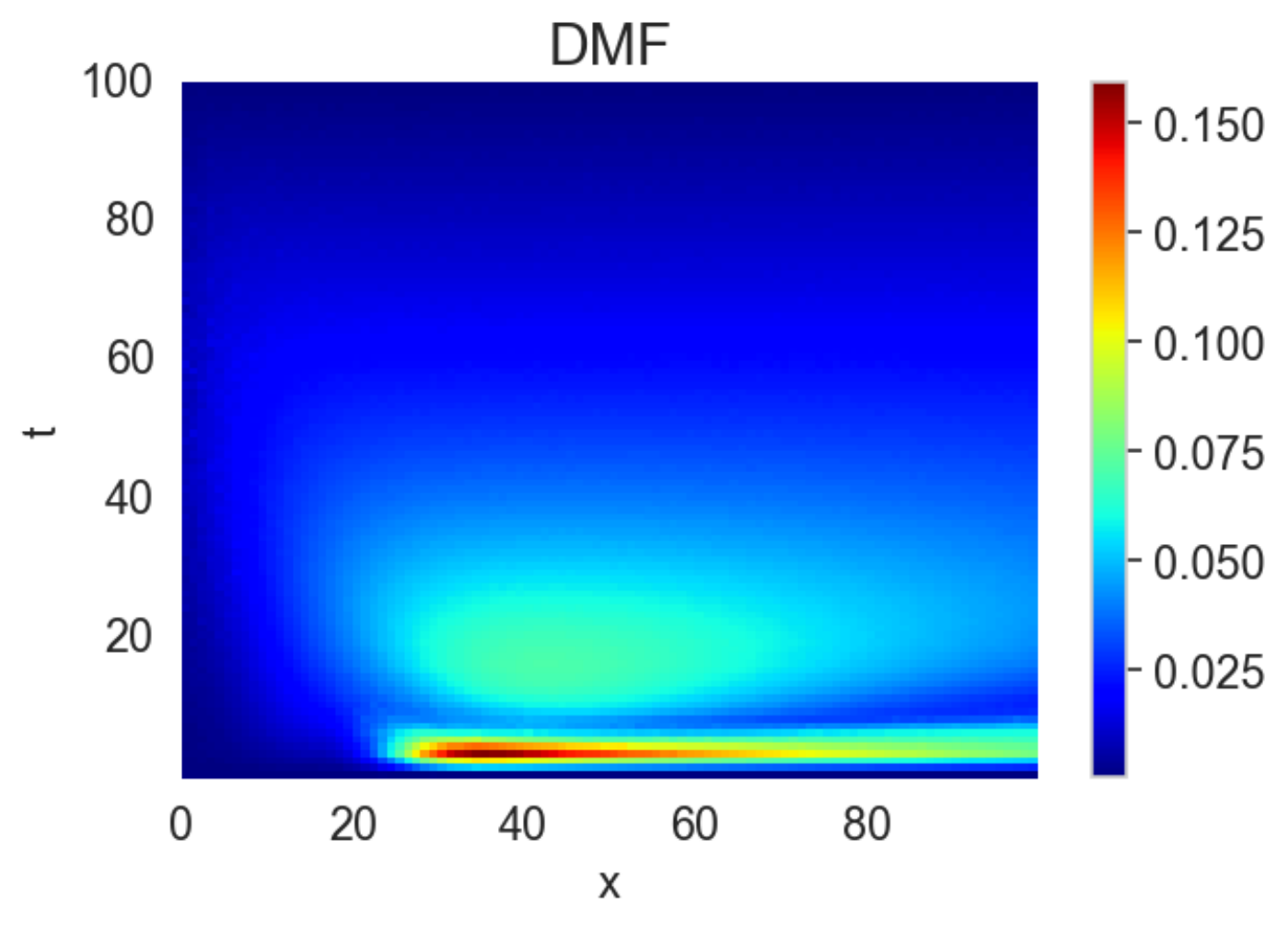}
    \end{minipage}%
    \hfill
    \\
    \begin{minipage}[t]{0.2\linewidth}
    \centering
    \includegraphics[width=3.25cm]{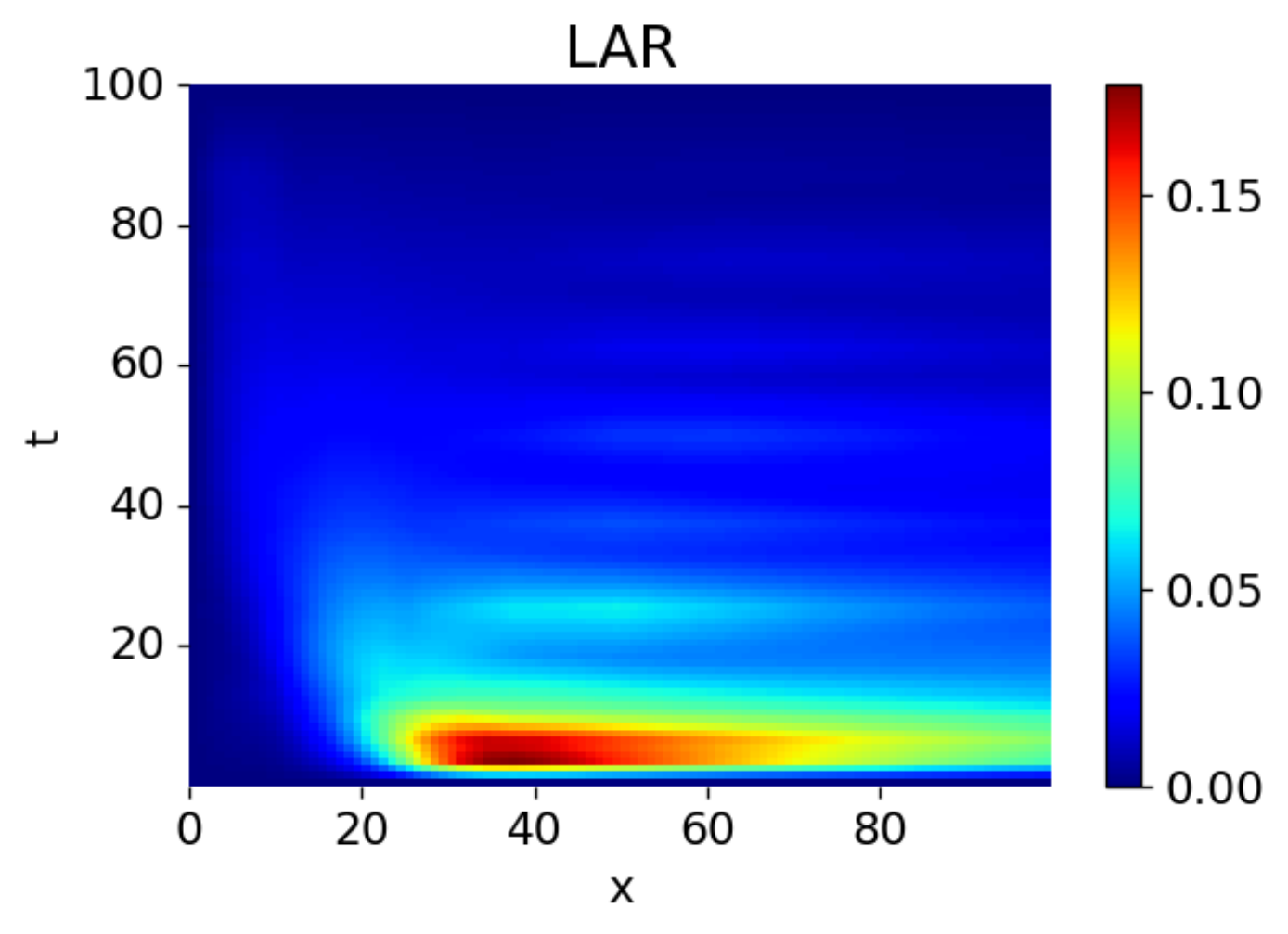}
    \end{minipage}%
    \hfill
    \begin{minipage}[t]{0.2\linewidth}
    \centering
    \includegraphics[width=3.25cm]{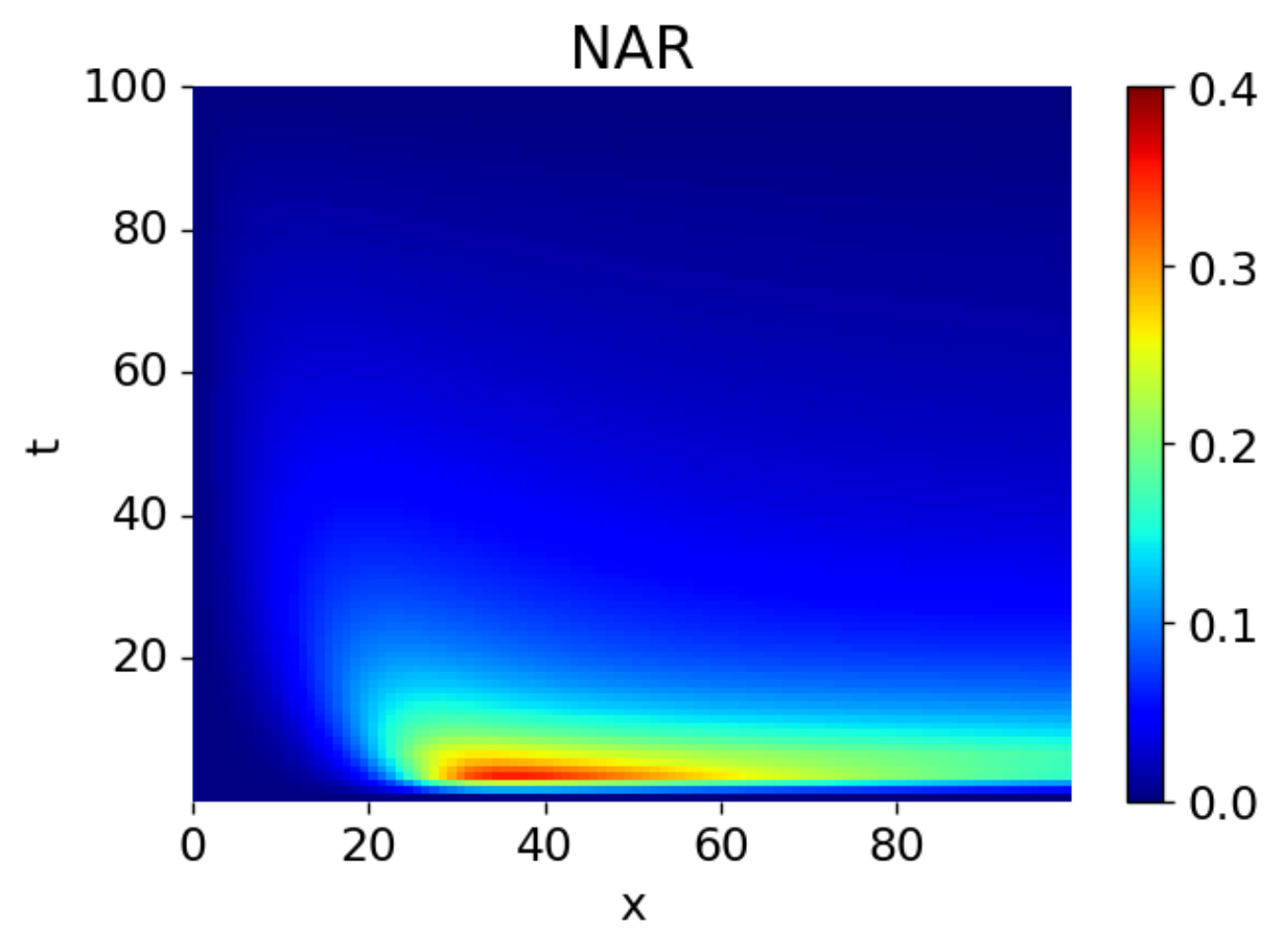}
    \end{minipage}%
    \begin{minipage}[t]{0.2\linewidth}
    \centering
    \includegraphics[width=3.25cm]{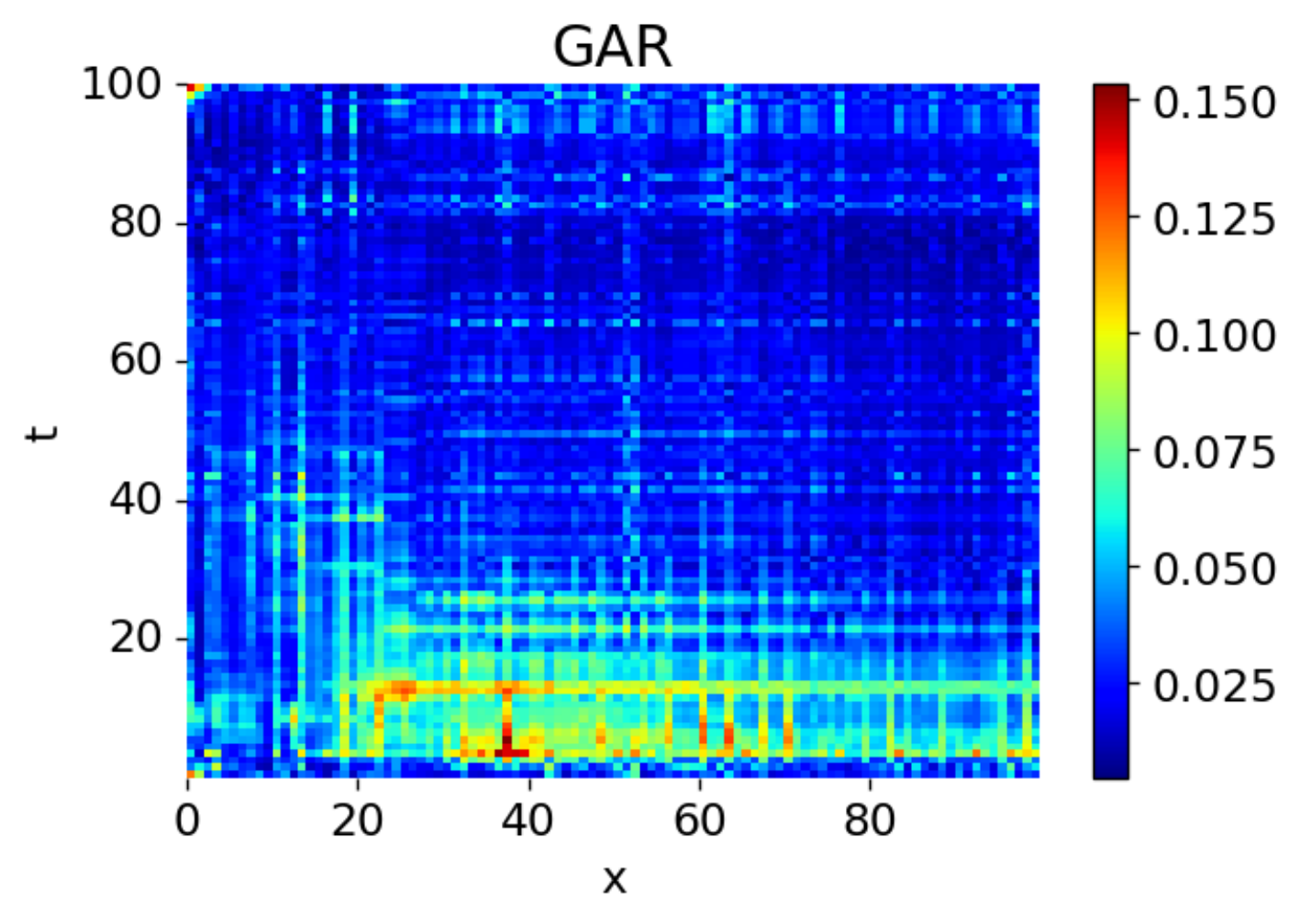}
    \end{minipage}%
    \hfill
    \begin{minipage}[t]{0.2\linewidth}
    \centering
    \includegraphics[width=3.25cm]{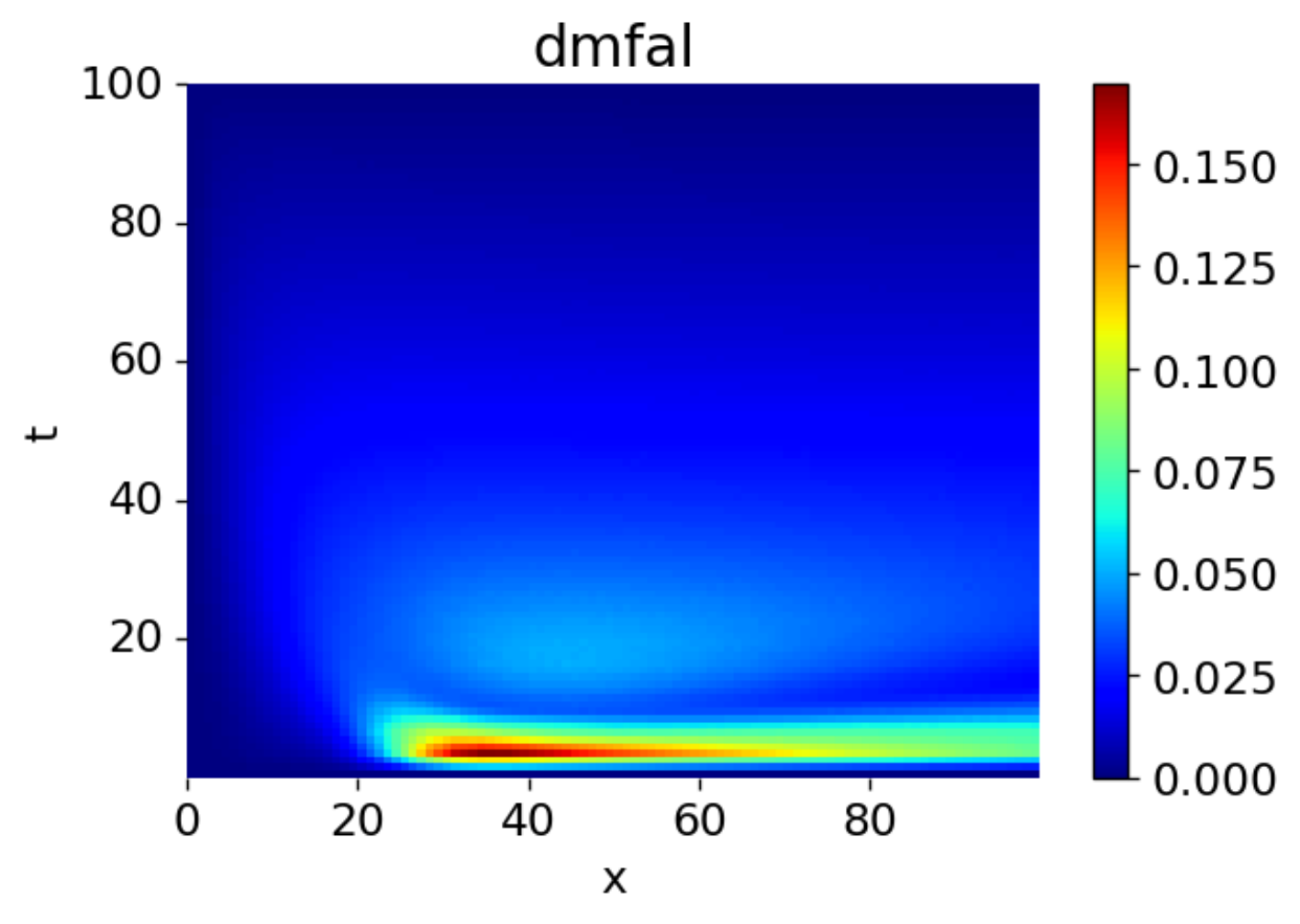}
    \end{minipage}%
    \hfill
    \begin{minipage}[t]{0.2\linewidth}
    \centering
    \includegraphics[width=3.25cm]{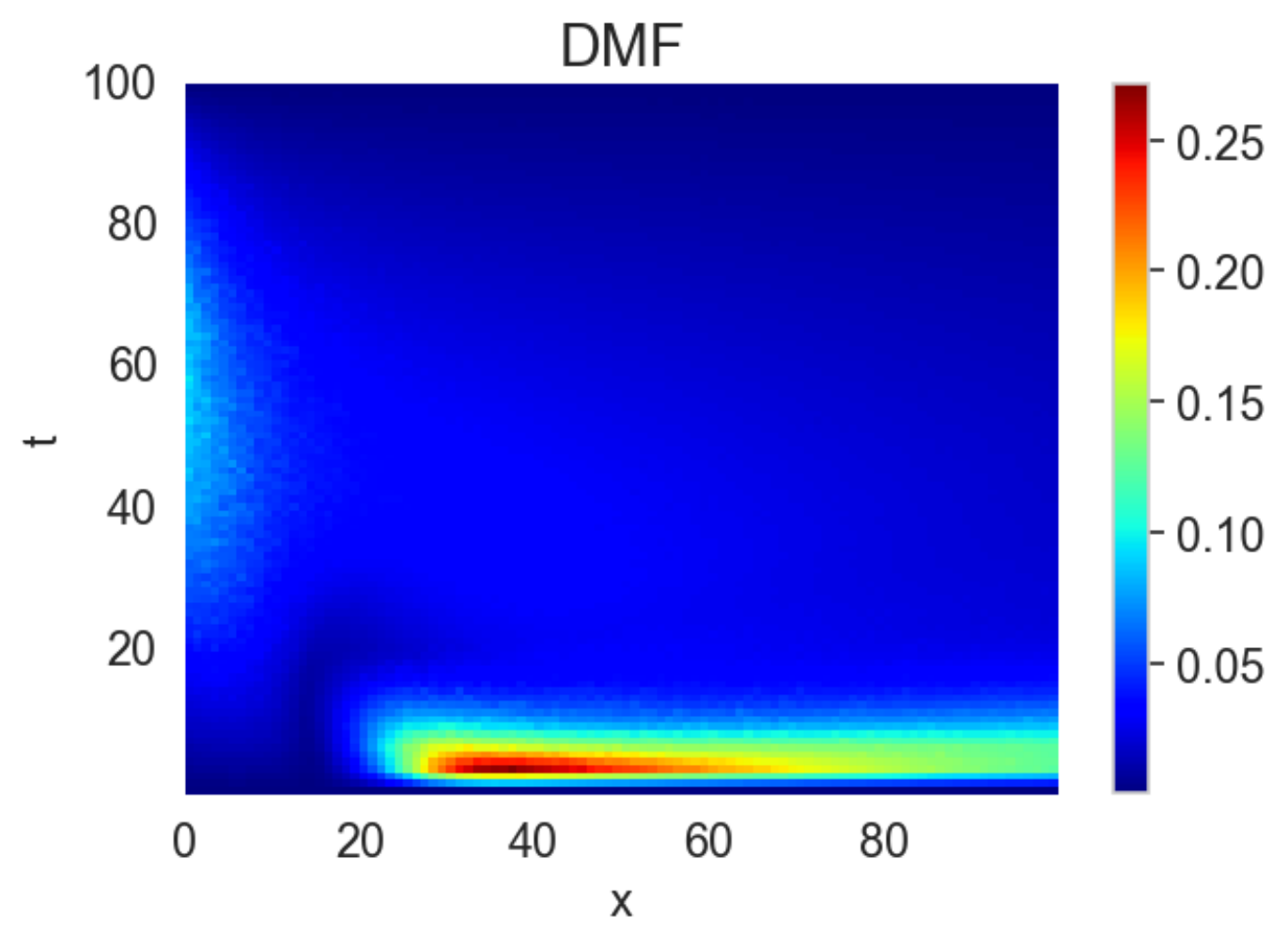}
    \end{minipage}%
    \hfill
    \\
    \begin{minipage}[t]{0.2\linewidth}
    \centering
    \includegraphics[width=3.25cm]{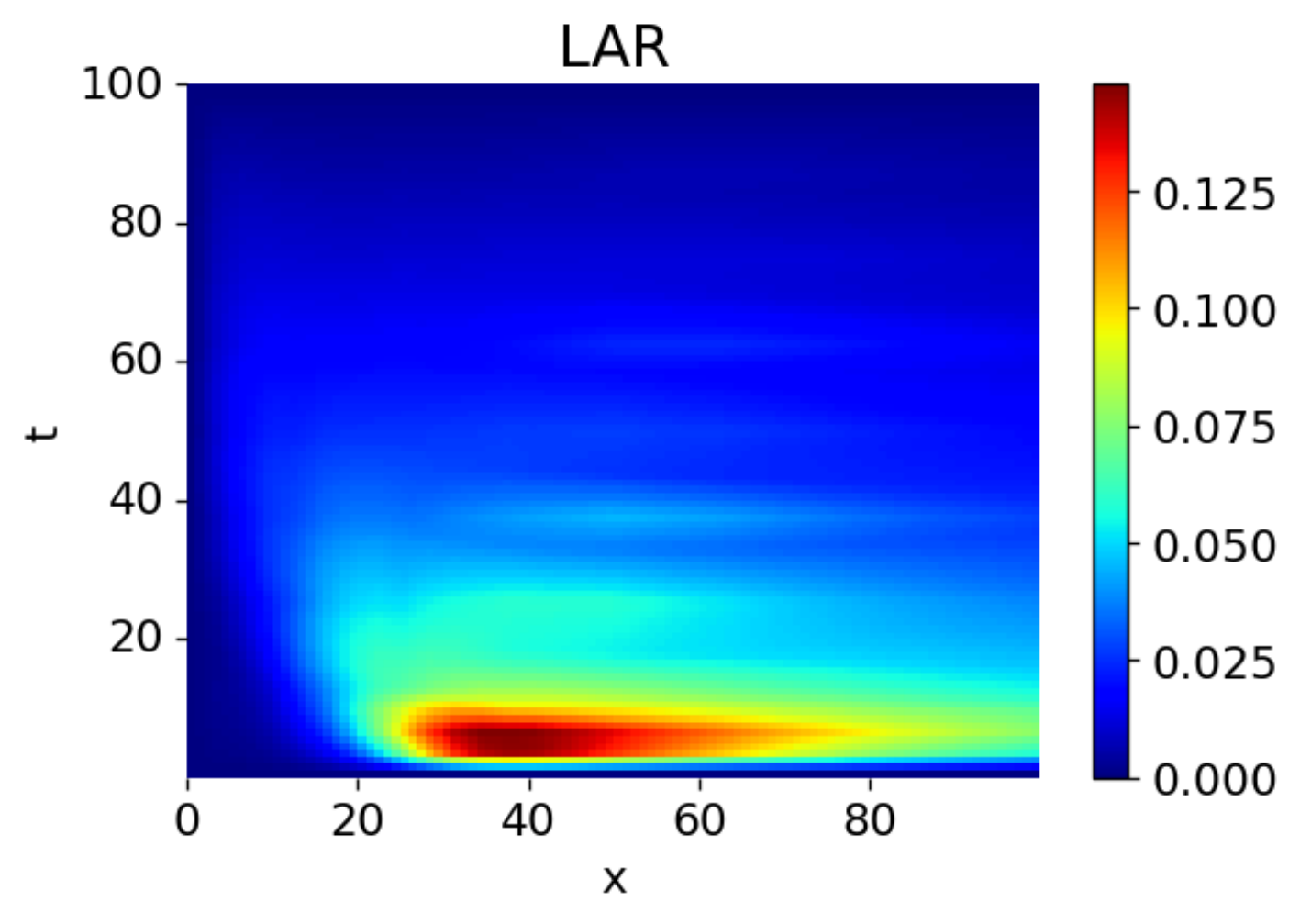}
    \end{minipage}%
    \hfill
    \begin{minipage}[t]{0.2\linewidth}
    \centering
    \includegraphics[width=3.25cm]{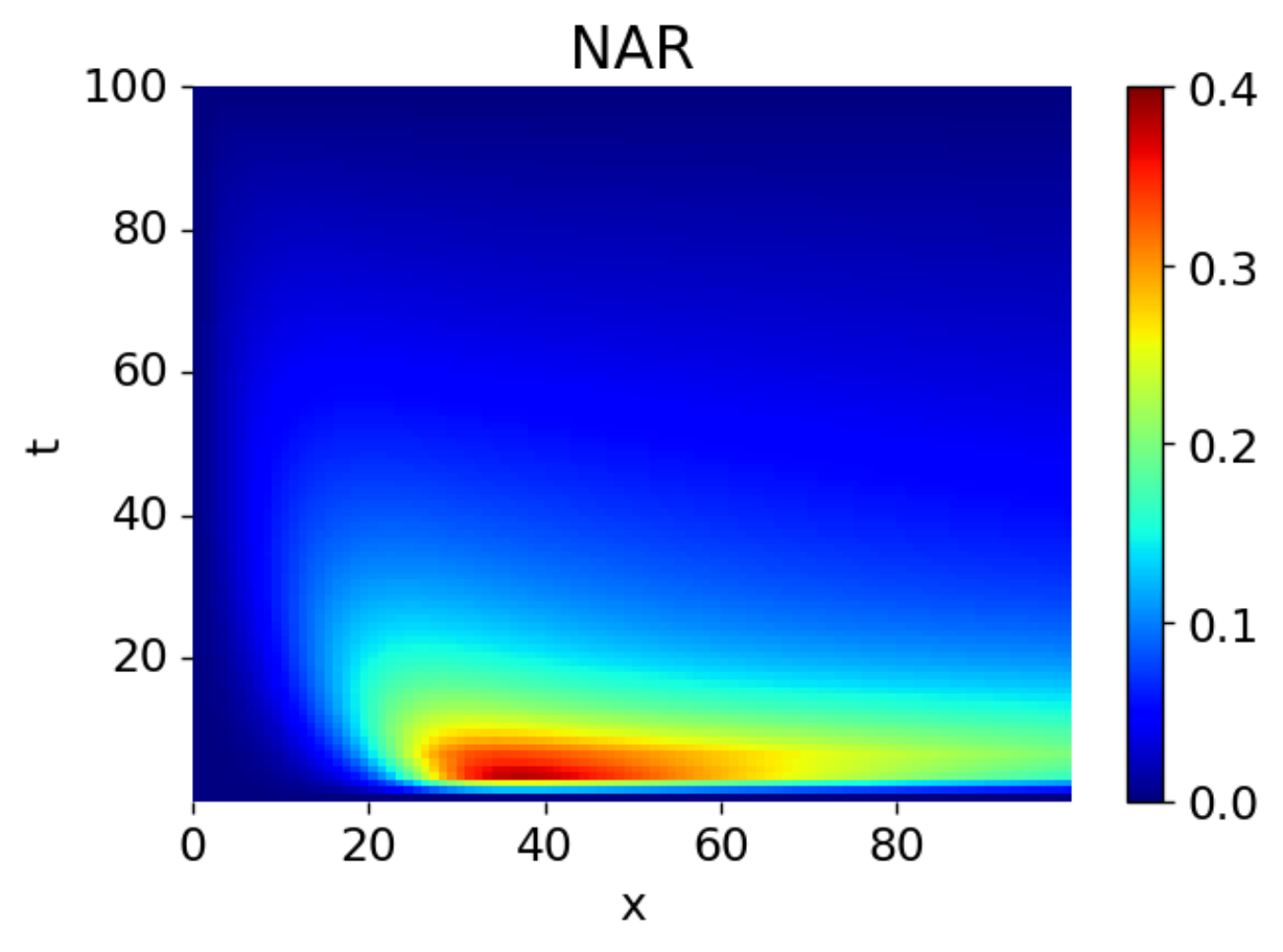}
    \end{minipage}%
    \begin{minipage}[t]{0.2\linewidth}
    \centering
    \includegraphics[width=3.25cm]{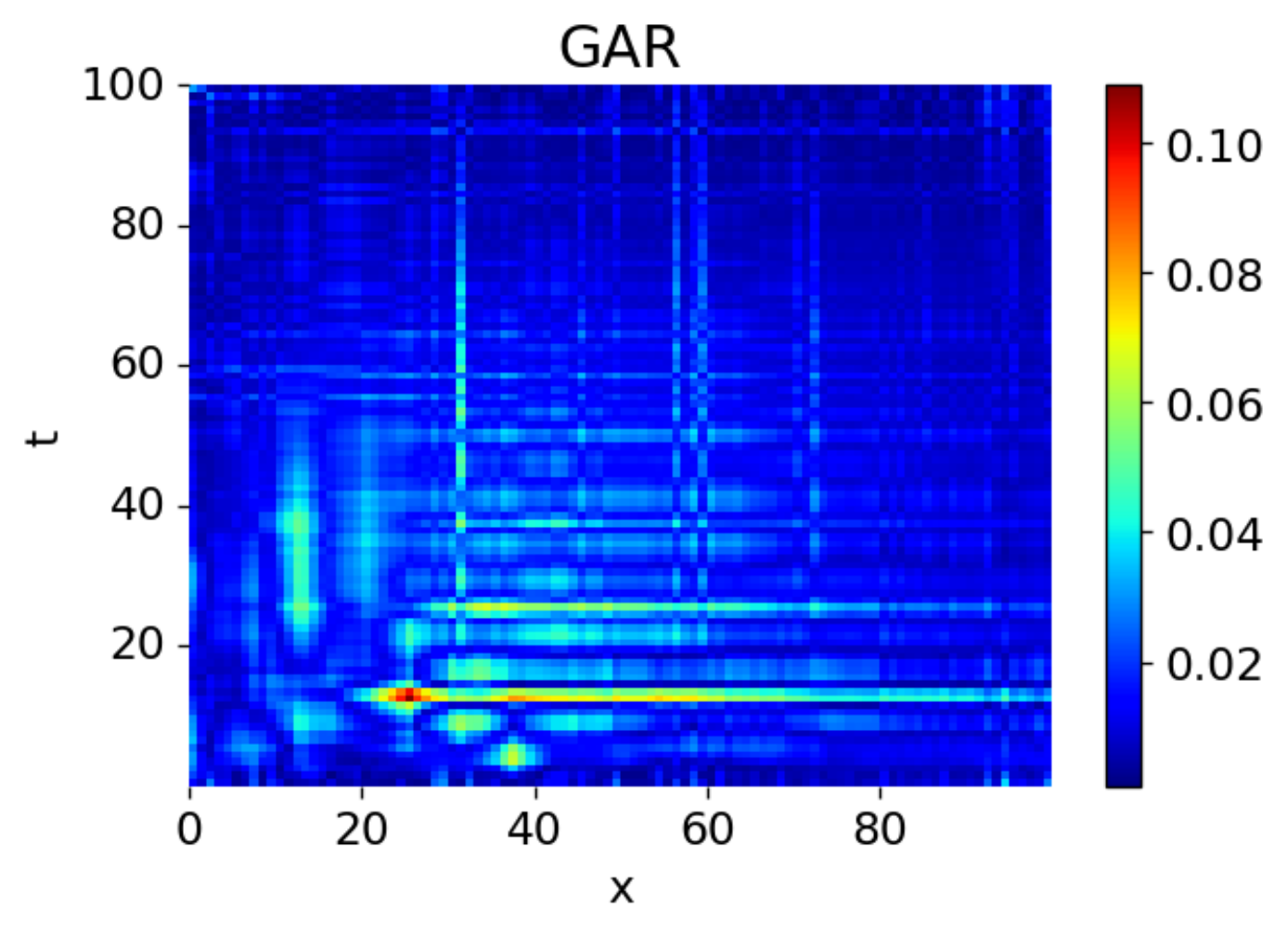}
    \end{minipage}%
    \hfill
    \begin{minipage}[t]{0.2\linewidth}
    \centering
    \includegraphics[width=3.25cm]{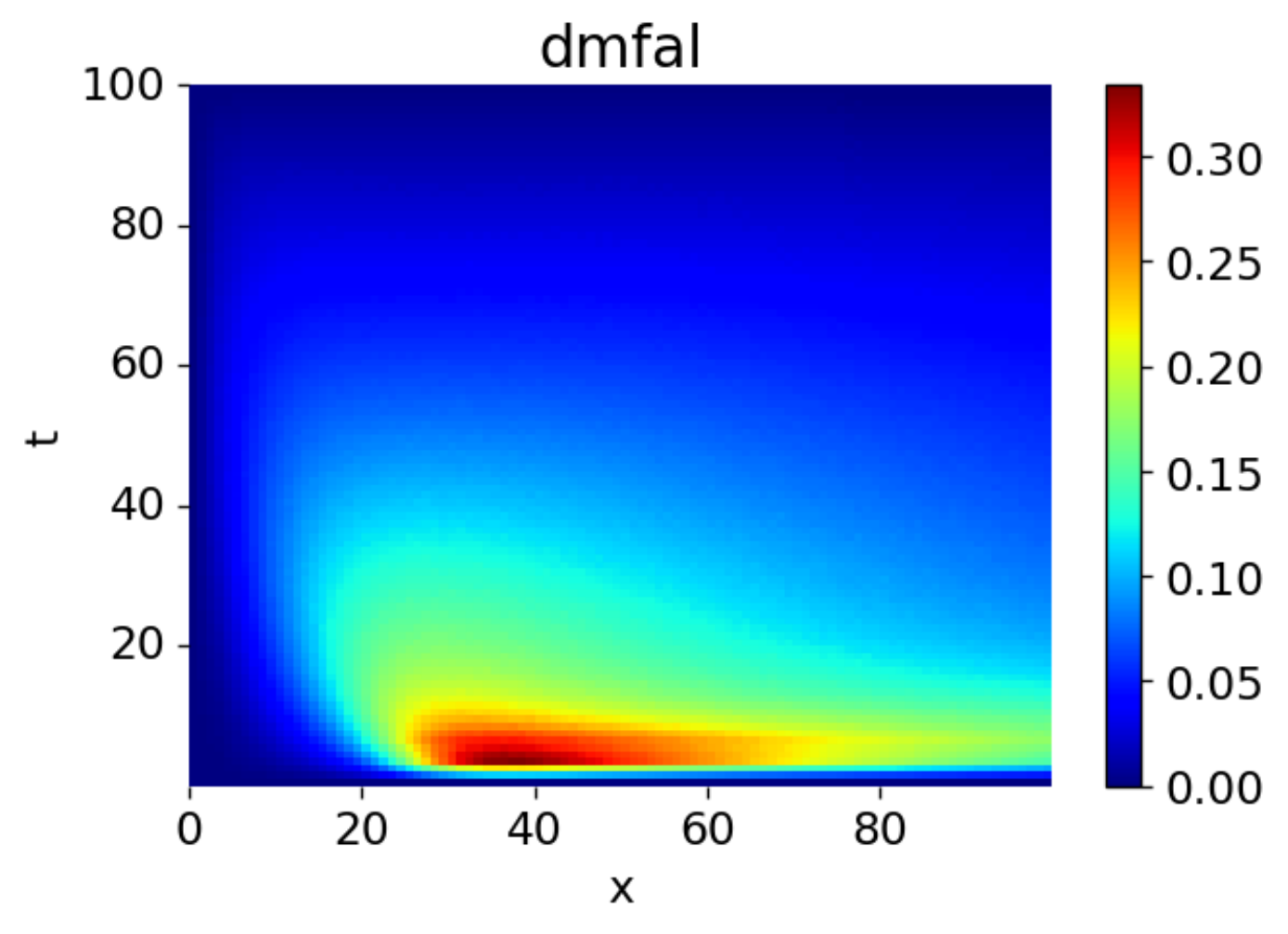}
    \end{minipage}%
    \hfill
    \begin{minipage}[t]{0.2\linewidth}
    \centering
    \includegraphics[width=3.25cm]{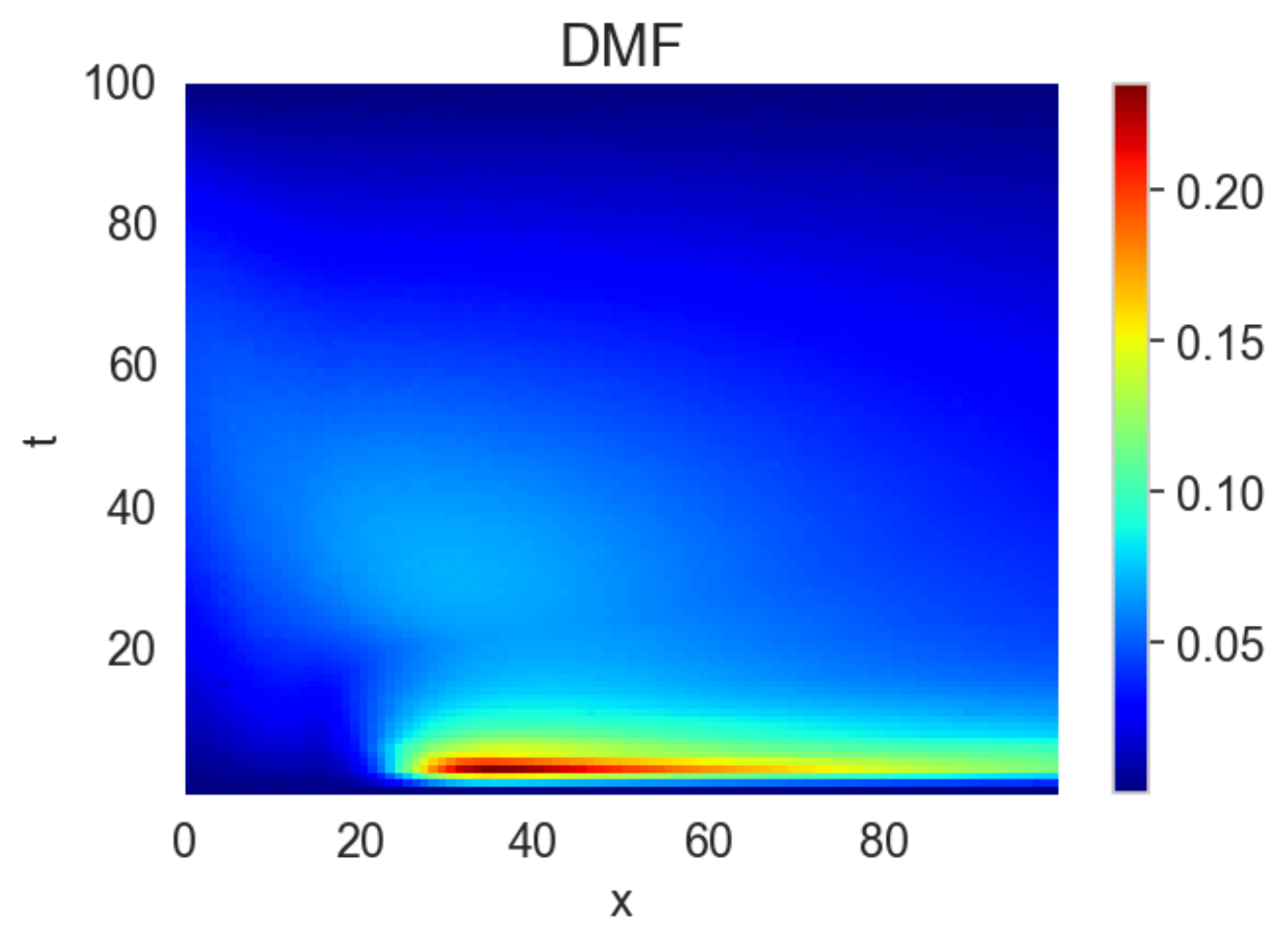}
    \end{minipage}%
    \hfill
    \\\begin{minipage}[t]{0.2\linewidth}
    \centering
    \includegraphics[width=3.25cm]{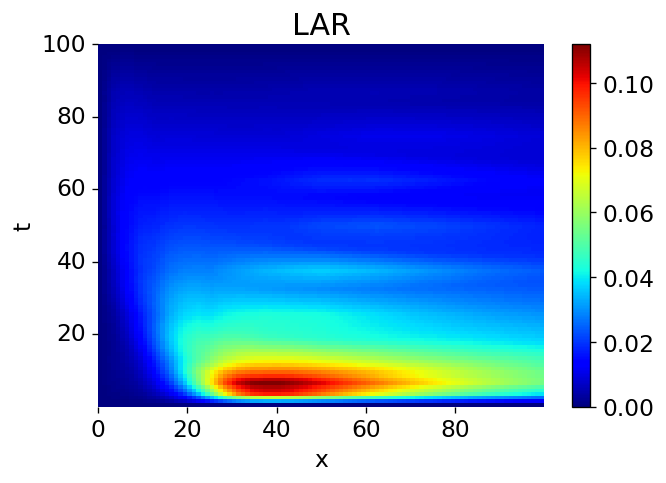}
    \end{minipage}%
    \hfill
    \begin{minipage}[t]{0.2\linewidth}
    \centering
    \includegraphics[width=3.25cm]{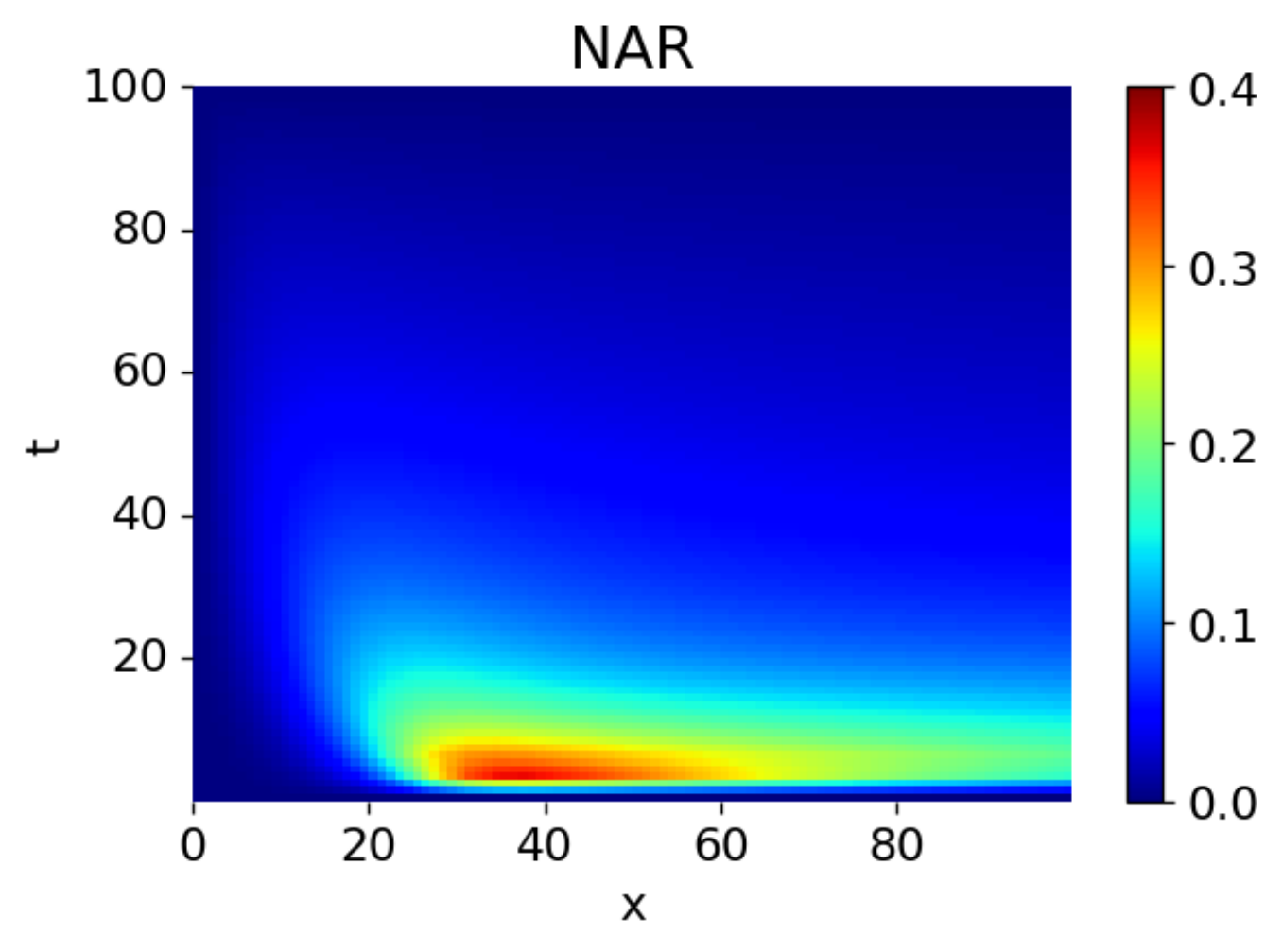}
    \end{minipage}%
    \begin{minipage}[t]{0.2\linewidth}
    \centering
    \includegraphics[width=3.25cm]{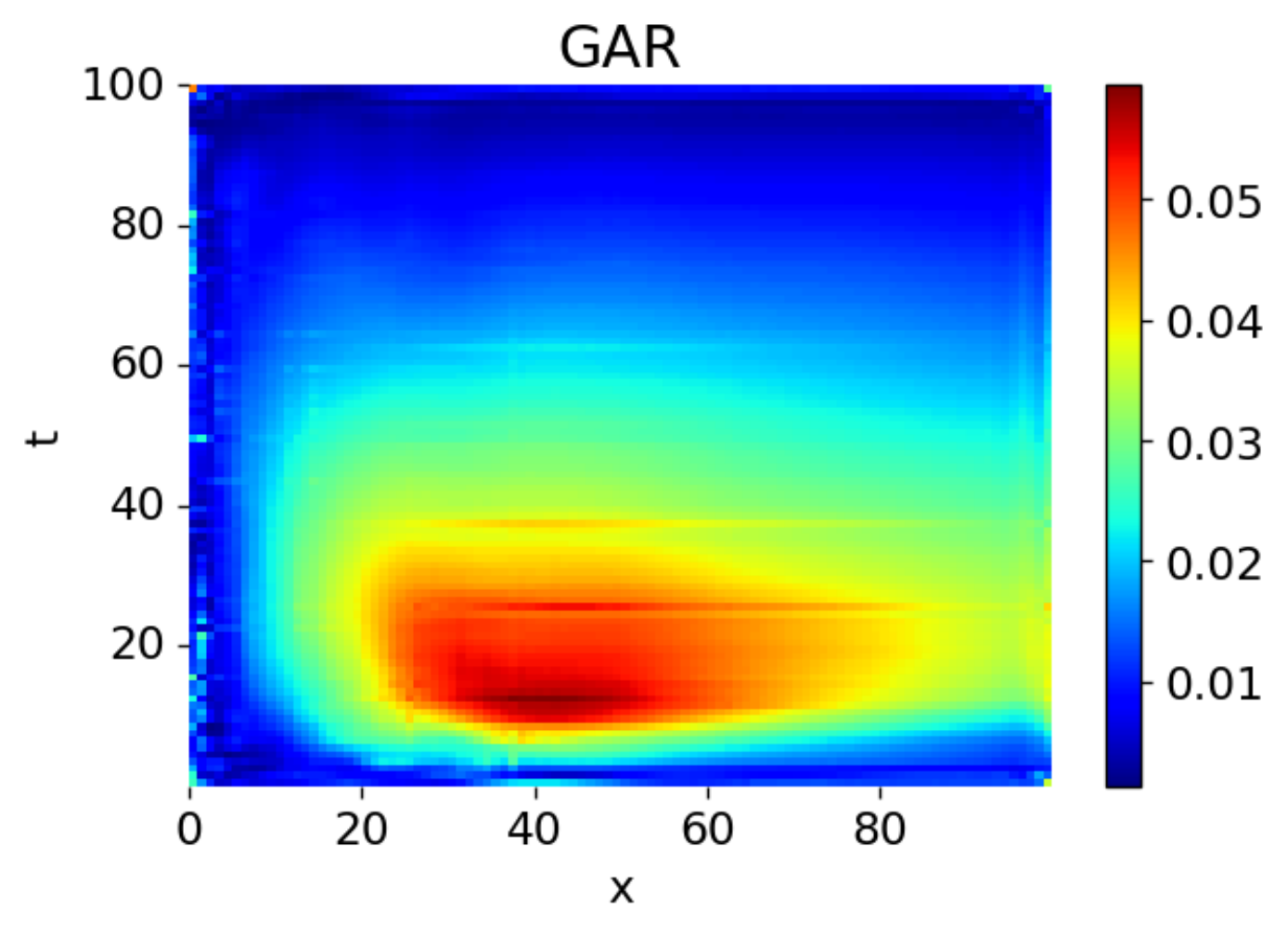}
    \end{minipage}%
    \hfill
    \begin{minipage}[t]{0.2\linewidth}
    \centering
    \includegraphics[width=3.25cm]{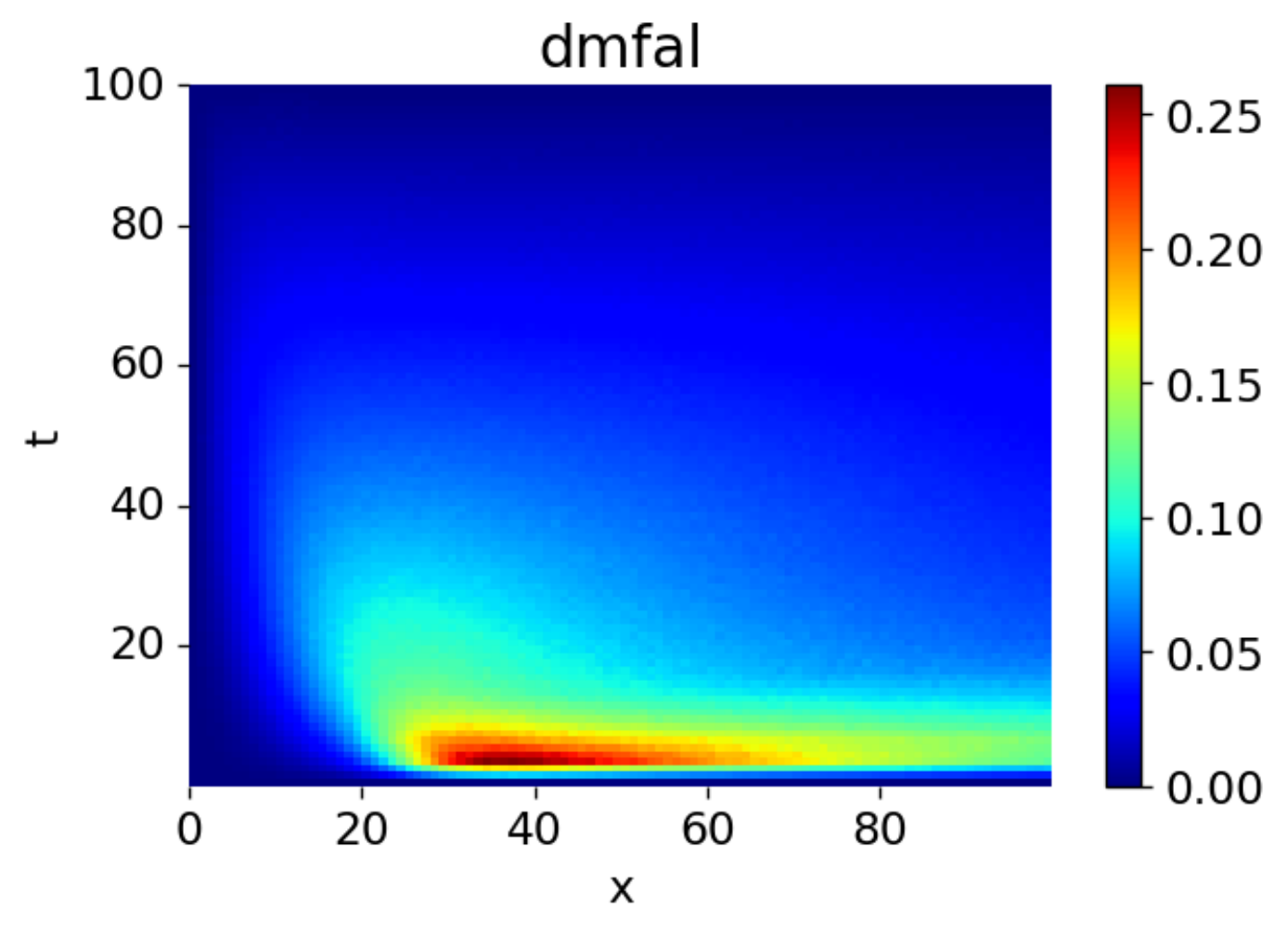}
    \end{minipage}%
    \hfill
    \begin{minipage}[t]{0.2\linewidth}
    \centering
    \includegraphics[width=3.25cm]{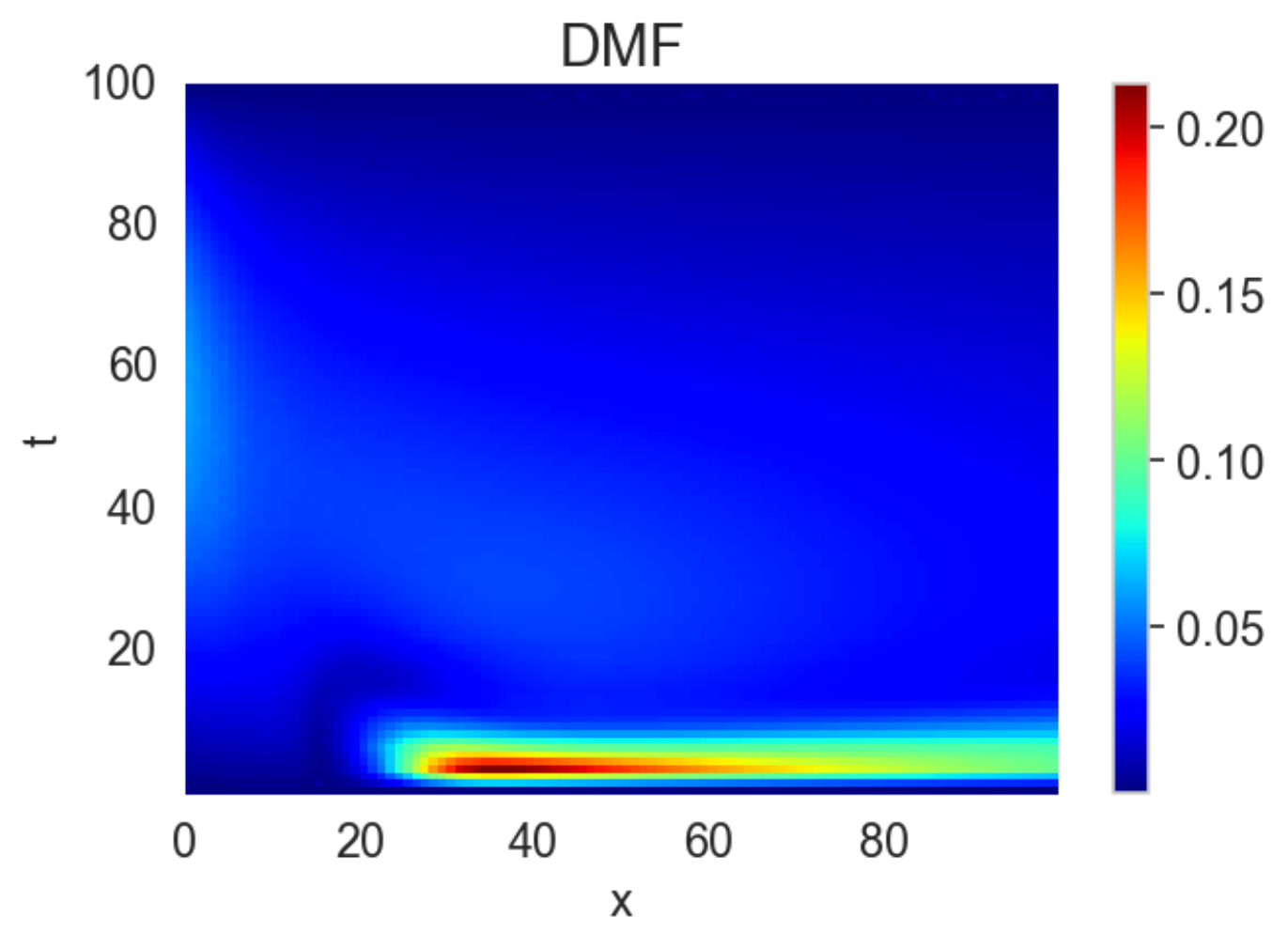}
    \end{minipage}%
    \hfill
    
    \caption{Comparison of MAE fields between different methods on Burger dataset, LAR, NAR, GAR, dmfal, DMF (left to right columns), using high-fidelity data $\{4,8,16,32\}$ (top to bottom rows) with fixed 32 low-fidelity data.}
    \label{exp_24}
\end{figure}

To show the detailed results. We use Fig.~\ref{exp_24} to show the pixel-wise mean (overall testing points) absolute error (MAE) field of $f_H(\x_H)$ and $\y_H$ on Burger's equation.
We can see that for all methods, the main error concentrates at the early stage of the system and reduces as the simulation time passes.
LAR and NAR show a consistently inferior accuracy compared to dmfal and DMF; GAR shows strong artifacts, which is consistent with the finding from the original work. Overall, DMF shows the best performance compared to the competitors, indicating its superiority over other methods in reconstructing the predictive fields consistently with different numbers of training points.


\noindent \textbf{Poisson's equation} is a typical elliptic PDE in mechanical engineering and physics for modelling potential fields, such as gravitational and electrostatic fields~\citep{chapra2011numerical}. It can be written as
    \begin{equation}
    \frac{\partial^{2} u}{\partial x^{2}}+\frac{\partial^{2} u}{\partial y^{2}}=0.
    \end{equation}
It is a generalization of Laplace's equation~\citep{persides1973laplace}.
    Despite its simplicity, Poisson's equation is commonly encountered in physics and is regularly used as a fundamental test case for surrogate models~\citep{tuo2014surrogate}. 
    In our experiment, we impose Dirichlet boundary conditions on a 2D spatial domain with $\textbf{x} \in [0,1] \times [0,1]$. The input parameters consist of the constant values of the four borders and the centre of the rectangular domain, which vary from $0.1$ to $0.9$ each. 
    To generate the matching potential fields as outputs, we sampled the input parameters equally. Using regular rectangular meshes and a first-order centre difference scheme, the PDE is solved using the finite difference method. We utilized an $8\times8$ mesh for the coarsest-level solution and $32\times32$ for the finest mesh used by the improved solver. 
    {The outputs are also upscaled using interpolation to $100\times 100$ nodes, and the experiment settings keep the same with Burger. 
    The results are shown in Fig.~\ref{exp22}.
    As can be seen, DMF performs worse than LAR and GAR when the high-fidelity data are limited, \ie 4 or 8 samples. 
    However, the RMSE of DMF keeps dropping with the increasing of high-fidelity samples while the performances of the latter two methods do not change considerably, which is consistent with the conclusion drawn from the Burger's equation. 
    As for other baselines, DC cigp, NAR, and dmfal are surpassed by DMF for all experiment settings. The detailed pixel-wise MAE comparisons are shown in Fig.~\ref{exp_23}.
    As the figures show,  the error of DMF concentrates in the centre area first,
    and then reduces gradually both in area and magnitude; 
    NAR and dmfal perform worse than DMF obviously; LAR and dmfal have a smaller variance when high-fidelity data are rare but their RMSEs fail to decrease with the increment of high-fidelity data in the corner area. 
    In these two experiment, we can see that DMF has a significant advantage over the SOTA methods when high-fidelity data are sufficient, which gives it a great potential in practical applications where accuracy is the most important.
   
   \begin{figure}[htbp]
    \begin{minipage}[t]{0.2\linewidth}
    \centering
    \includegraphics[width=3.25cm]{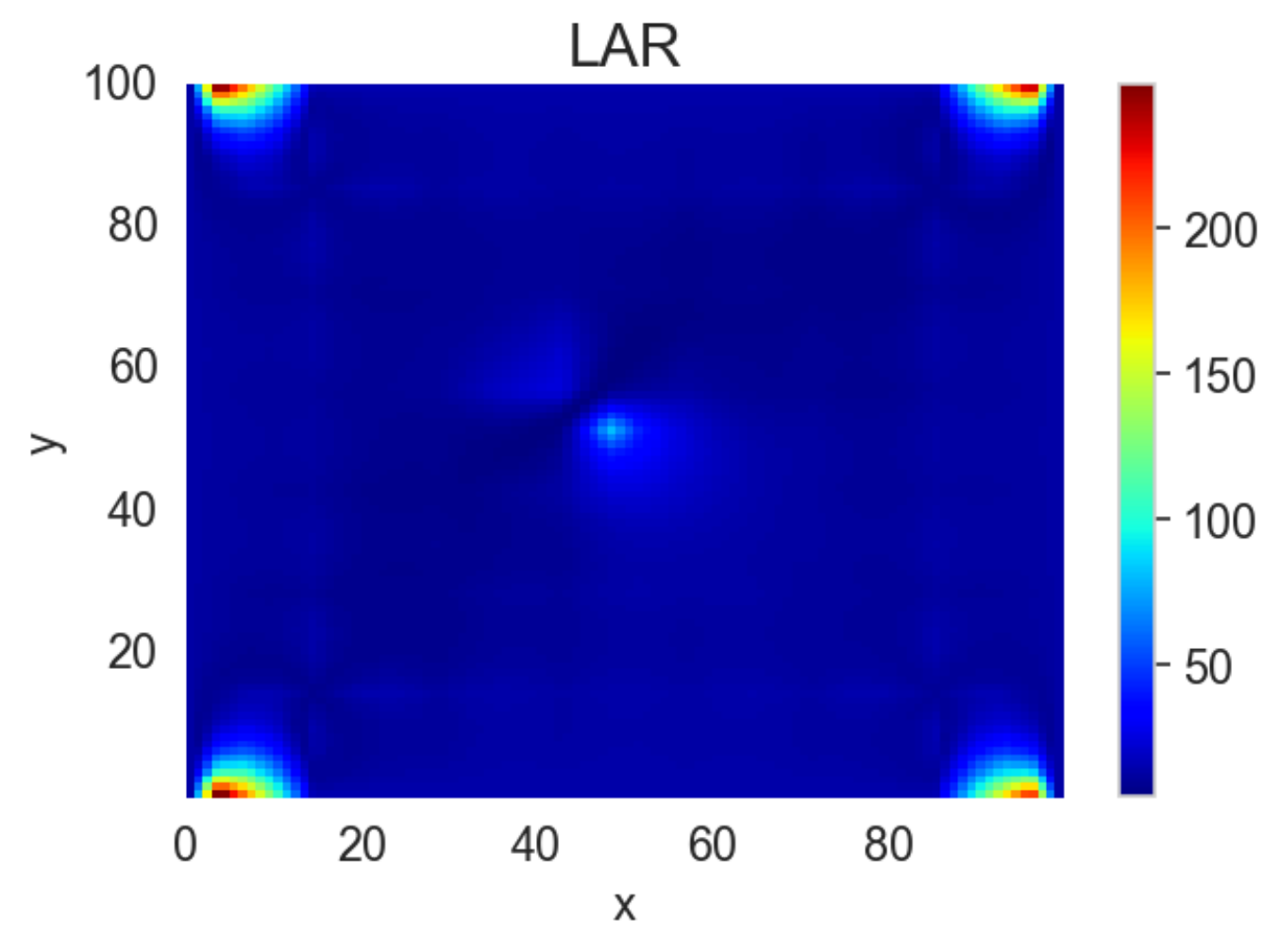}
    \end{minipage}%
    \hfill
    \begin{minipage}[t]{0.2\linewidth}
    \centering
    \includegraphics[width=3.25cm]{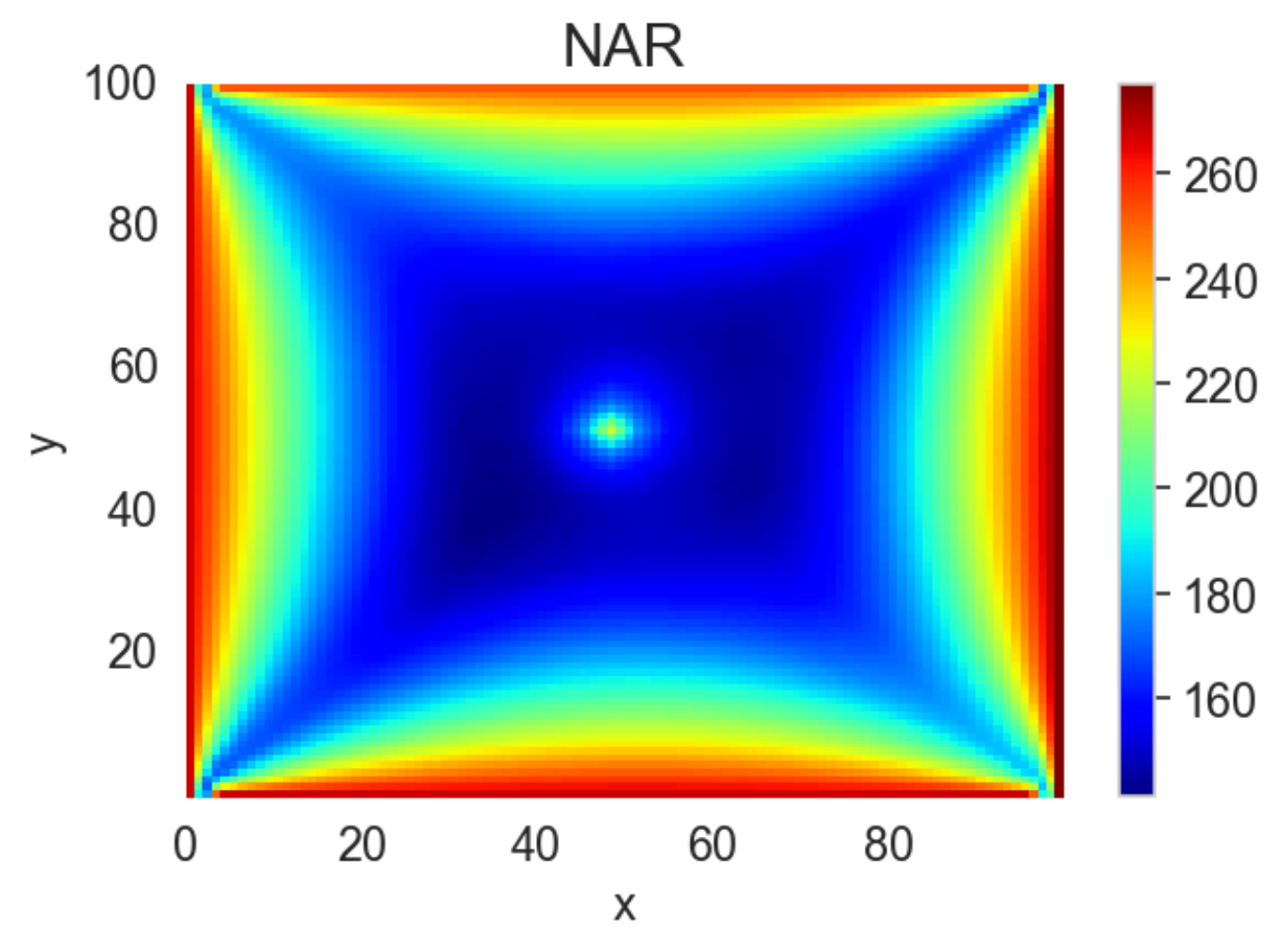}
    \end{minipage}%
    \begin{minipage}[t]{0.2\linewidth}
    \centering
    \includegraphics[width=3.25cm]{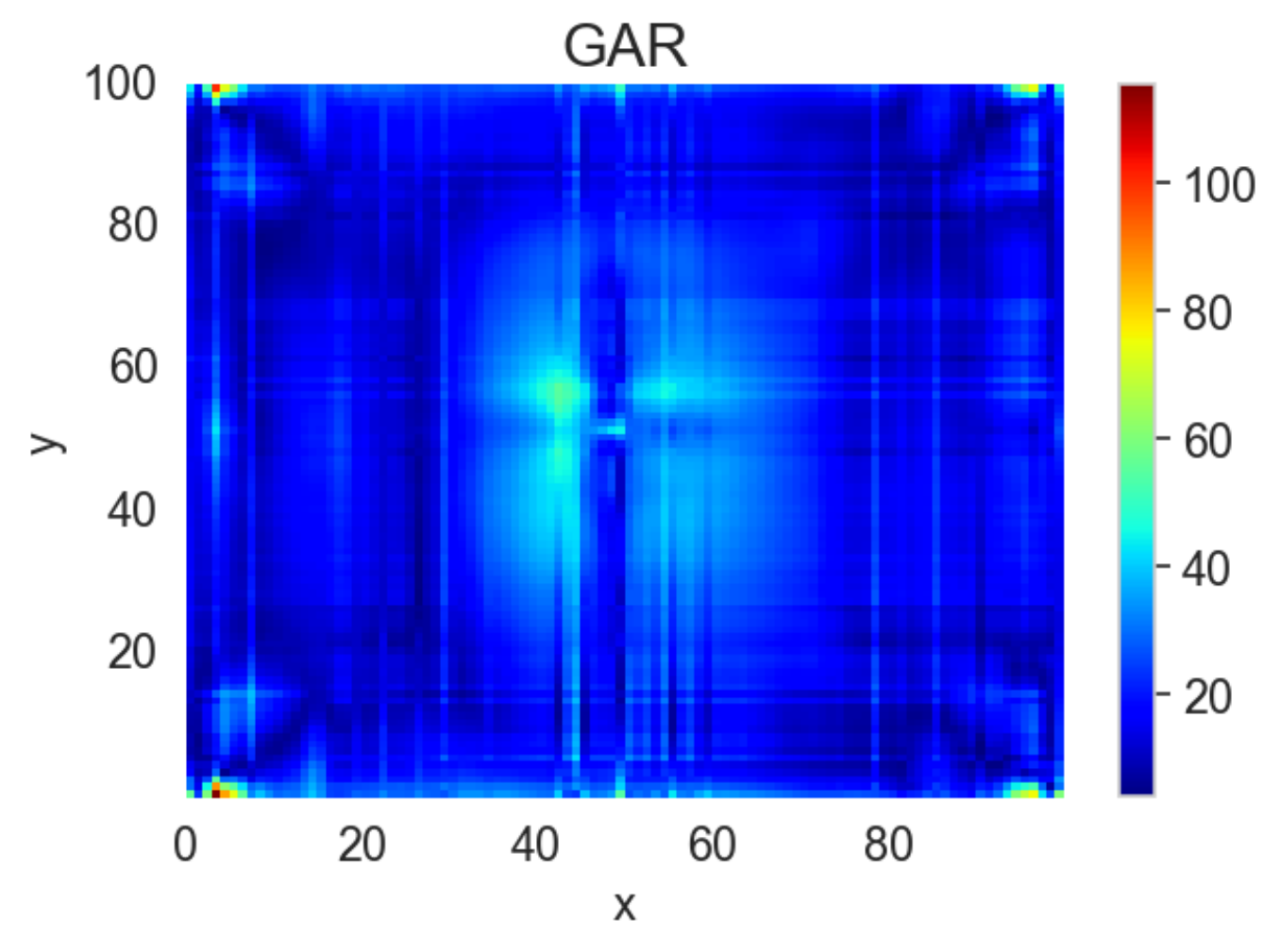}
    \end{minipage}%
    \hfill
    \begin{minipage}[t]{0.2\linewidth}
    \centering
    \includegraphics[width=3.25cm]{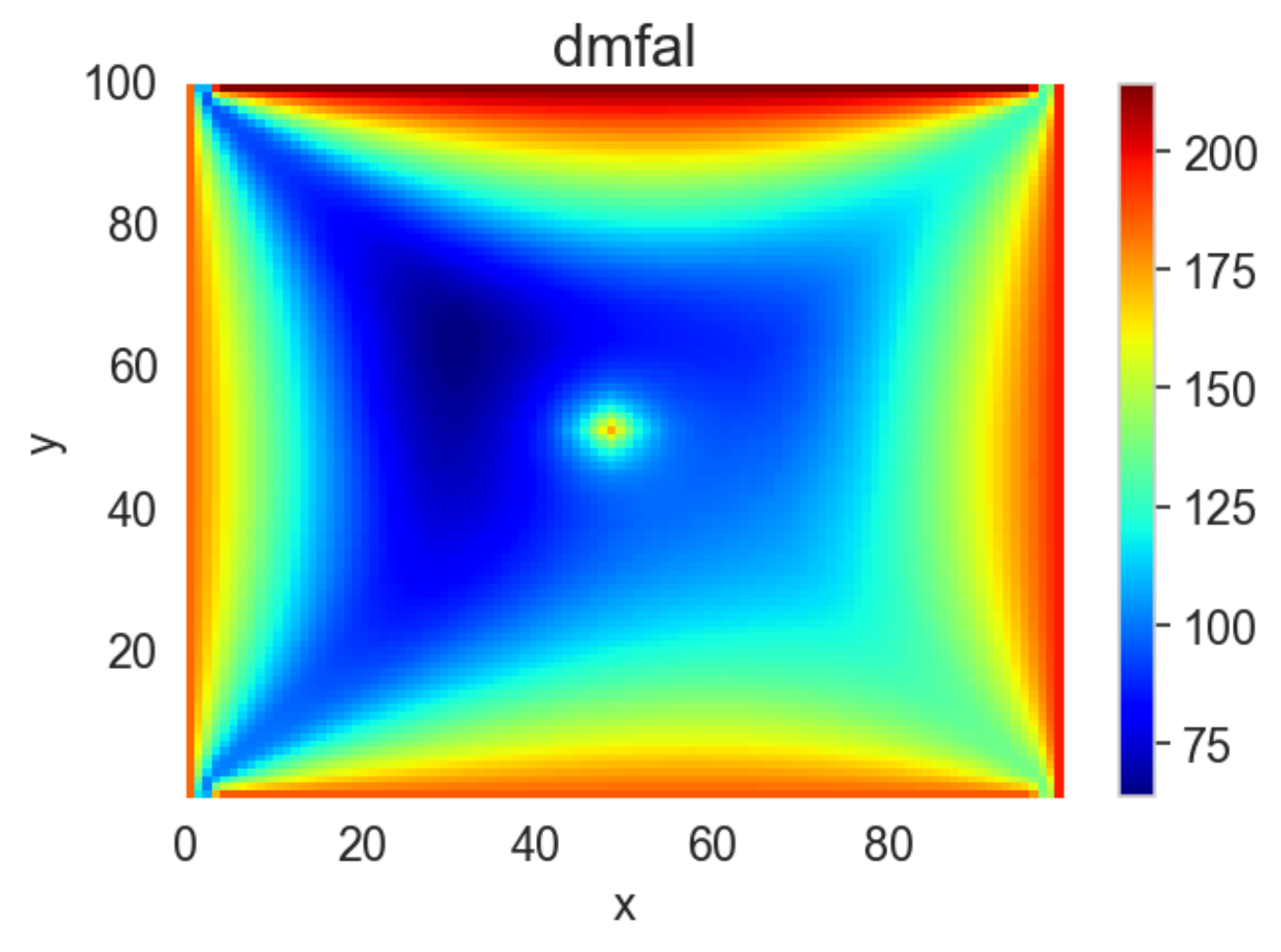}
    \end{minipage}%
    \hfill
    \begin{minipage}[t]{0.2\linewidth}
    \centering
    \includegraphics[width=3.25cm]{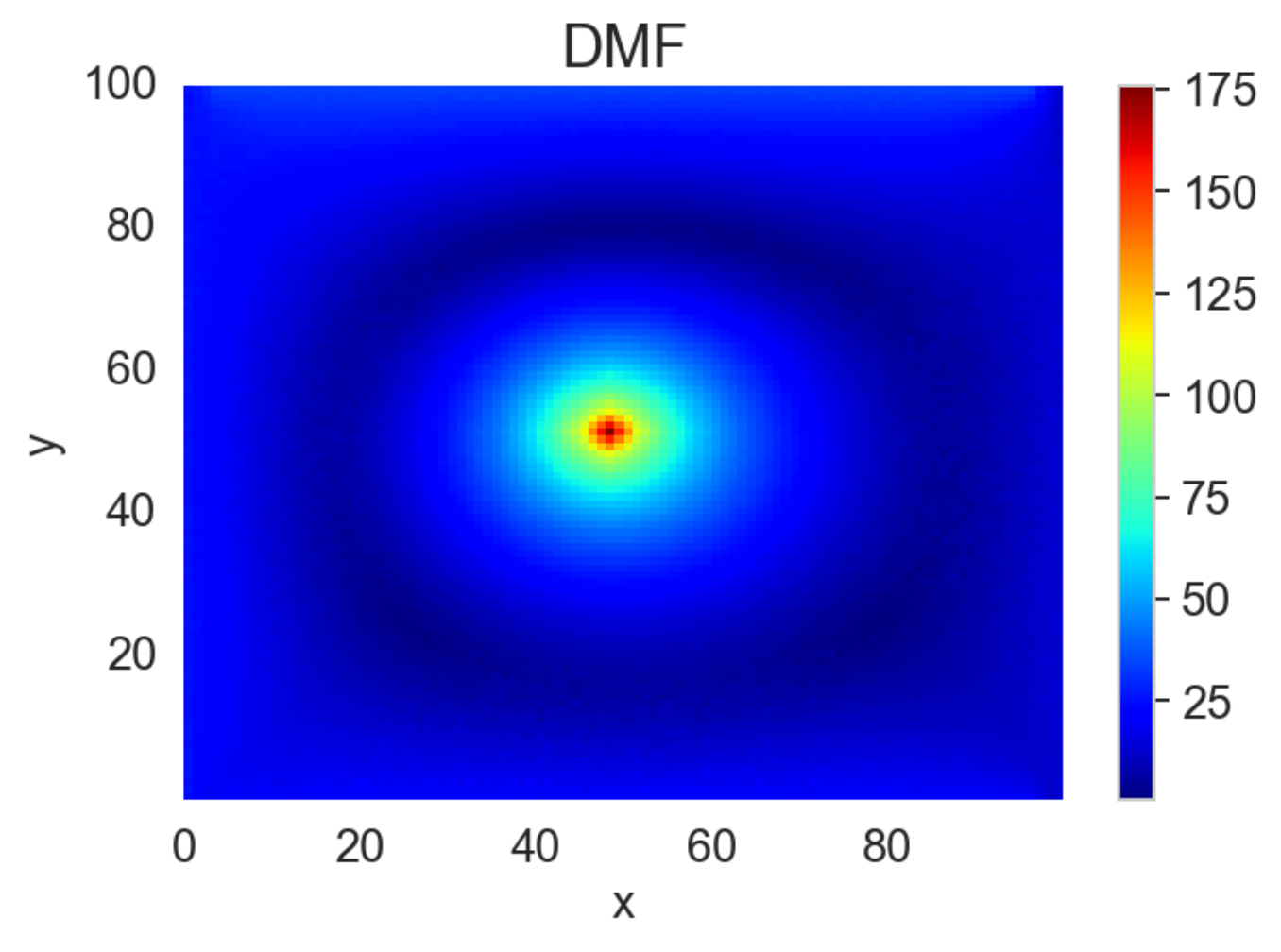}
    \end{minipage}%
    \hfill
    \\
    \begin{minipage}[t]{0.2\linewidth}
    \centering
    \includegraphics[width=3.25cm]{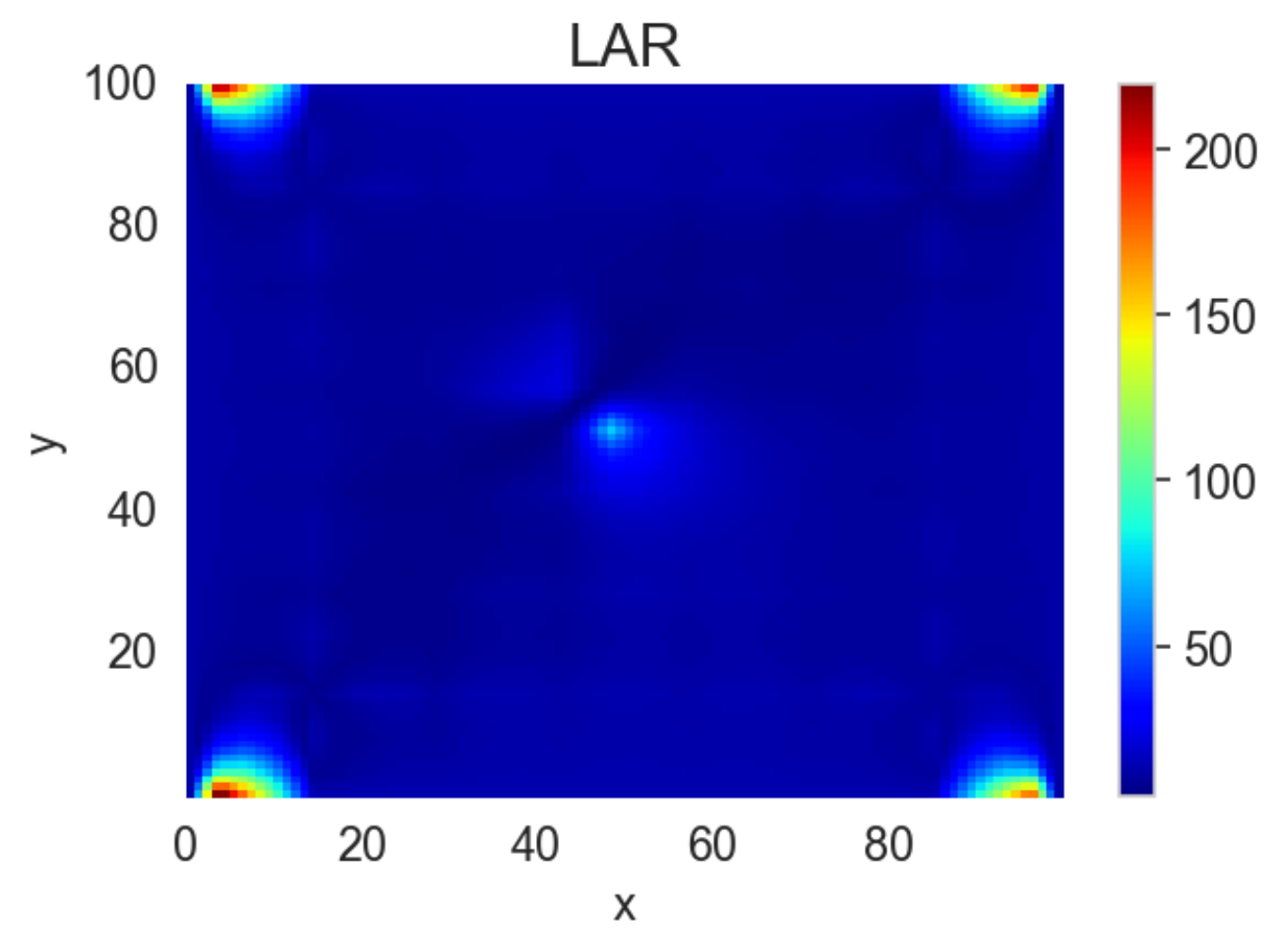}
    \end{minipage}%
    \hfill
    \begin{minipage}[t]{0.2\linewidth}
    \centering
    \includegraphics[width=3.25cm]{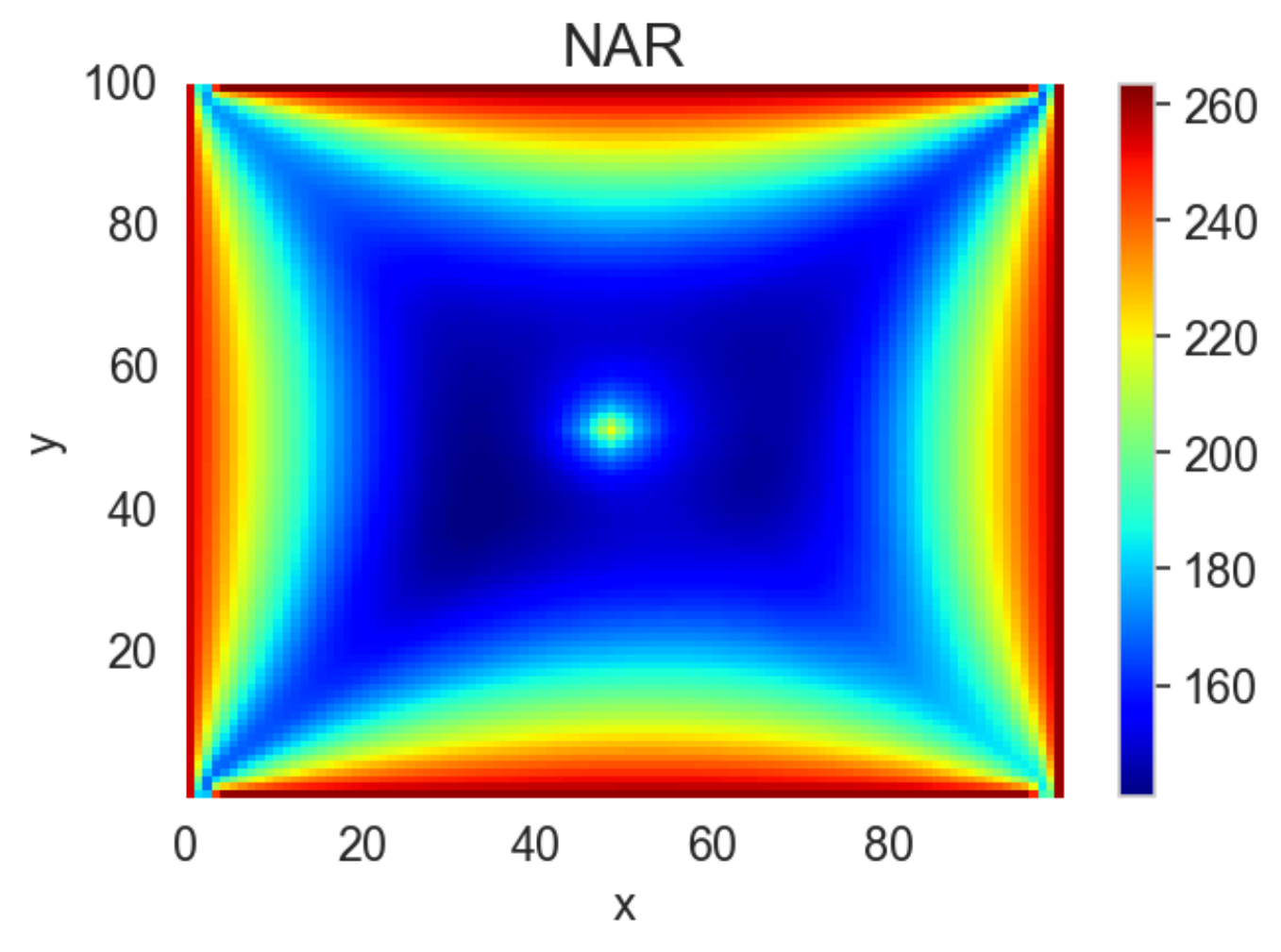}
    \end{minipage}%
    \begin{minipage}[t]{0.2\linewidth}
    \centering
    \includegraphics[width=3.25cm]{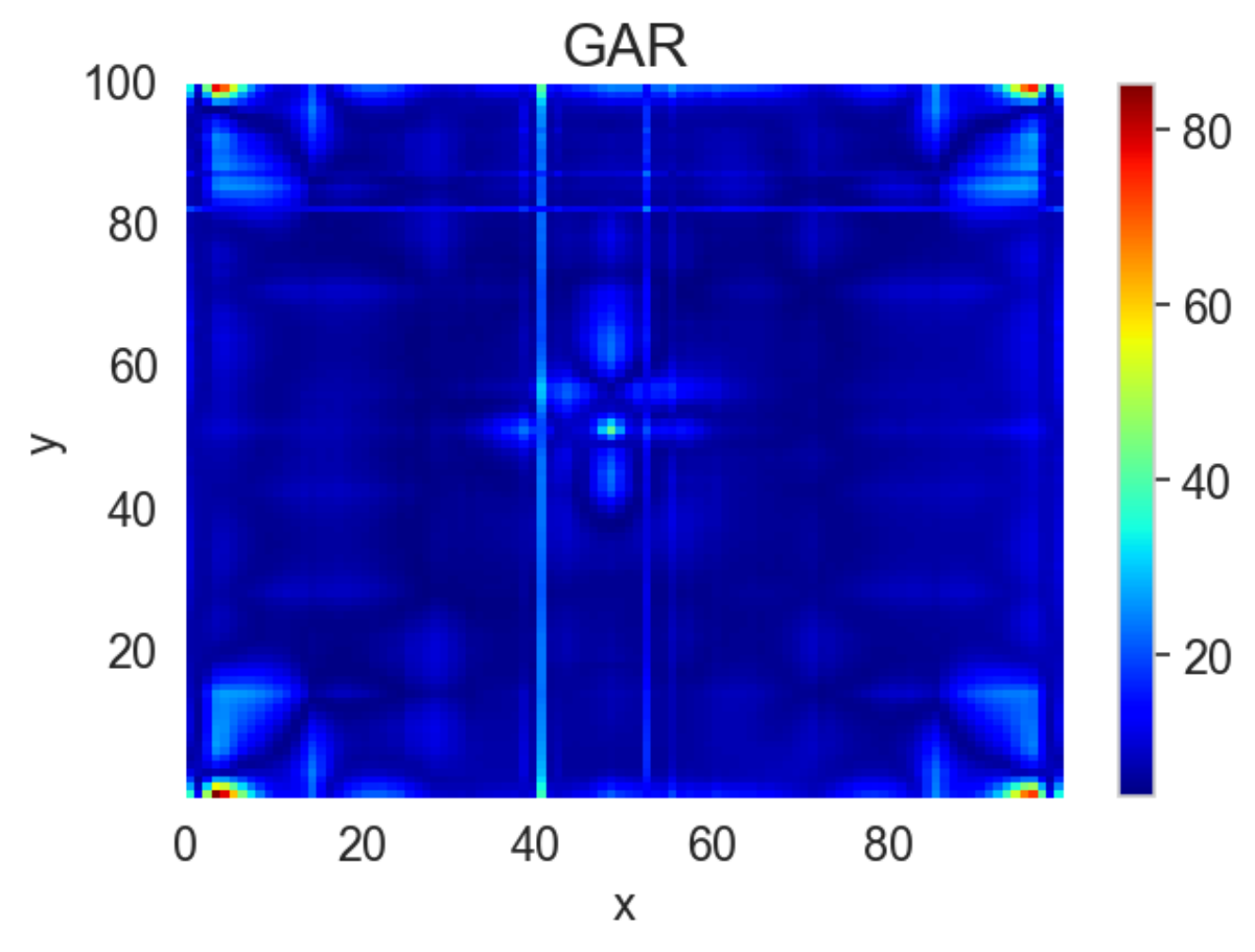}
    \end{minipage}%
    \hfill
    \begin{minipage}[t]{0.2\linewidth}
    \centering
    \includegraphics[width=3.25cm]{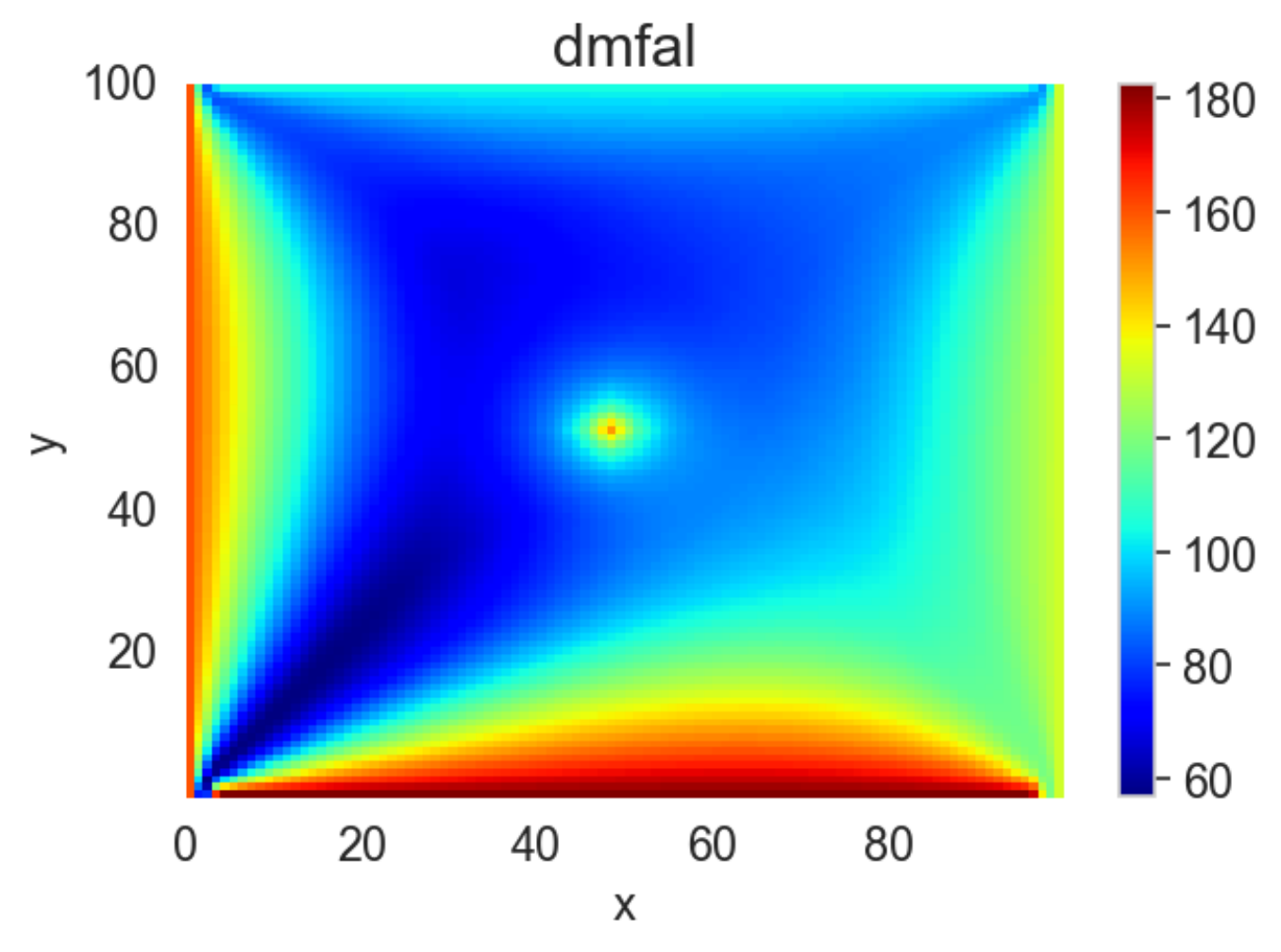}
    \end{minipage}%
    \hfill
    \begin{minipage}[t]{0.2\linewidth}
    \centering
    \includegraphics[width=3.25cm]{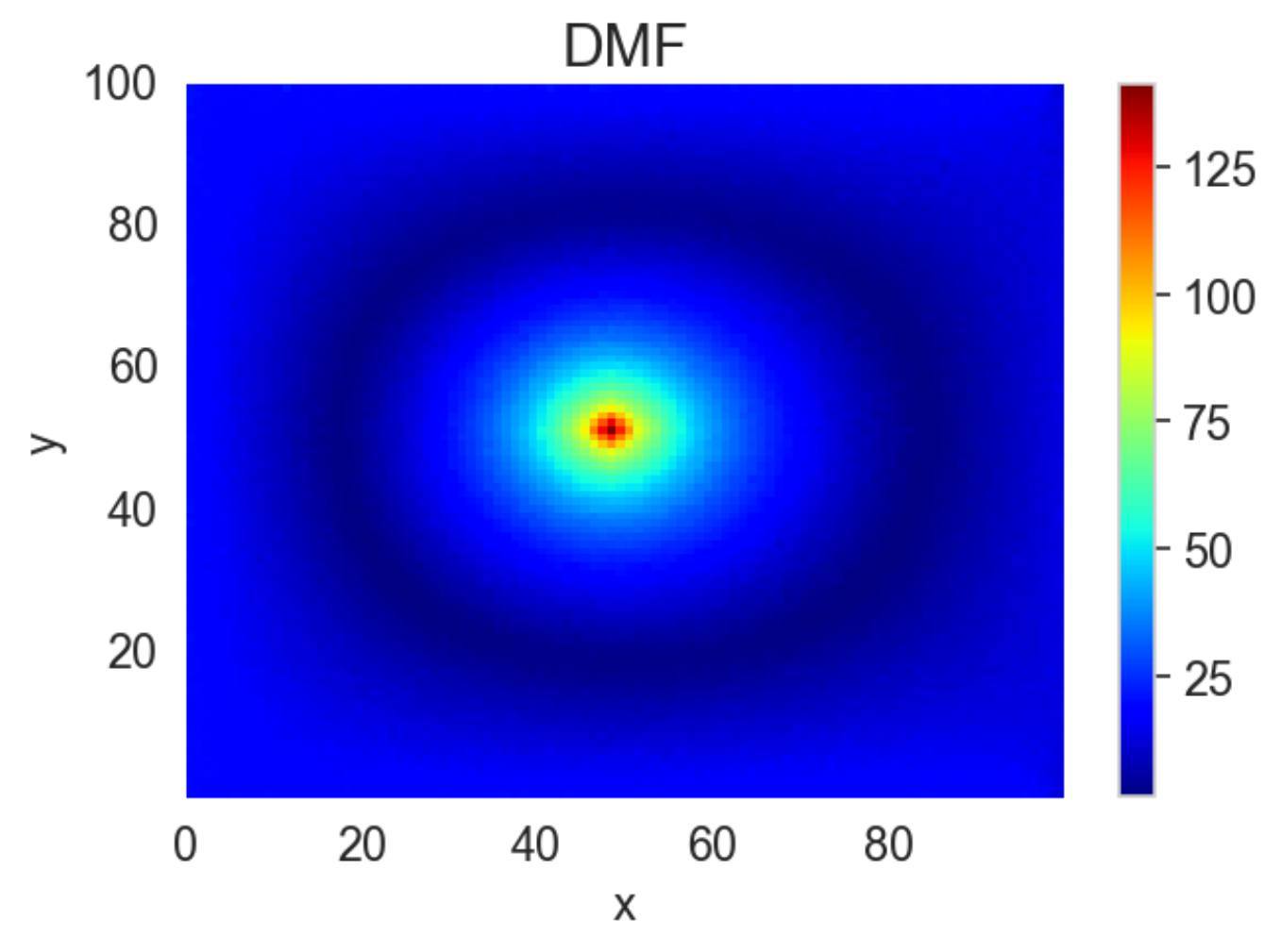}
    \end{minipage}%
    \hfill
    \\
    \begin{minipage}[t]{0.2\linewidth}
    \centering
    \includegraphics[width=3.25cm]{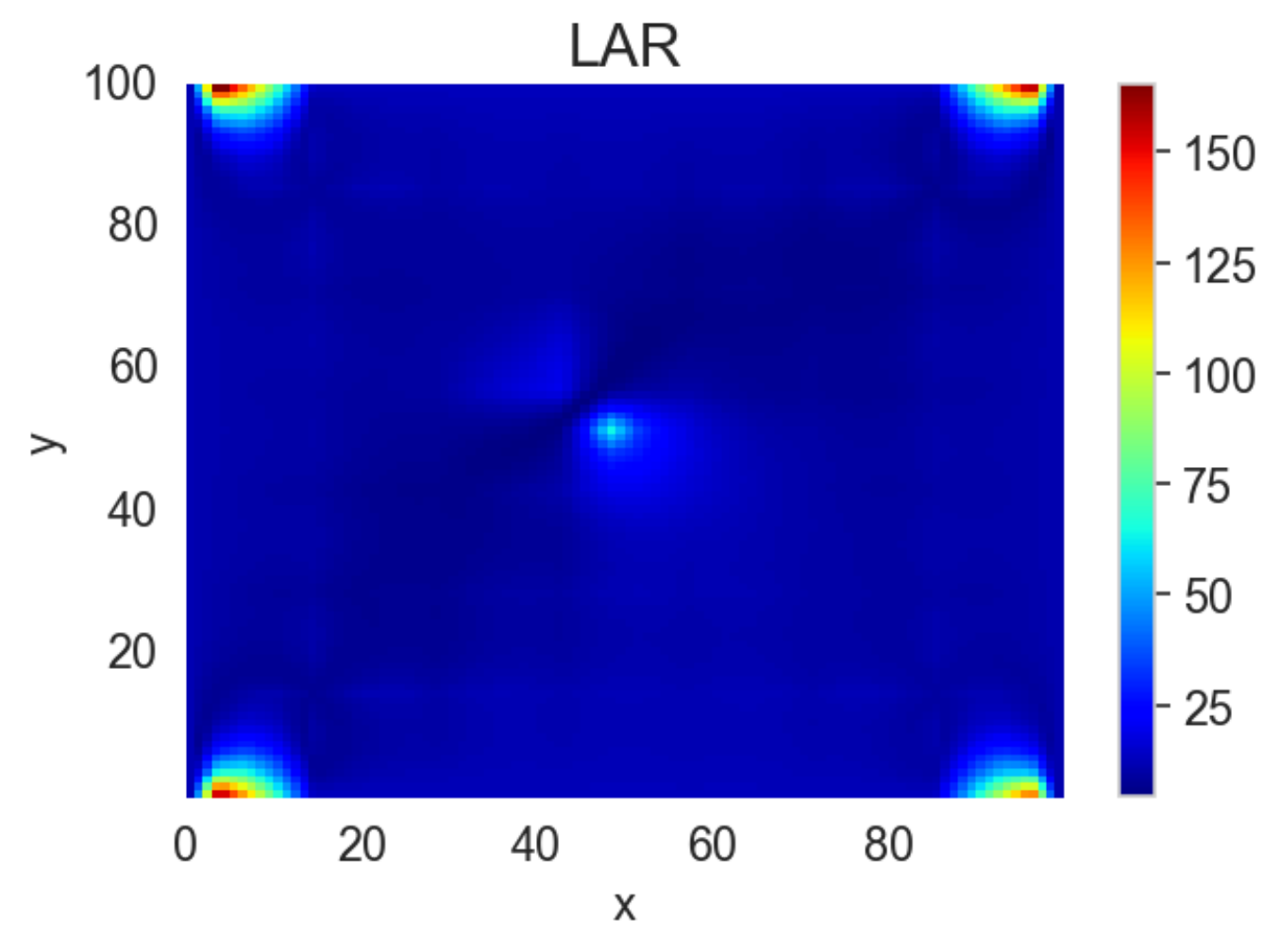}
    \end{minipage}%
    \hfill
    \begin{minipage}[t]{0.2\linewidth}
    \centering
    \includegraphics[width=3.25cm]{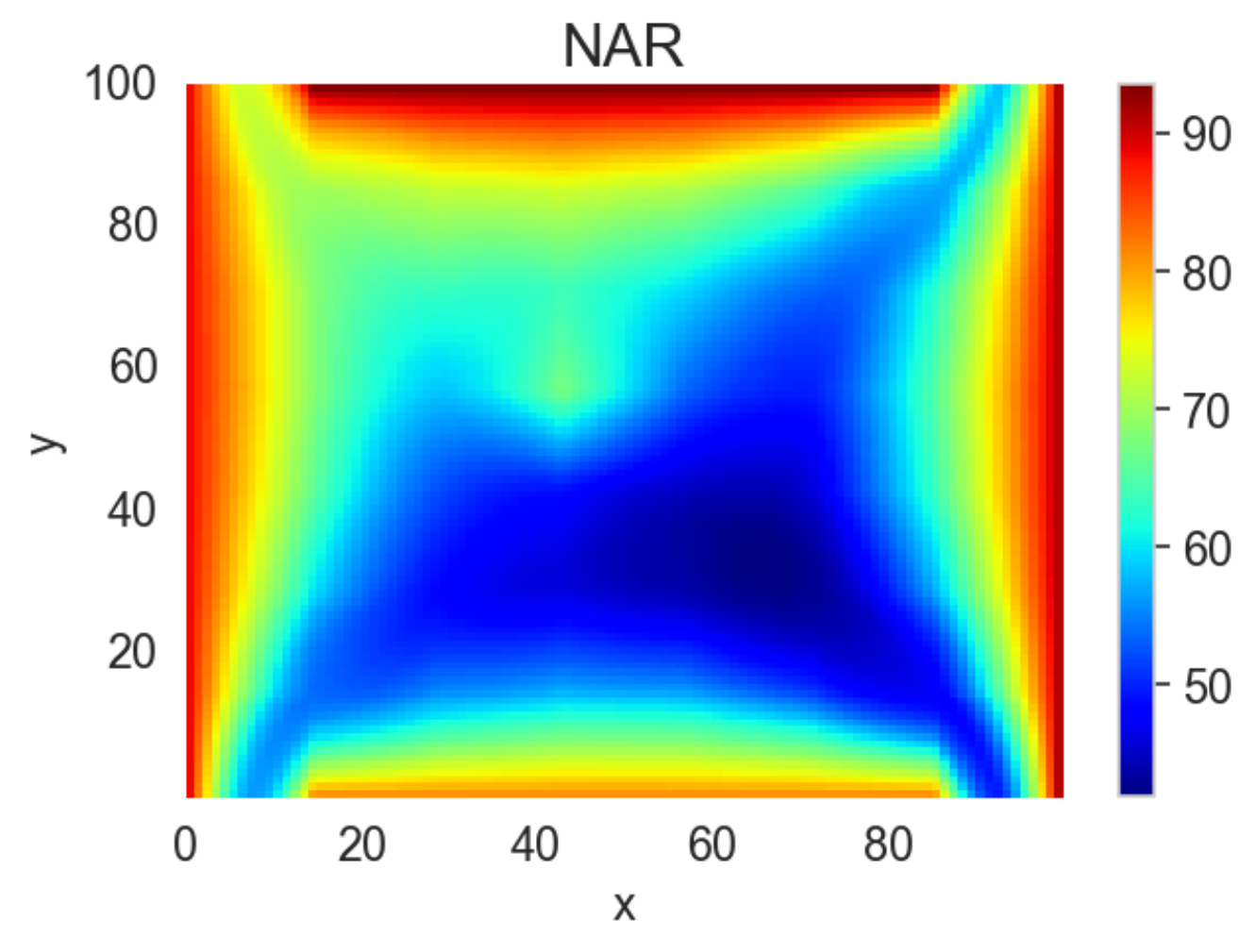}
    \end{minipage}%
    \begin{minipage}[t]{0.2\linewidth}
    \centering
    \includegraphics[width=3.25cm]{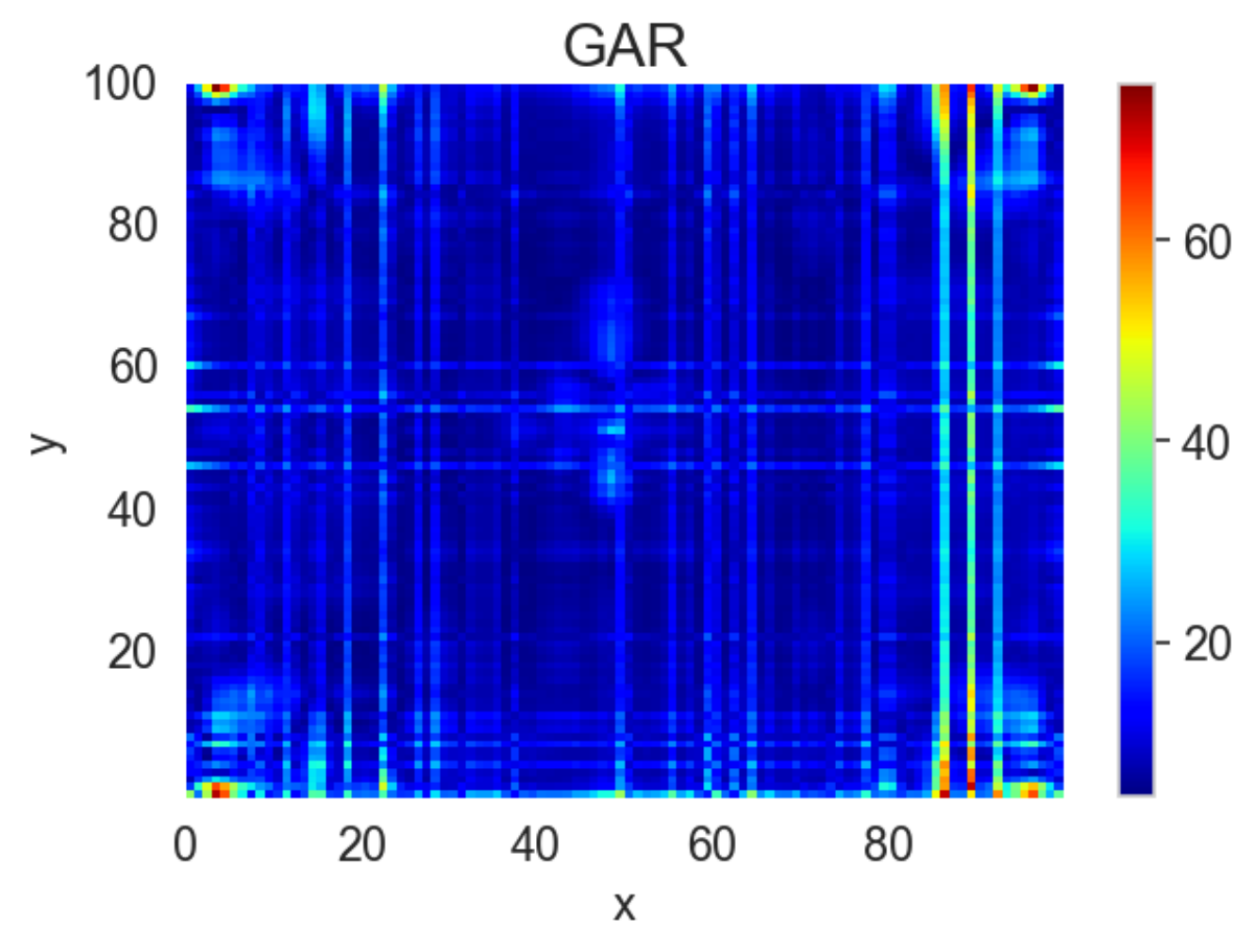}
    \end{minipage}%
    \hfill
    \begin{minipage}[t]{0.2\linewidth}
    \centering
    \includegraphics[width=3.25cm]{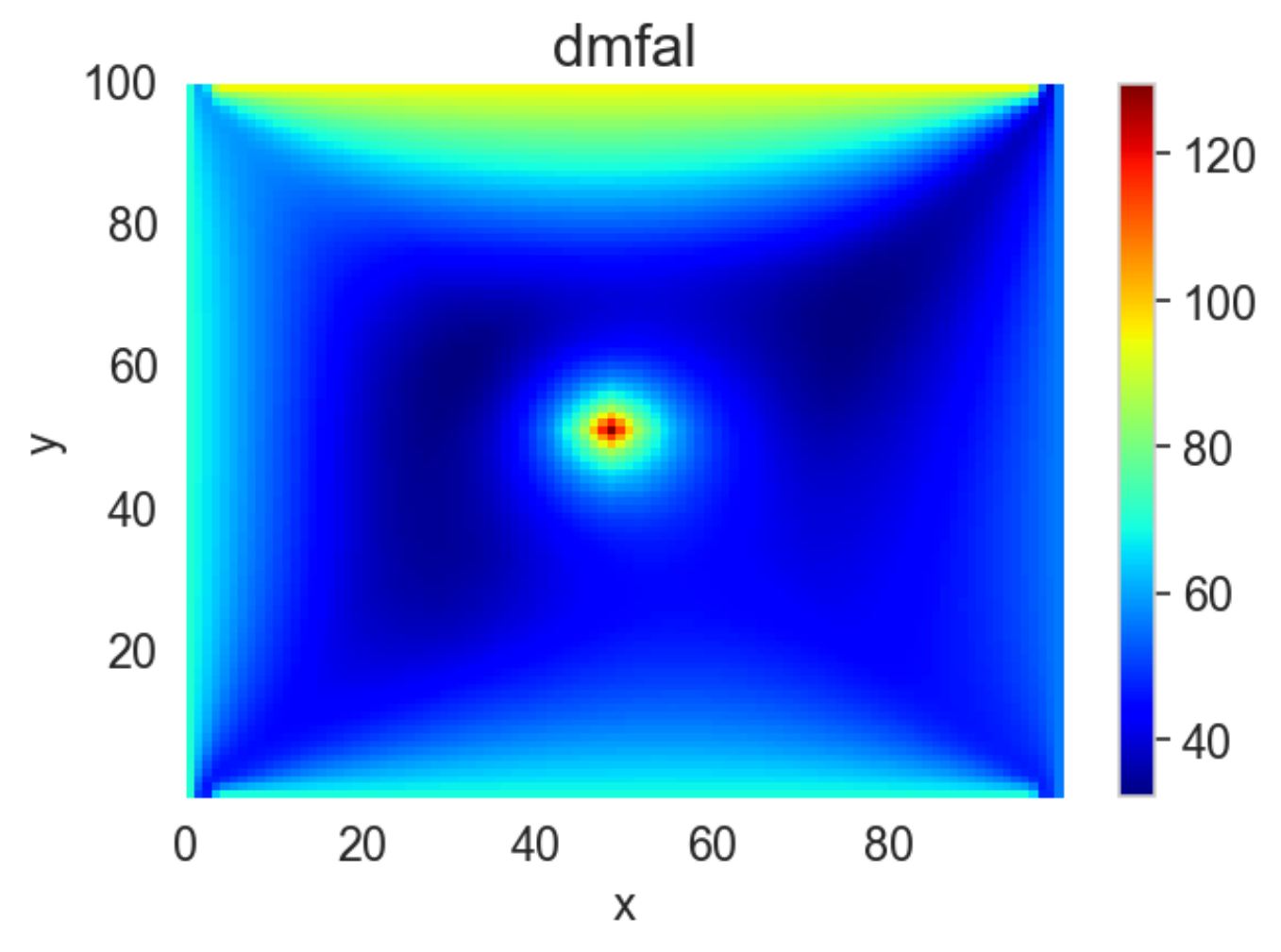}
    \end{minipage}%
    \hfill
    \begin{minipage}[t]{0.2\linewidth}
    \centering
    \includegraphics[width=3.25cm]{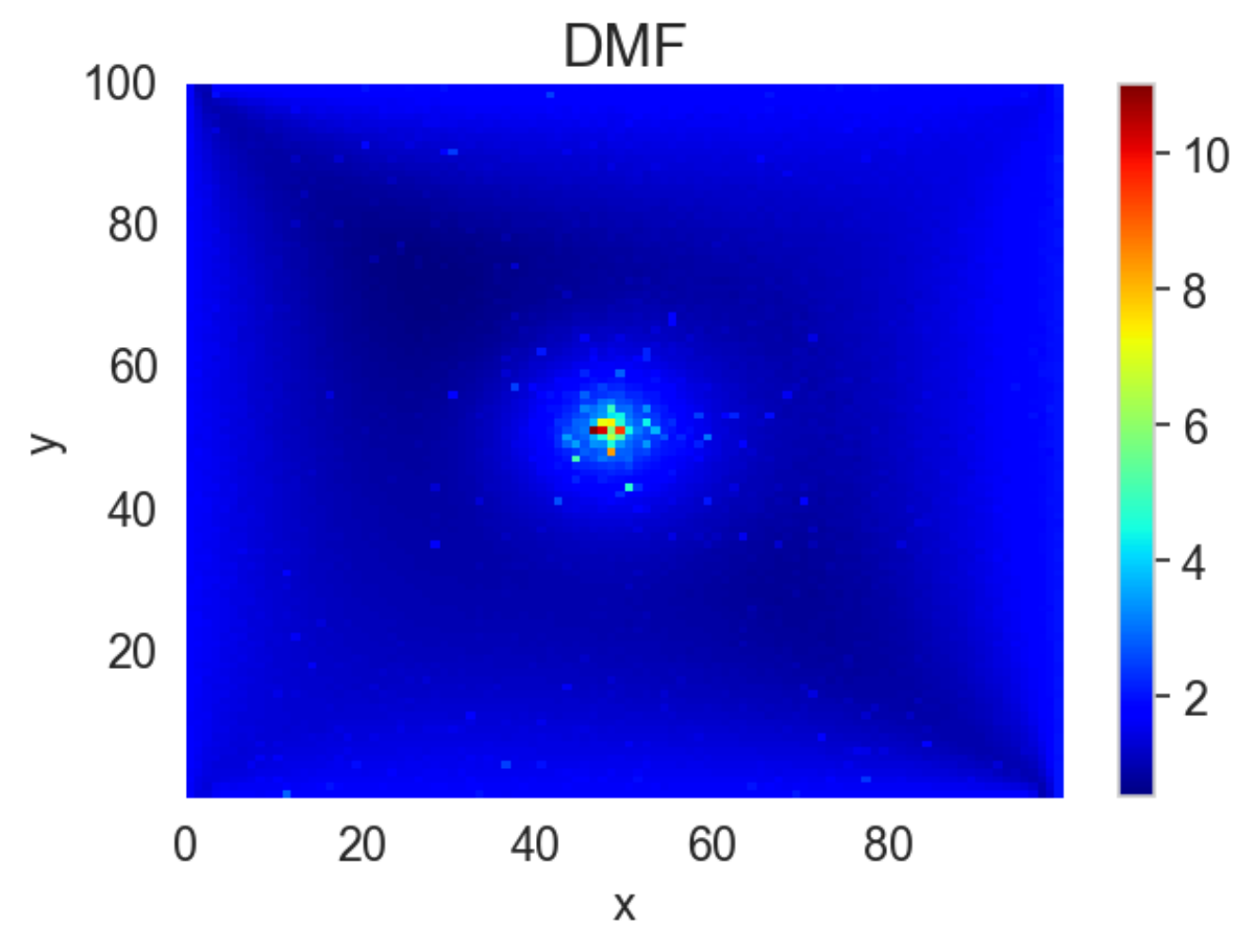}
    \end{minipage}%
    \hfill
    \\\begin{minipage}[t]{0.2\linewidth}
    \centering
    \includegraphics[width=3.25cm]{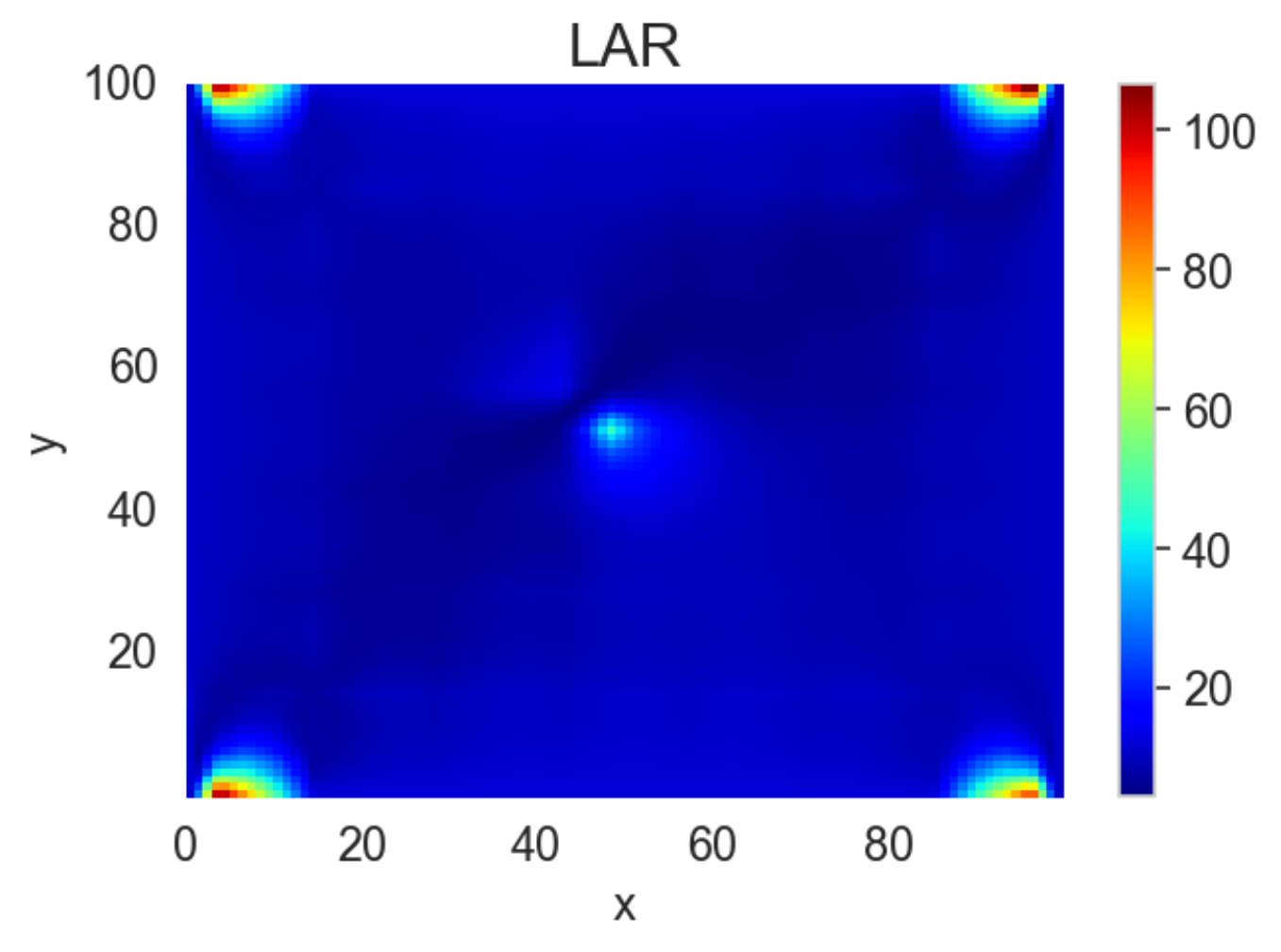}
    \end{minipage}%
    \hfill
    \begin{minipage}[t]{0.2\linewidth}
    \centering
    \includegraphics[width=3.25cm]{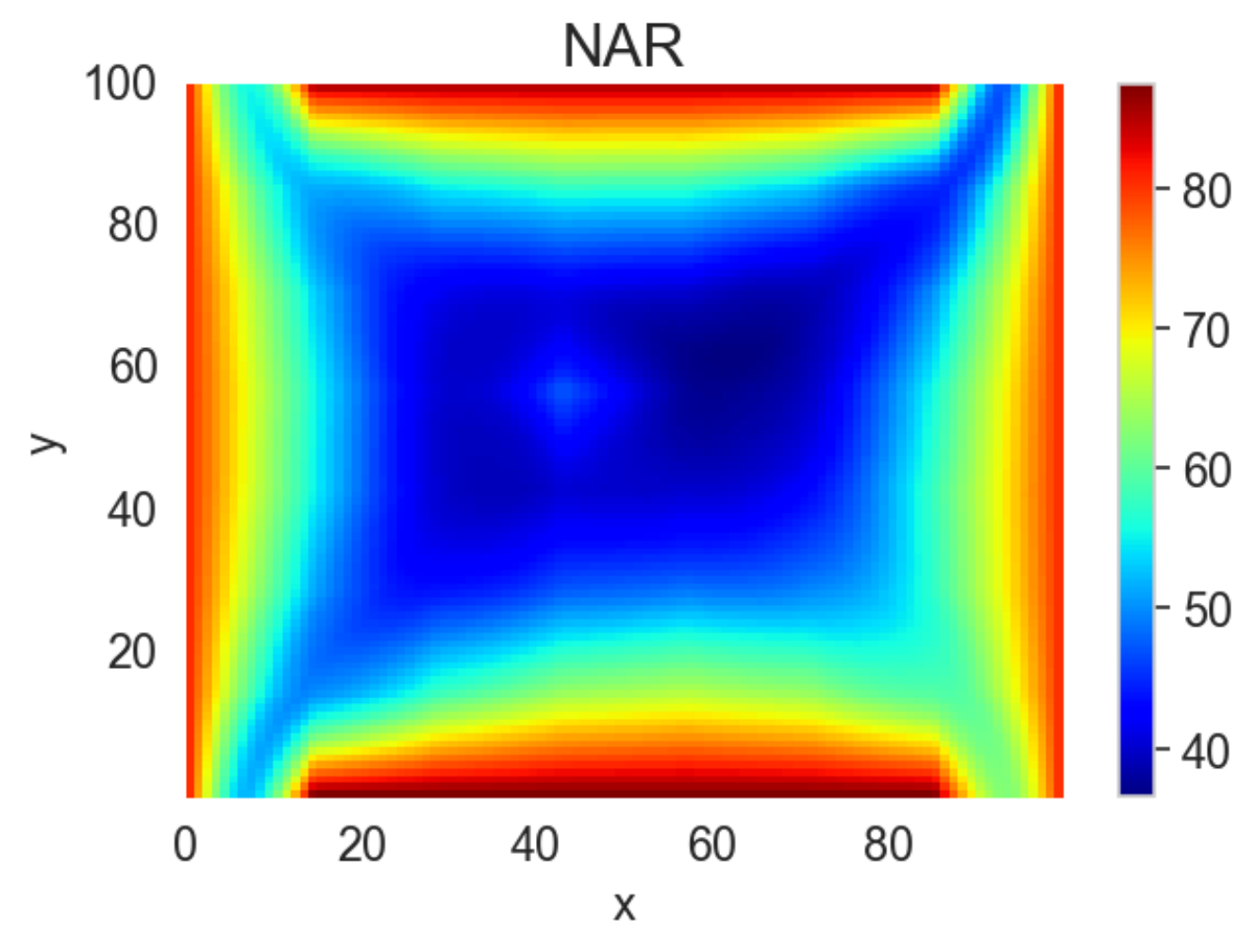}
    \end{minipage}%
    \begin{minipage}[t]{0.2\linewidth}
    \centering
    \includegraphics[width=3.25cm]{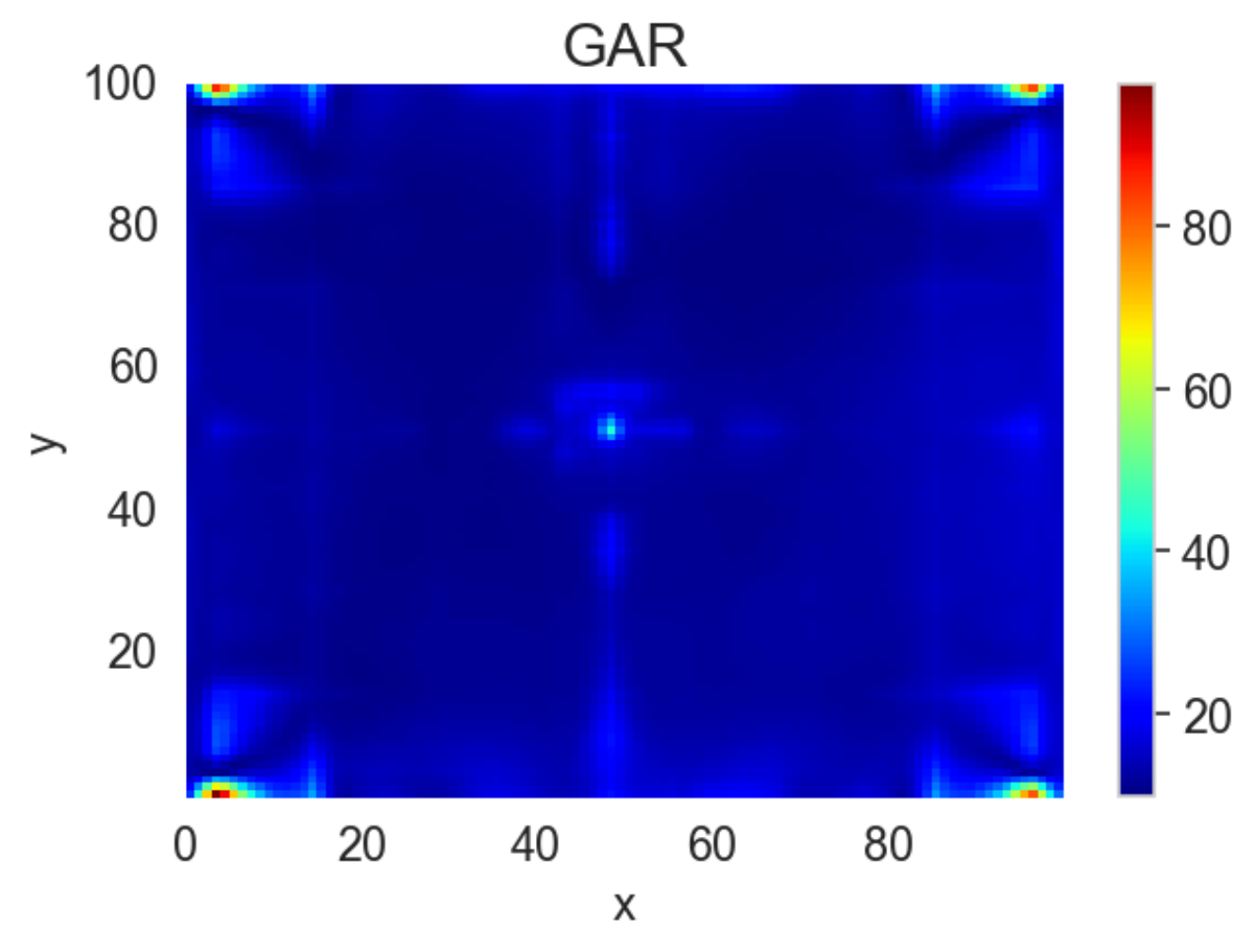}
    \end{minipage}%
    \hfill
    \begin{minipage}[t]{0.2\linewidth}
    \centering
    \includegraphics[width=3.25cm]{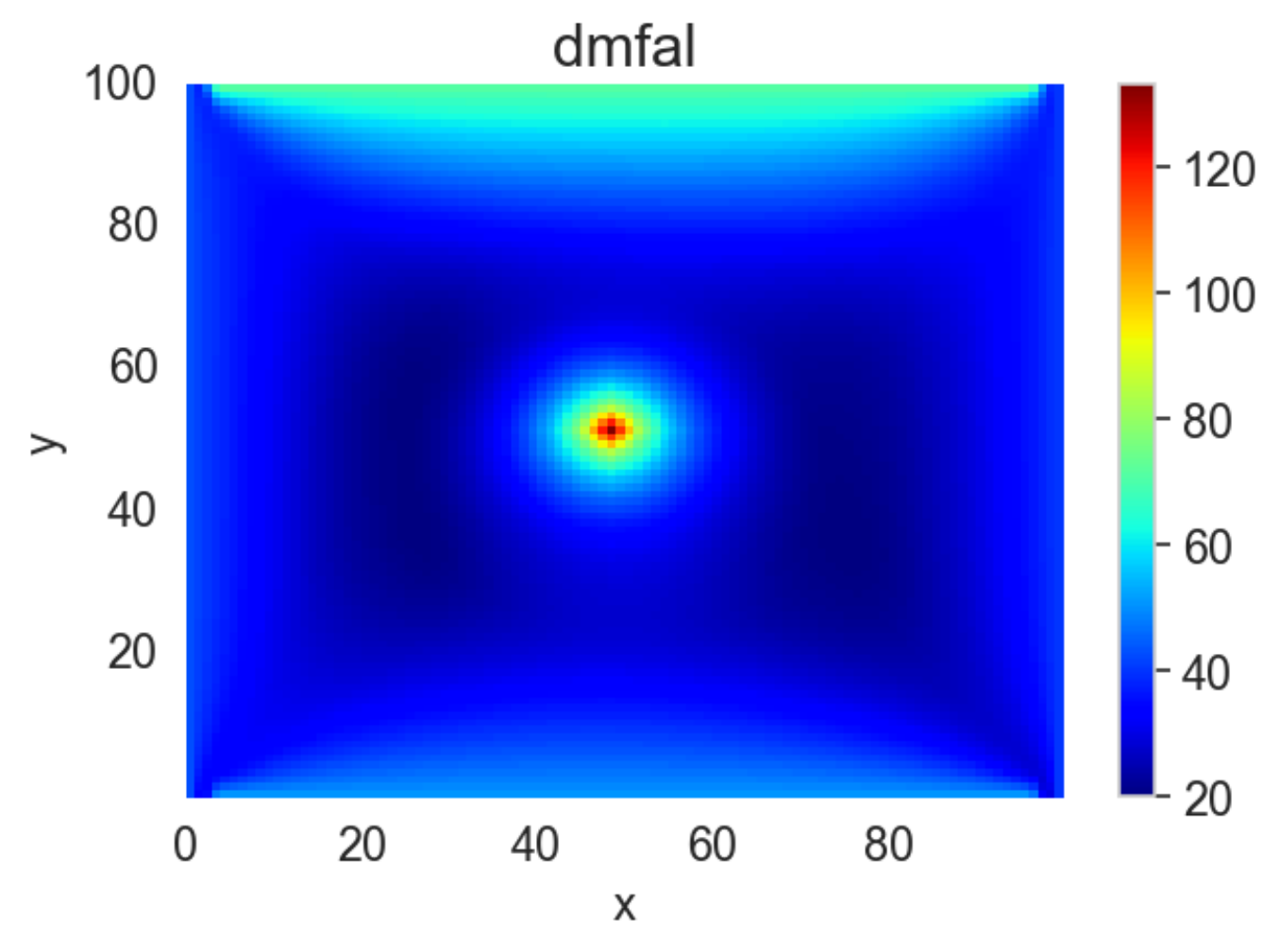}
    \end{minipage}%
    \hfill
    \begin{minipage}[t]{0.2\linewidth}
    \centering
    \includegraphics[width=3.25cm]{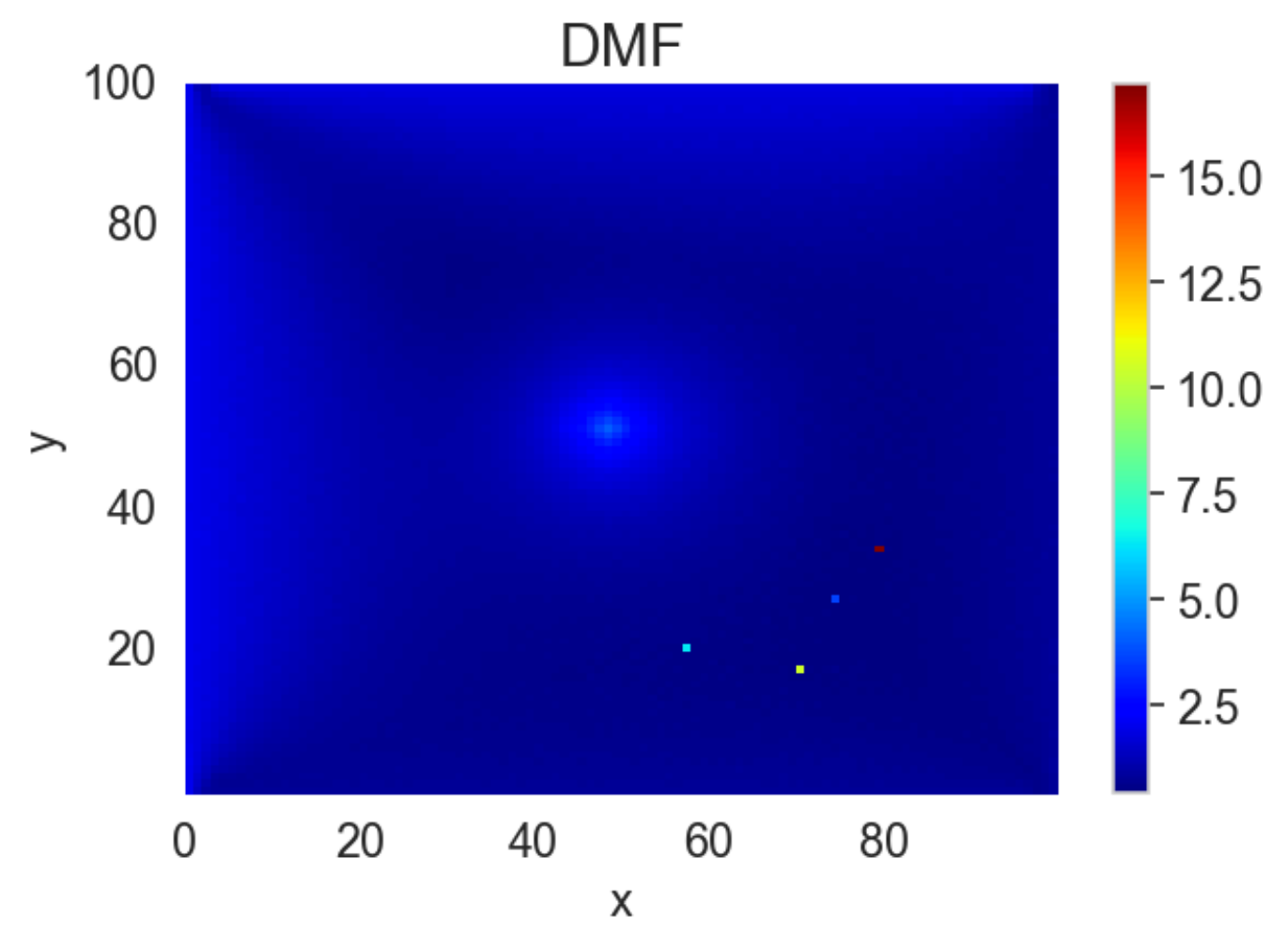}
    \end{minipage}%
    \hfill
    
    \caption{Comparison of Absolute Errors (AE) between different methods on Poisson dataset, LAR, NAR, GAR, dmfal, DMF (left to right columns), using high-fidelity data $\{4,8,16,32\}$ (top to bottom rows) with fixed 32 low-fidelity data.}
    \label{exp_23}
\end{figure}




\begin{figure}[!htbp]
    \centering
    \includegraphics[width=5cm]{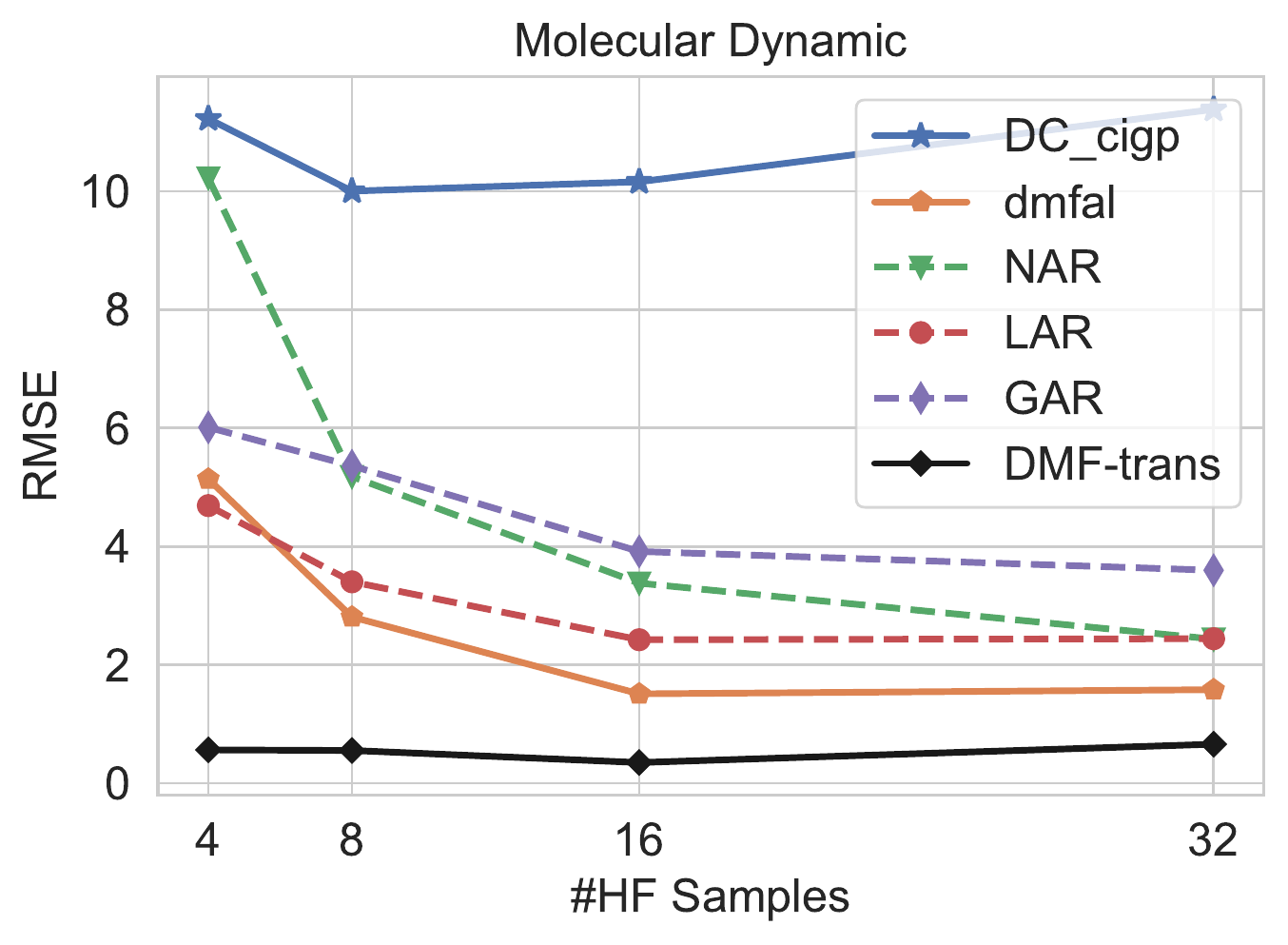}
    \includegraphics[width=5cm]{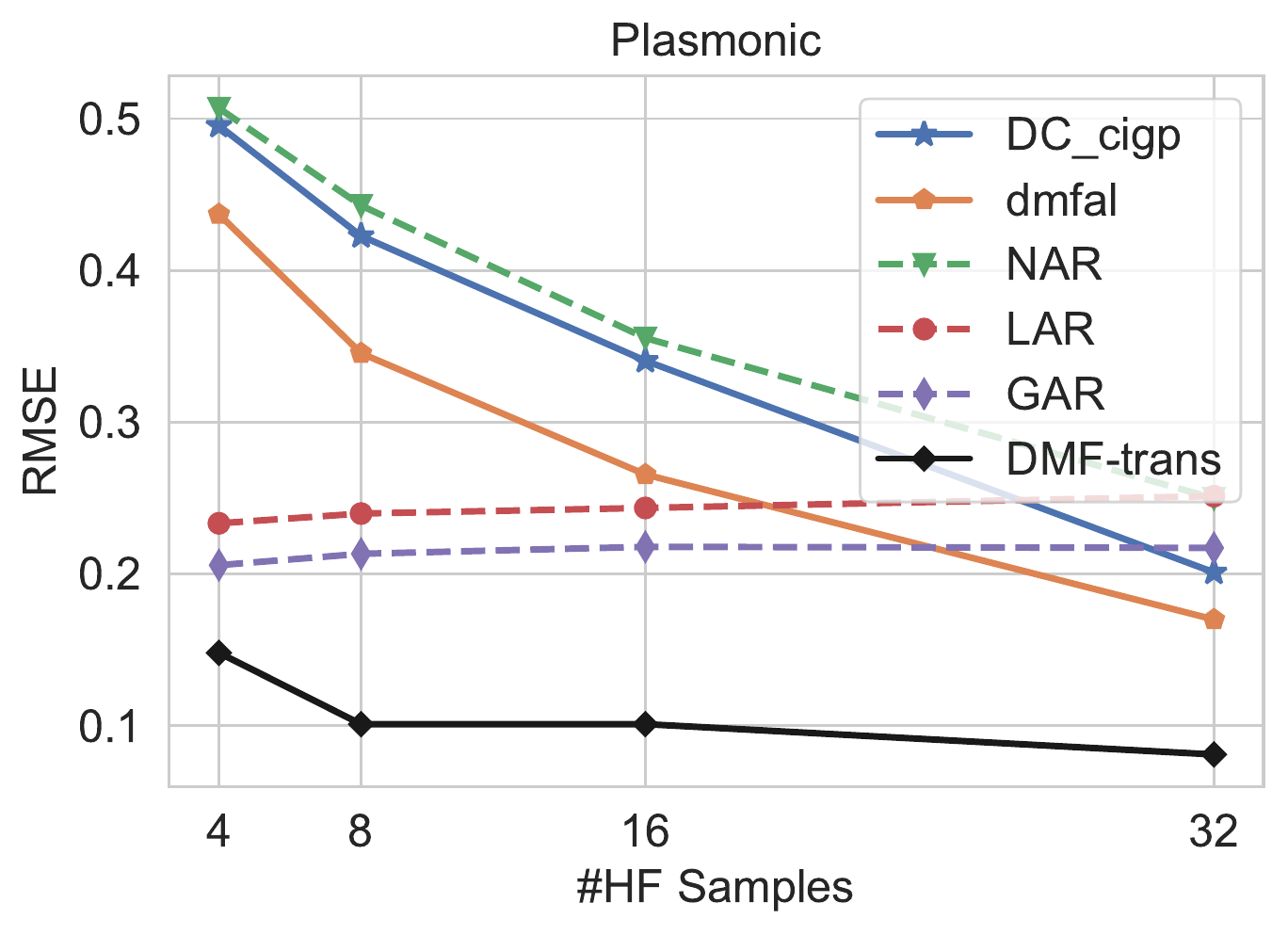}
    \includegraphics[width=5cm]{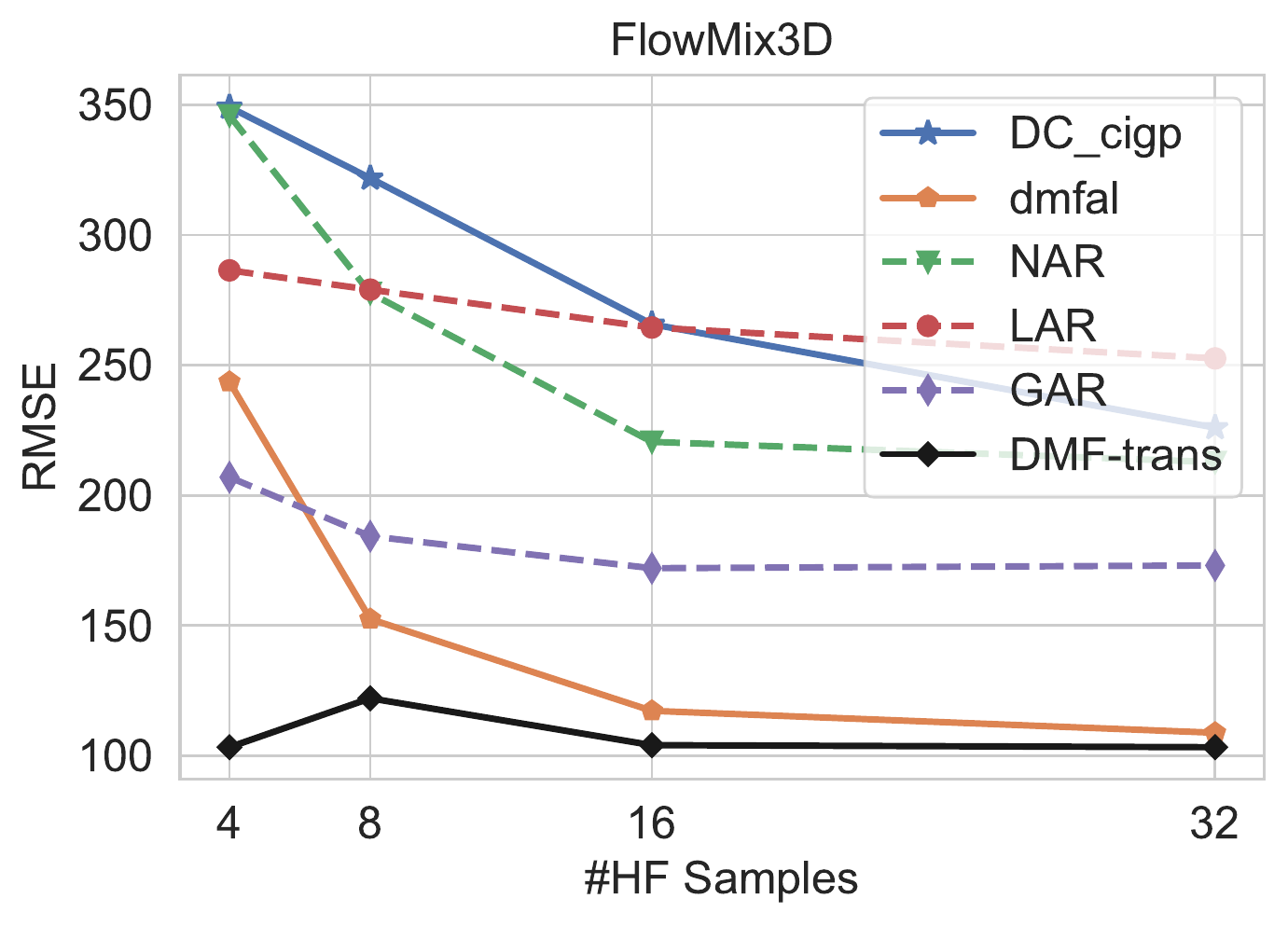}
    \caption{Comparison of SOTA methods on real test examples}
    \label{exp35}
\end{figure}

\subsection{Real-World Applications}
Finally, we assess our model on multi-fidelity data generated from three real-world applications, namely, molecular dynamics (MD) simulation, plasmonic array design, and fluid flow simulation.
The detailed problem and experimental settings will be discussed soon in the following subsections.
We first demonstrate the significant improvement of DMF on these real-world applications with only a small number of high-fidelity training samples.
Following the same experimental setup as in previous sections, we compare DMF with other SOTA methods on these real-world applications and show the RMSE against an increasing number of high-fidelity training samples in Fig.~\ref{exp35}.
It is clear that DMF outperforms other SOTA methods in all three applications by a significant margin.
In particular, DMF achieves the best performance with only 4 high-fidelity training samples in all three problems, which is much less than the number of high-fidelity training samples required by other SOTA methods.
Even with 32 high-fidelity training samples, the competitors still cannot catch up with DMF with just only 4 high-fidelity training samples.
This indicates that DMF is able to learn the underlying relationship between high- and low-fidelity data with a small number of high-fidelity training samples with the help of automatic machine learning as proposed in this work.
Another interesting observation is that the other deep-learning-based methods, dmfal, also achieves good performance in all three applications particularly when the number of high-fidelity training samples is sufficient.
This highlights the potential of deep learning in multi-fidelity fusion over traditional machine learning methods.
With the help of transfer learning, DMF can leverage the knowledge from low-fidelity data more effectively and unleash the great potential of deep learning in multi-fidelity fusion.

%




\noindent \textbf{Molecular Dynamics simulation}:
We look at a molecular dynamics simulation~\citep{lee2016computational} to demonstrate the effectiveness of DMF in high-dimensional data.
The interatomic interaction in the area of molecular dynamics (MD) is defined using a potential energy expiration. The Lennard-Jones (LJ) potential~\citep{lee2016computational} is a well-known example for simulating these interactions in the MD system: 
\begin{equation}
u\left(r_{ij}\right)=4\varepsilon \left [ \left(\frac{\sigma}{r_{ij}}\right)^{12}-\left(\frac{\sigma}{r_{ij}}\right)^6 \right ],
\label{eq:LJ1}
\end{equation}
where $\epsilon$ is the potential well depth, $\sigma$ defines the length scale for this pairwise interatomic interaction model,
and $r_{ij}$ denotes the pairwise distance between particles $i$ and $j$. When the integration time step used to solve the MD system's Lagrangian governing equations is too large during this potential energy model's MD simulation, the resulting numerical simulation suffers greatly from numerical instability. To prevent such a numerical instability, we set a limit on the magnitude of repulsive interactions when the ratio of $\frac{\sigma}{r_{ij}}$ exceeds 1.2, which may cause slight inaccuracies in the simulation results but will not affect the experiment conclusion as such a setup is applied across all the MD simulations.
%
%
%
We assume $\sigma$, $\varepsilon$, $m$ or molecular mass to be $3\angstrom$, $1\frac{\text{kcal}}{\text{mol}}$ and 12.01$\frac{\text{g}}{\text{mol}}$, respectively. Here, the boundary conditions on all sides of the cubic simulation box (with a width of 27.05$\angstrom$) are  viewed as intermittent.
Additionally, we define a uniform temperature and density grid as $T\times \rho: [500, 1000]\text{K}\times[36.27,701.29]\frac{\text{kg}}{\text{m}^3}$ ($\rho^{\ast}: [0.05,0.95]$) on 114 sample points, where  $\rho$ and $\rho^{\ast}$ are density and dimensionless density equal to $\frac{Nm}{V}$ and $\frac{N\sigma^3}{V}$, respectively, and the mass of each particle ($m$) and simulation box size ($L=V^{\frac{1}{3}}$) is set to 12.01 g/mol and 27.05$\angstrom$. 
%
   
\begin{figure}
\centering
\includegraphics[width=8cm]{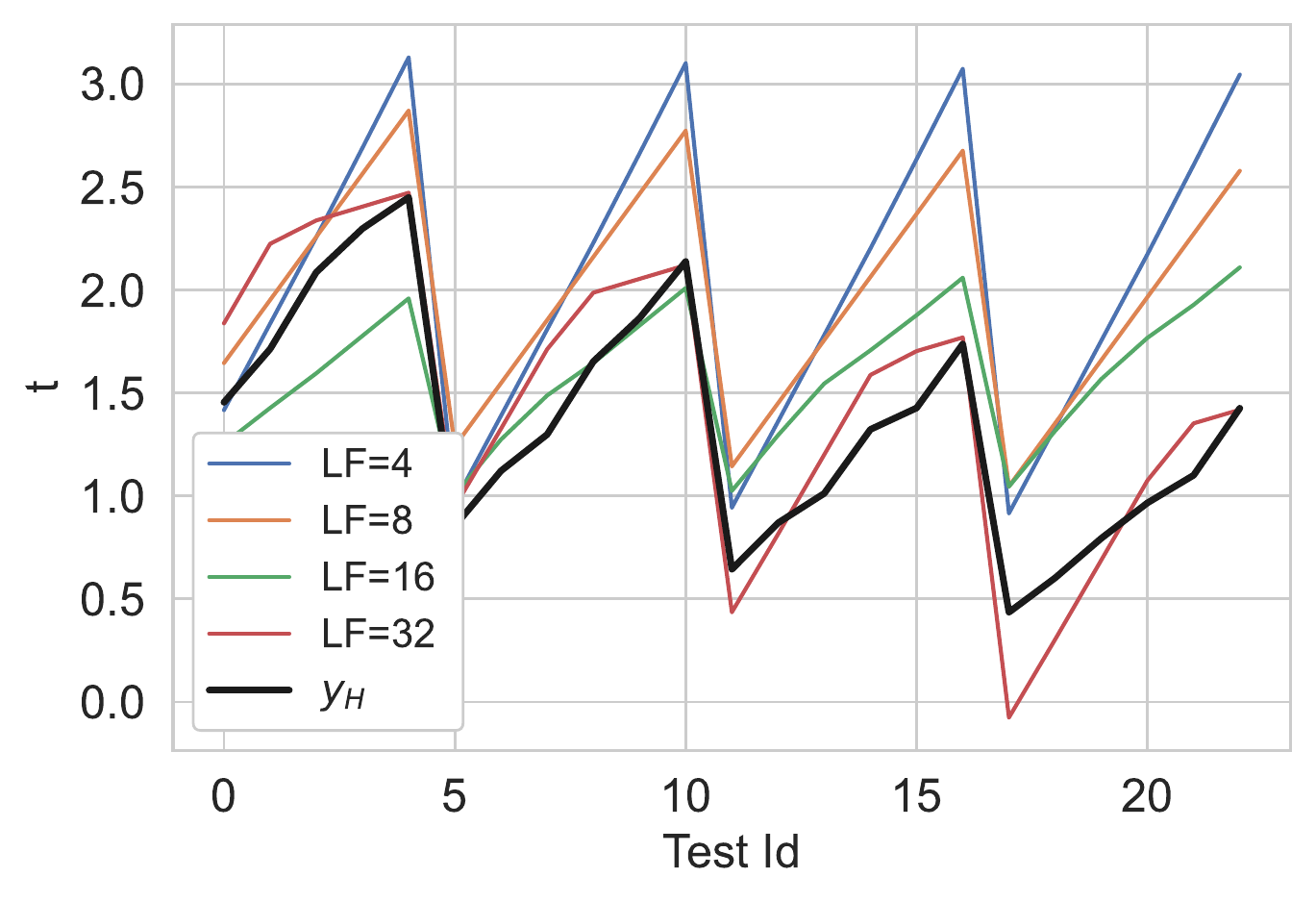}
\caption{The ground-truth temperature profiles for 23 test cases vs. DMF predictions with different numbers of low-fidelity training data.}
\label{mol_detail}
\end{figure}

For this problem, we are interested in the temperature profile of the system, which is obtained by averaging the temperature of each particle in the system.
The ground truth temperature profiles are obtained by running the high-fidelity MD simulation model with a time step of 1 fs.
In contrast, the low-fidelity model is run with a time step of 20 fs.
We fix 4 high-fidelity data and increase the number of low-fidelity data from 4 to 32 and compare the predictions of DMF with the ground truth temperature profiles on 23 test samples. The RMSE results are demonstrated in Figure \ref{exp35}.

In addition to the overall RMSE of Figure \ref{exp35}, now we show the detailed average temperature predictions of DMF with respect to ground truth on 23 test samples for the Molecular Dynamics simulation model.
Predictions based on different numbers of low-fidelity data are shown in Figure \ref{mol_detail}. As can be seen, as the amount of low-fidelity data increases, the maximum errors decrease, and the predictions gradually approach the ground truth. 
With 32 low-fidelity data, the maximum error is about 0.1 K, which is much smaller than the maximum error of 0.5 K obtained by the Gaussian process regression (GPR) method~\citep{lee2016computational} and the maximum error of 0.3 K obtained by the deep neural network (DNN) method~\citep{lee2016computational}.

\noindent\textbf{Plasmonic nanoparticle arrays}:
Coupled Dipole Approximation (CDA) is a technique for imitating the optical response of a collection of similar metallic nanoparticles that are not magnetic and have dimensions much smaller than the wavelength of light (in this case, 25 nm).
The extinction and scattering efficiencies $Q_{ext}$ and $Q_{sc}$ for plasmonic systems with varying numbers of scatterers are defined as the QoIs in this work.
%
We examined particle arrays resulting from Vogel spirals (see Fig. \ref{vogel}). Since the number of interactions of incident waves from particles influences the magnetic field, the number of nanoparticles in a plasmonic array substantially affects the local extinction field caused by plasmonic arrays. The configurations of Vogel spirals with particle numbers in the set $\{2,25,50\}$ that define fidelities F1 through F3 are depicted in Fig. \ref{vogel}.
$\lambda\in[200,800]$ nm, $\alpha_{vs}\in[0,2\pi]$ rad, and $a_{vs}\in(1,1500)$ were determined to be the parameter space. 
These are, respectively, the incidence wavelength, the divergence angle, and the scaling factor. A Sobol sequence was utilized to choose inputs. The computing time required to execute CDA increases exponentially as the number of nanoparticles increases. Consequently, the proposed sampling approach results in significant reductions in computational costs.
    
\begin{figure}[htbp]
    \centering
    \includegraphics[width=0.45\textwidth,trim={8cm 0 8cm 0},clip]{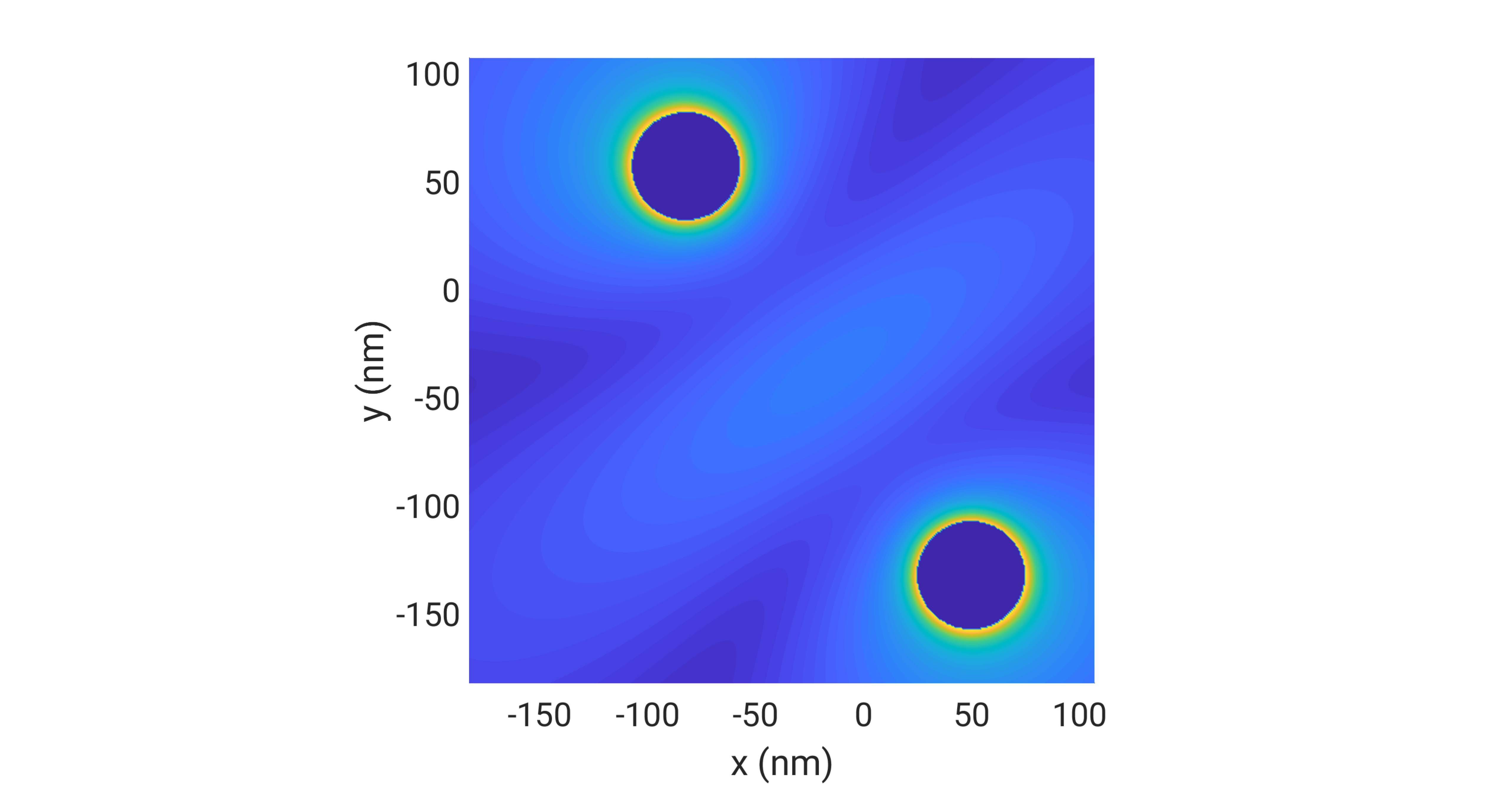}
    \includegraphics[width=0.45\textwidth,trim={8cm 0 8cm 0},clip]{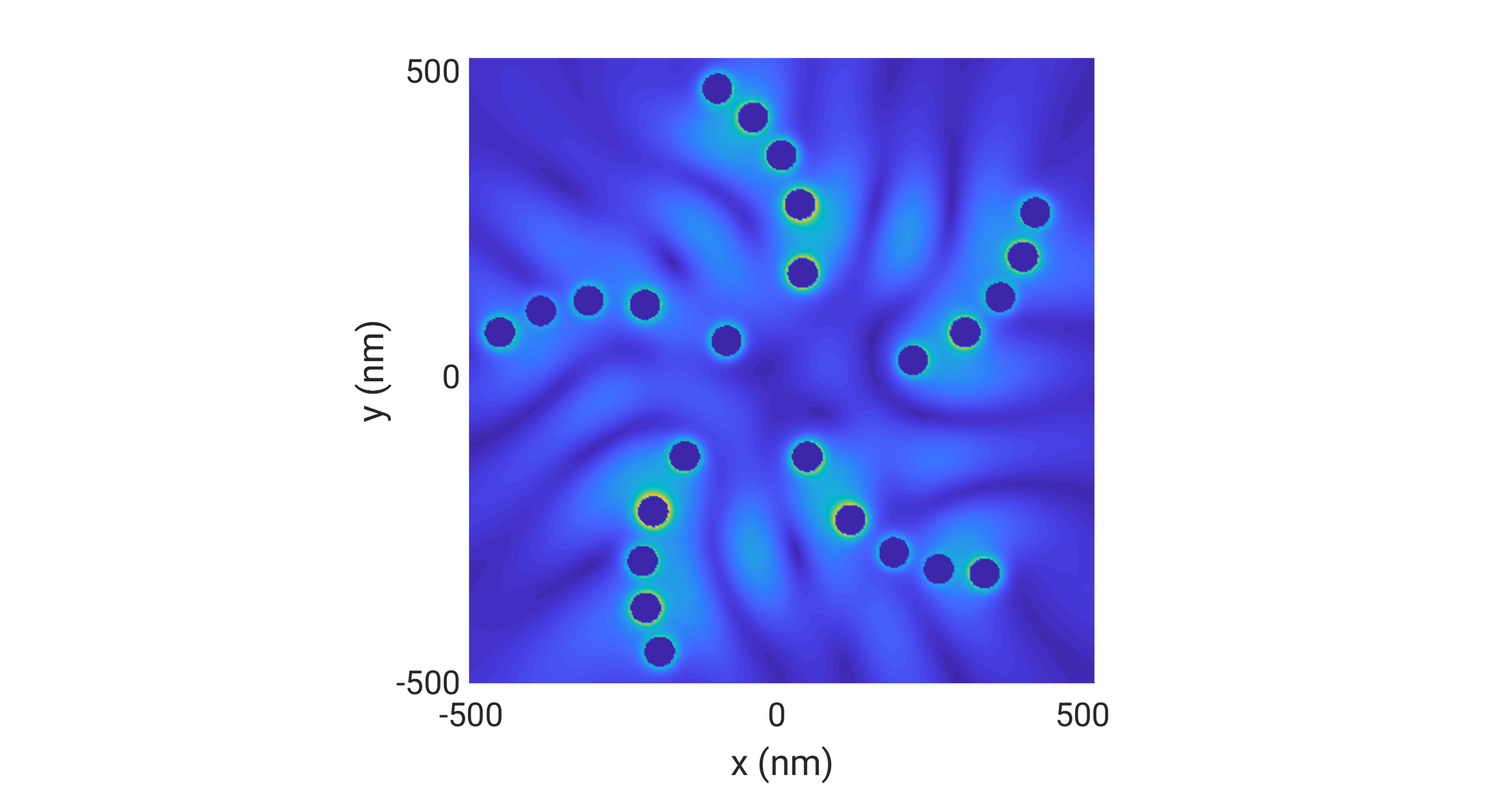}
    \includegraphics[width=0.45\textwidth,trim={8cm 0 8cm 0},clip]{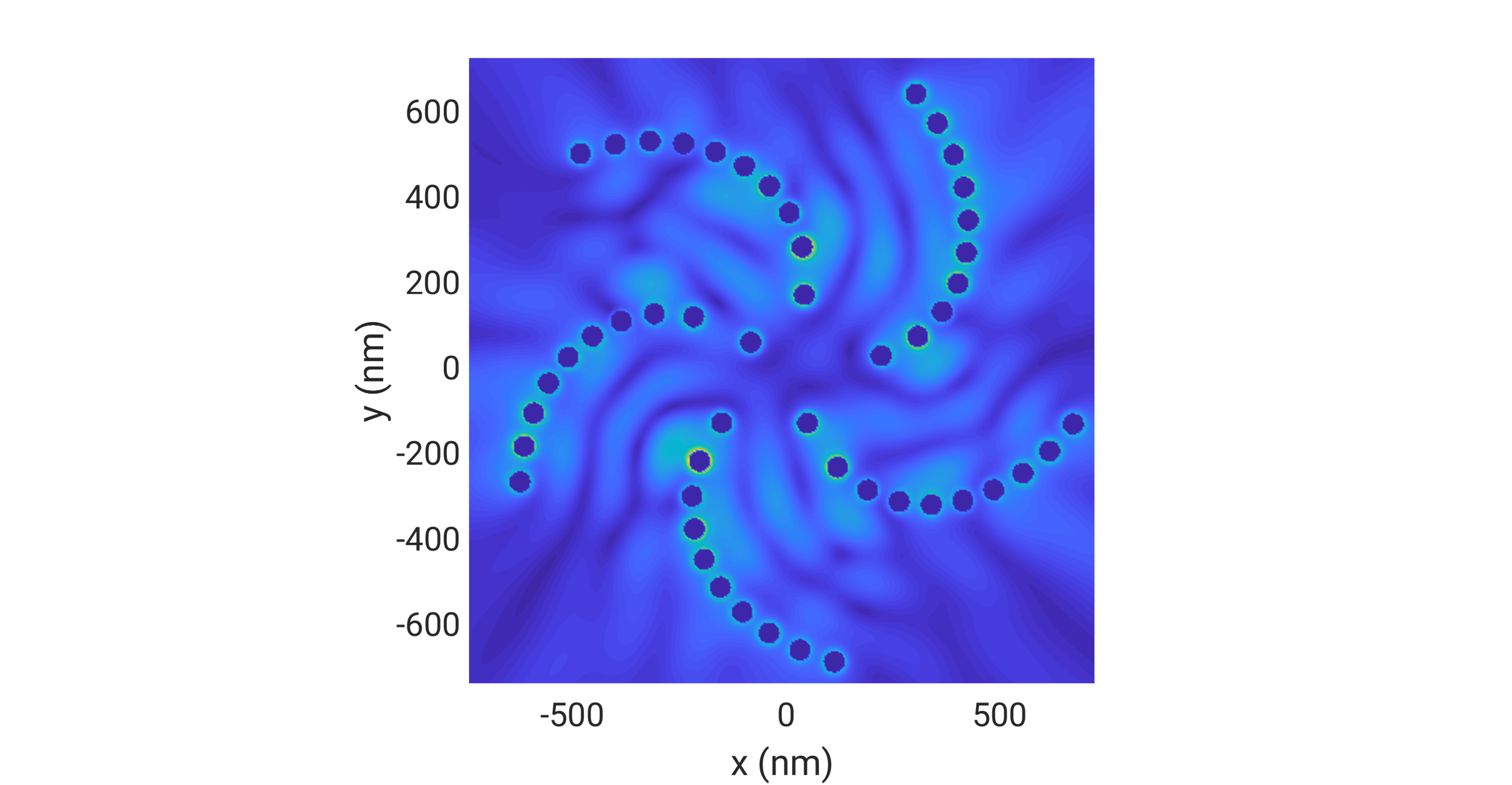}
    \includegraphics[width=0.45\textwidth,trim={8cm 0 8cm 0},clip]{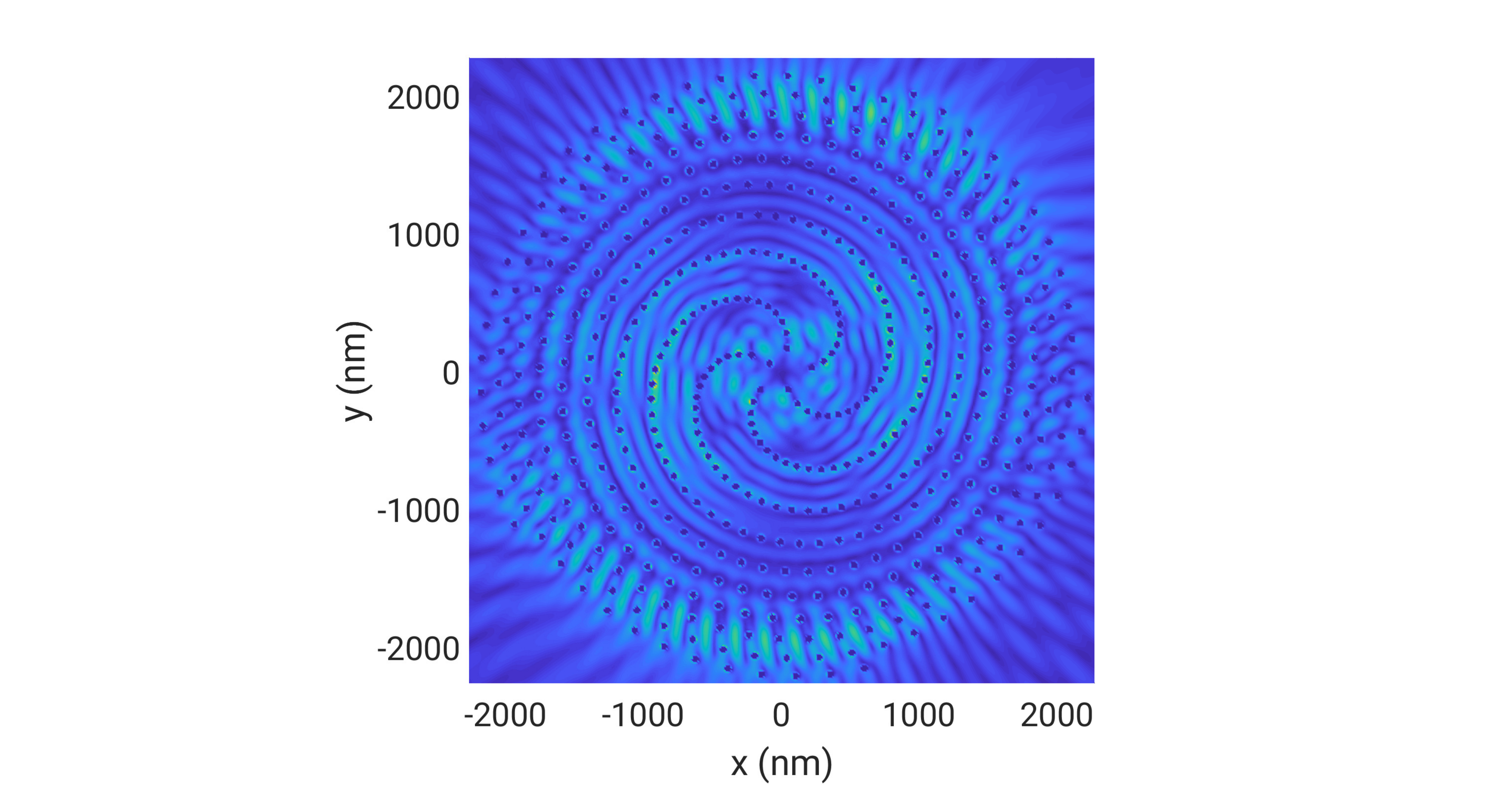}
    \caption{Sample configurations of Vogel spirals with $\{2,25,50,500\}$ particles.}
\label{vogel}
\end{figure} 
    
The solution of the local electric fields, $E_loc(r_j)$, for each nano-sphere, can be used to calculate a plasmonic array's response to electromagnetic radiation. Considering $N$ metallic particles defined by the same volumetric polarizability $\alpha(\omega)$ and situated at vector coordinates $\r_i$, it is possible to calculate the local field $\E_{loc}(\r_j)$ by solving~\citep{guerin2006effective} the corresponding linear equation.
    
    \begin{equation}\label{eqFoldyLax}
    \E_{loc}(\mathbf r_i)=\mathbf E_0({\mathbf r_i})-\frac{\alpha k^2}{\epsilon_0} \sum_{j=1,j\neq i}^{N} \mathbf{\tilde{G}}_{ij} \mathbf E_{loc}(\mathbf r_j)
    \end{equation}
    in which $\mathbf E_0(\r_i)$ is the incident field, $k$ is the wave number in the background medium, $\epsilon_0$ denotes the dielectric permittivity of vacuum ($\epsilon_0 = 1$ in the CGS unit system), and $\mathbf{\tilde{G}}_{ij}$ is constructed from $3\times3$ blocks of the overall $3N\times3N$ Green's matrices for the $i$th and $j$th particles. $\mathbf{\tilde{G}}_{ij}$ is a zero matrix when $j = i$, and
    otherwise calculated as
    \begin{equation}\label{Gij}\tilde{\bf G}_{ij}=\frac{\exp(ikr_{ij})}{r_{ij}}\left\{{\bf I}-\widehat{\r}_{ij}\widehat{\r}_{ij}^T-\left[\frac{1}{ikr_{ij}}+\frac{1}{(kr_{ij})^2}({\bf I}-3\widehat{\r}_{ij}\widehat{\r}_{ij}^T)\right]\right\}
    \end{equation}
    where $\widehat{\r}_{ij}$ denotes the unit position vector from particles $j$ to $i$ and $r_{ij}=|\r_{ij}|$. 
    By solving Equations~\ref{eqFoldyLax} and \ref{Gij}, the entire local fields $\mathbf E_{loc}(\r_i)$, the scattering and extinction cross-sections are calculated as a result, and the more accurate numerical solution is detailed in \citep{razi2019optimization}.
    $Q_{ext}$ and $Q_{sc}$ are derived by normalizing the scattering and extinction cross-sections relative to the array's entire projected area. We considered the Vogel spiral class of particle arrays, which is described by~\citep{christofi2016probing}
    \begin{equation}\label{eqVSa}
    \rho_n = \sqrt{n}a_{vs} \quad \mbox{and}\quad 
    \theta_n = n\alpha_{vs},
    \end{equation}
    where $\rho_n$ and $\theta_n$ represent the radial distance and polar angle of the $n$-th particle in a Vogel spiral array, respectively. Therefore, the Vogel spiral configuration may be uniquely defined by the incidence wavelength $\lambda$, the divergence angle $\alpha_{vs}$, the scaling factor $a_{vs}$, and the number of particles $n$. 
    
    {In this problem, we fix the number of \lf training data to 96 samples, and the numbers of high-fidelity data are $\{4, 16,64, 96\}$; we evaluate on 34 withhold test samples. 
    The experimental results are shown in Fig.~\ref{34}, which shows the performance by comparing the predicted values $f_H(\x)$ and the true values $\y_H$ of the high-fidelity model with the $R^2$ statistics. 
    The $R^2$ statistics are calculated by the following formula:
    \begin{equation}
    R^2 = 1 - \frac{\sum_{i=1}^{n} (y_i - \hat{y}_i)^2}{\sum_{i=1}^{n} (y_i - \bar{y}_i)^2}
    \end{equation}
    where $\hat{y}_i$ is the predicted value of the $i$-th sample, and $\bar{y}_i$ is the mean value of the $i$-th sample.
    As can be seen, the predictions $f_H(\x)$ are close to the true values $\y_H$ when the number of high-fidelity data is only 4. 
    Nonetheless, the performance of the \hf model gradually improves as the number of high-fidelity data increases and we can see how the distributions of the scatter points are moving towards the line $y=x$ with the increase of the number of high-fidelity data.
    The trend is clearly shown with the $R^2$ statistics.
    It might seem that the improvement is very significant. We need to note that $R^2>0.95$ is a very high value, which means the strong correlations between two variables and the predictions $f_H(\x)$ are very close to the true values $\y_H$. It already indicates that the \hf model is very accurate in practice.
    %
    %

\begin{figure}[htbp]
    \begin{minipage}[t]{0.25\linewidth}
    \centering
    \includegraphics[width=4cm]{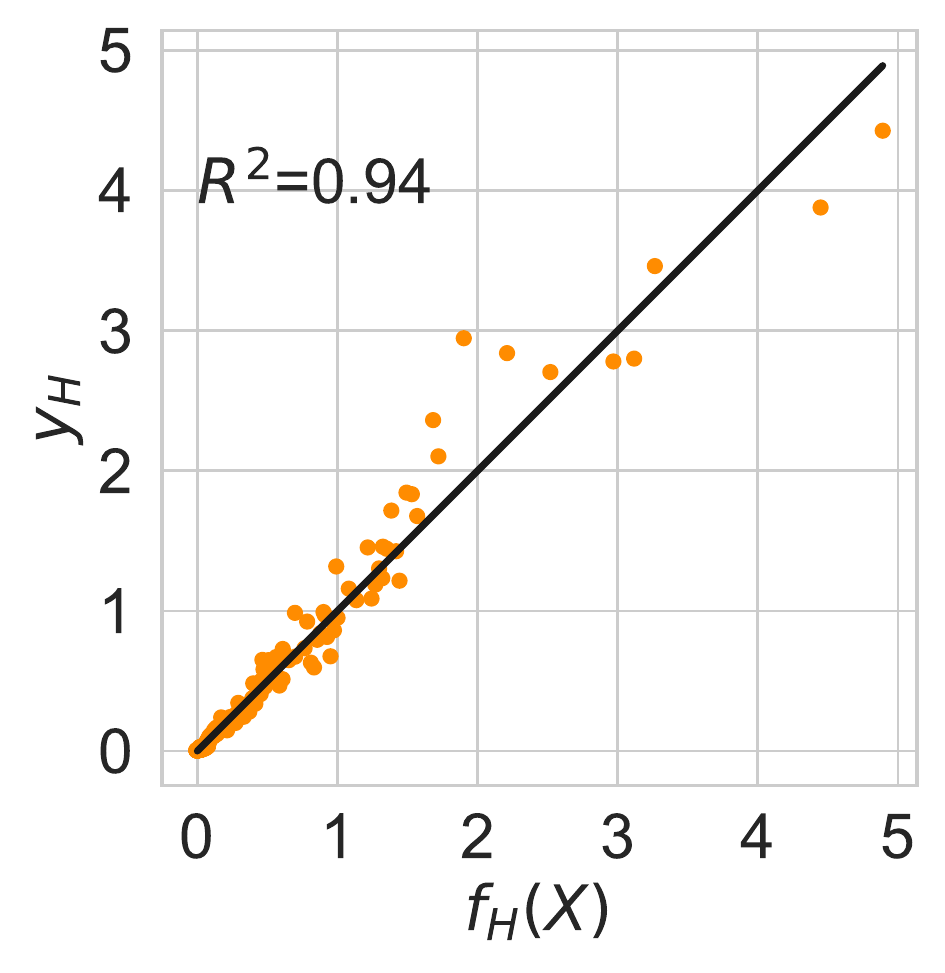}
    \end{minipage}%
    \hfill
    \begin{minipage}[t]{0.25\linewidth}
    \centering
    \includegraphics[width=4cm]{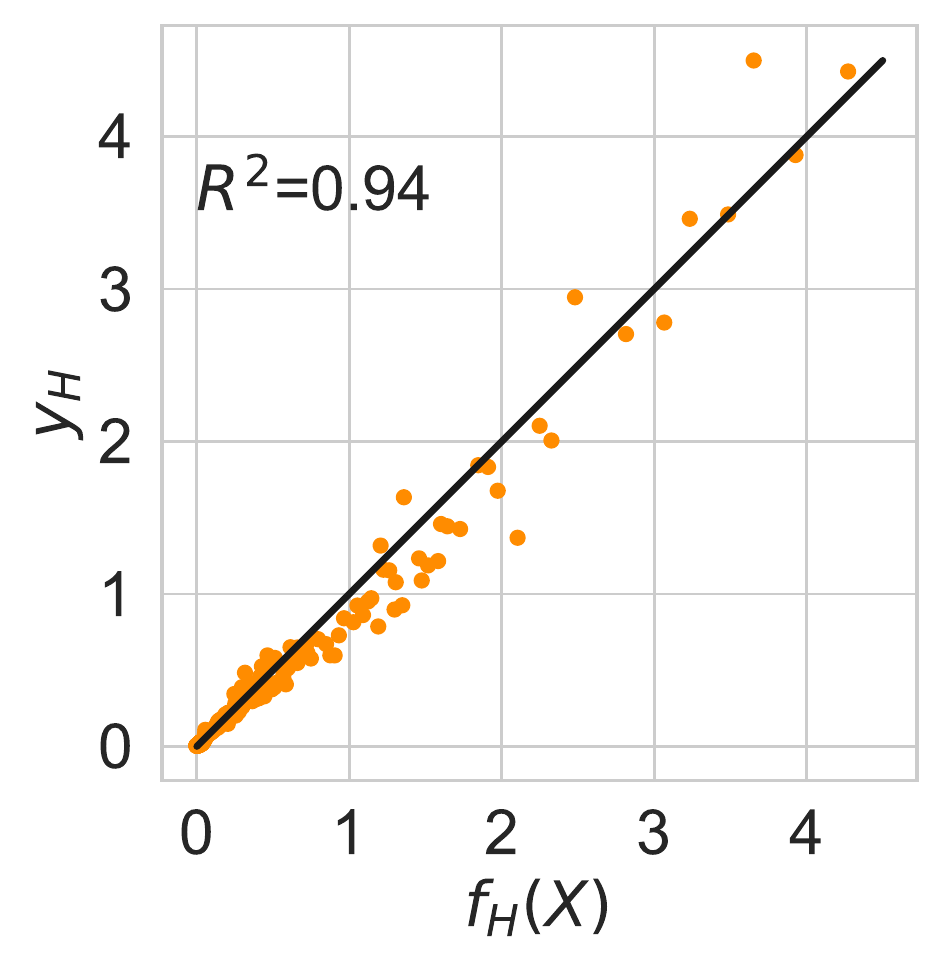}
    \end{minipage}%
    \begin{minipage}[t]{0.25\linewidth}
    \centering
    \includegraphics[width=4cm]{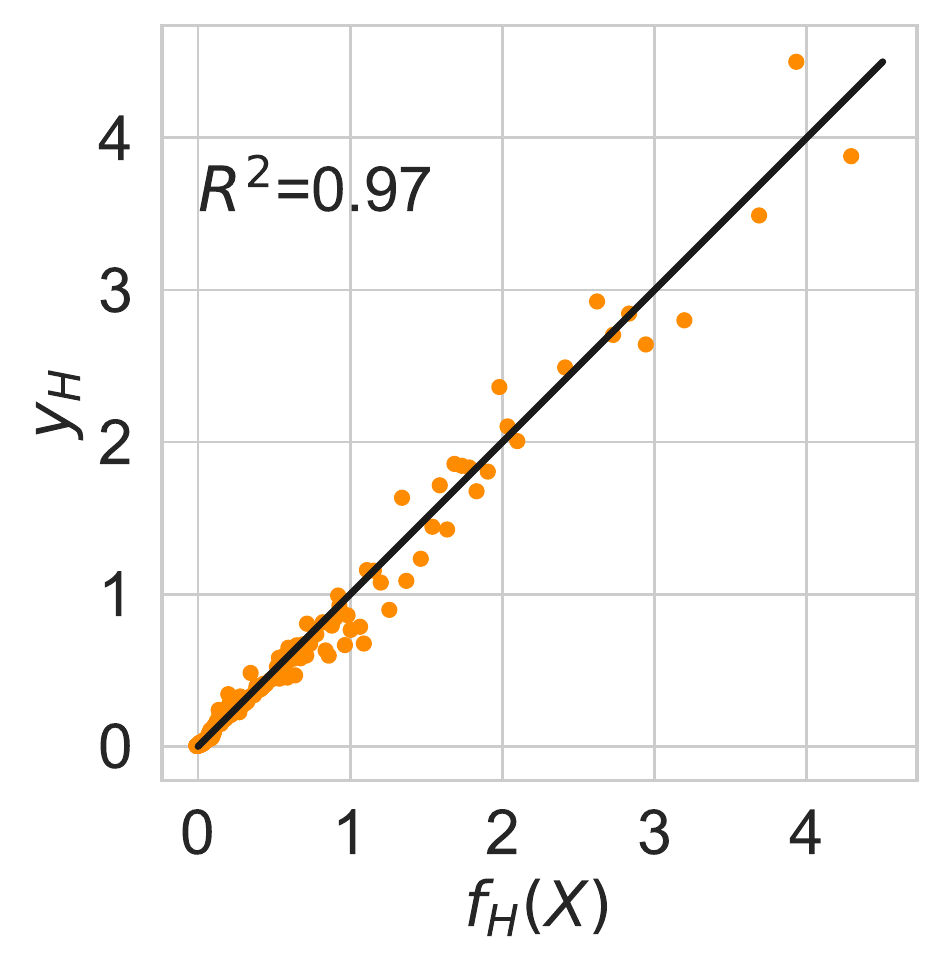}
    \end{minipage}%
    \hfill
    \begin{minipage}[t]{0.25\linewidth}
    \centering
    \includegraphics[width=4cm]{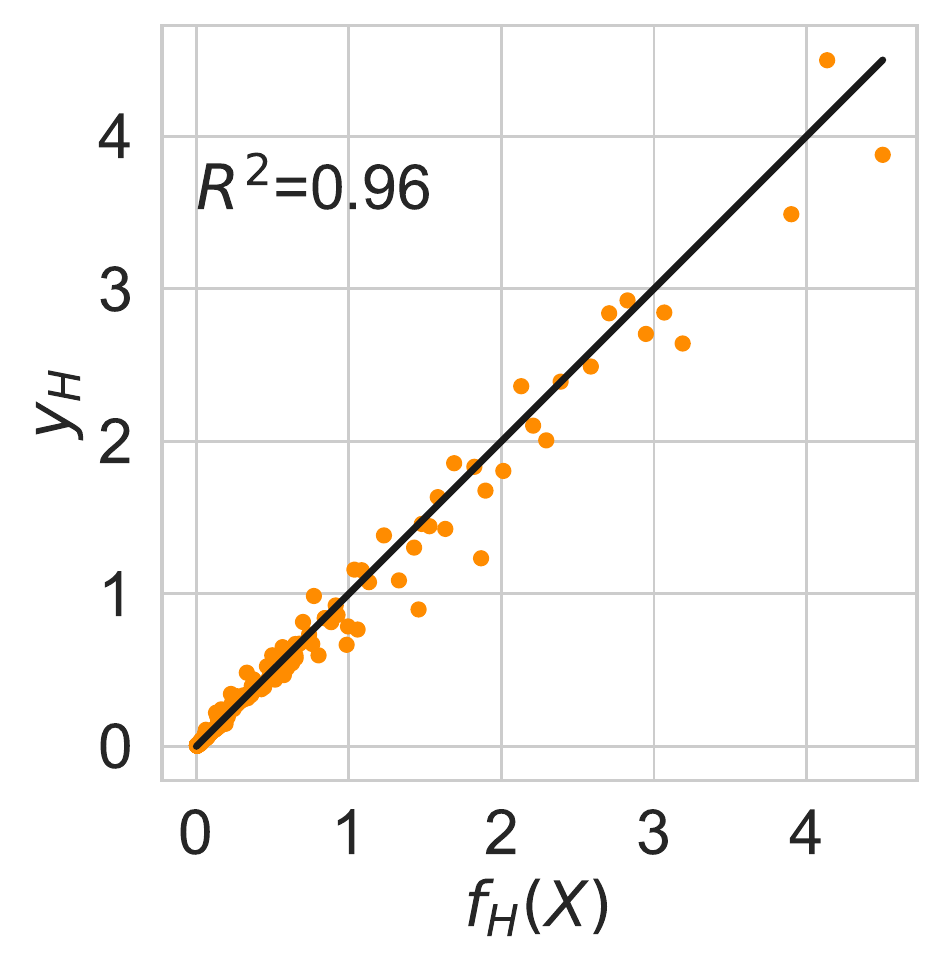}
    \end{minipage}%
    \hfill
    \caption{Relation between predictions $f_H(\x)$ and $\y_H$, the high-fidelity data are 4, 16, 64, 96 from left to right, with fixed 96 low-fidelity data. }
    \label{34}
\end{figure}

\noindent\textbf{Turbulent mixing flow in an elbow-shape pipe:} To model the turbulent mixing flows in an elbow-shape pipe, which is denoted briefly by FlowMix3D in this work, we can use different models, \eg from the Reynolds-averaged Navier–Stokes (RANS) equations to the Large Eddy Simulation (LES) equations.
\begin{figure}[htbp]
	\centering
    \includegraphics[width=0.49\textwidth]{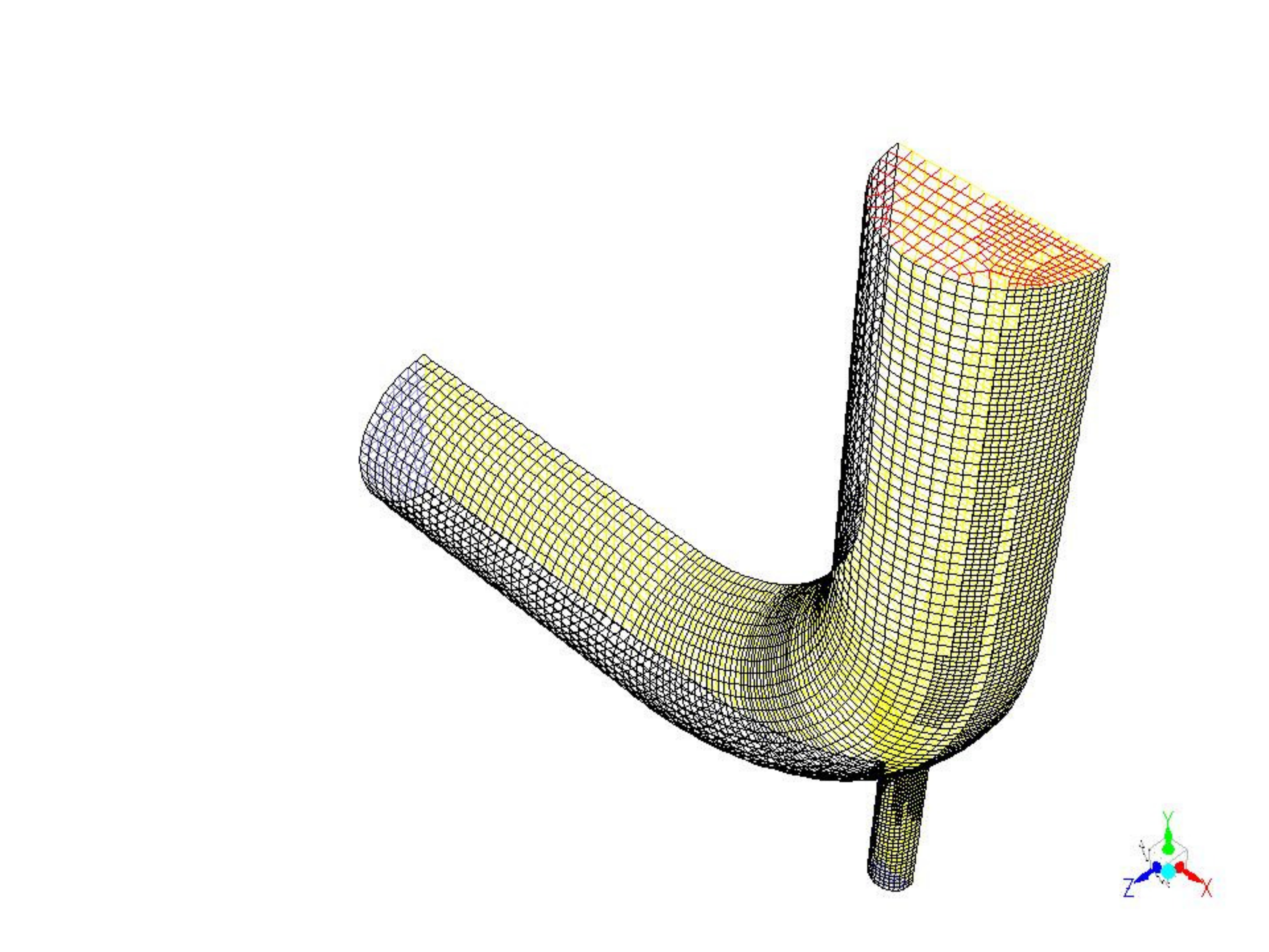}
    \includegraphics[width=0.49\textwidth]{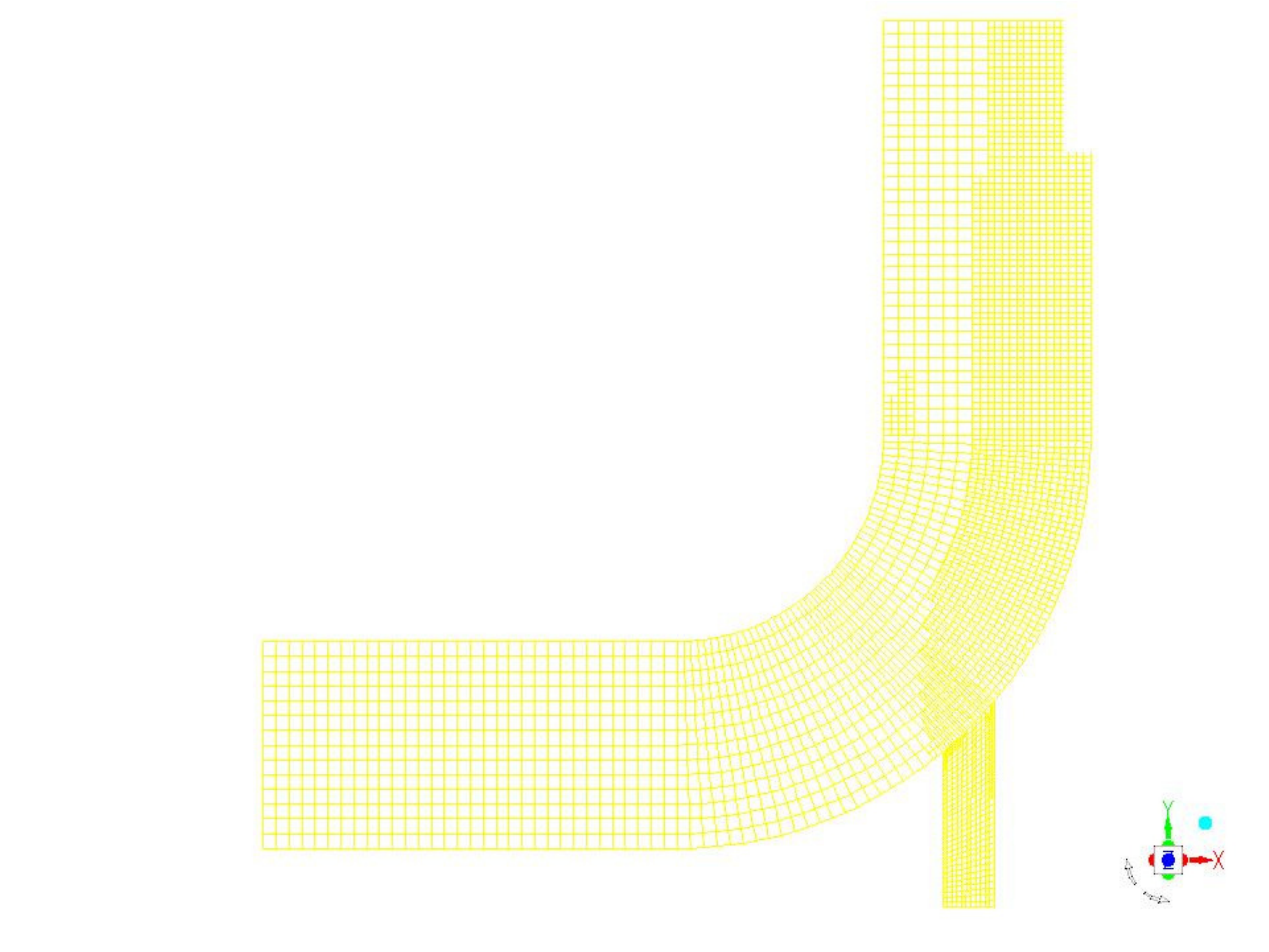}
	\caption{Computational domain and geometrical configuration of an elbow pipe and its plane of symmetry.}
	\label{fig mixflow}
\end{figure}
Both \hf and \lf are suboptimal solutions when it comes to the design and optimization of thermal-fluid systems.
In this problem, water enters from two inlets, the bottom left end of the pipe and a smaller inlet on the elbow, as shown in Fig.~\ref{fig mixflow}, From the top right. The pipe has a vertical upward exit for the water. At $t = 50 s$, the QoI were vectorized profiles of the velocity magnitude and pressure in circular cross sections of the elbow pipe, one located at the elbow junction (oriented at 45 degrees) and the other near the pipe exit (oriented at 0 degrees). The freestream velocity at the large inlet (with a diameter of 1 m) was chosen as the input parameter space, with values ranging from 0.2 to 2 m$ s^{-1}$, and the velocity at the smaller inlet (with a diameter of 0.5 m) was chosen with values ranging from 1.2 to 3 m$ s^{-1}$.

\section{Conclusion}
In the era of artificial intelligence, it is no longer a question of whether machine learning can be applied to engineering and scientific problems, but how to apply it effectively and efficiently.
In this paper, we propose a novel paradigm for learning a physics simulation efficiently and effectively by using 1) neural network architecture search, 2) multi-fidelity fusion through transfer learning, 3) continuous relaxation scheme, and 4) hyperparameter pruning technique.
Despite not being a rigorous Bayesian model (\eg AR) for multi-fidelity fusion, the proposed DMF method is able to emulate complex physics and outperform the SOTA methods in terms of accuracy and computational time on a variety of challenging benchmark problems and real-world examples to demonstrate its practicality.
We believe such an effort will inspire the development of deep learning based multi-fidelity fusion and other related problems in the future to achieve a new level of accuracy and efficiency in engineering and scientific applications.





We expect DMF to be an essential tool for understanding complex physics and an influential model in \mf fusion in the near future. The methods described in this paper appear to promise in producing accurate surrogate models capable of rapidly generating higher-fidelity data.

\bibliography{./dmf.bib}

\section{Appendix}\label{ap}

\subsection{Proof of Proposition 1}

We denote the loss of the model during training by $\mathcal{L}$,
\begin{equation}
    \mathcal{L}(\w,\Alpha)=\norm{f(\x,\w,\Alpha)-\y},
\end{equation}
where $\x $ is the input features, $\y$ represents \lf or \hf value. 
We train the $\w$ and $\Alpha$ respectively, where exists $\varepsilon_1$ and $\varepsilon_2$ so that $\Delta_{\w}\mathcal{L}(\w,\Alpha)<\varepsilon_1$ and $\Delta_{\Alpha}\mathcal{L}(\w,\Alpha)<\varepsilon_2$, therefore, we have: 

\begin{align}
    |\Delta_{\w,\Alpha}|&=|\mathcal{L}(\w+\delta \w,\Alpha+\delta\Alpha)-\mathcal{L}(\w,\Alpha)|\\
    &=|\mathcal{L}(\w+\delta \w,\Alpha+\delta\Alpha)-\mathcal{L}(\w+\delta \w,\Alpha)+\mathcal{L}(\w+\delta \w,\Alpha)-\mathcal{L}(\w,\Alpha)|\\
    &\leqslant |\mathcal{L}(\w+\delta \w,\Alpha+\delta\Alpha)-\mathcal{L}(\w+\delta \w,\Alpha)|+|\mathcal{L}(\w+\delta \w,\Alpha)-\mathcal{L}(\w,\Alpha)|\\
    &<\varepsilon_1+\varepsilon_2.
\end{align}

We have a Cauchy sequence $\mathcal{L}_n=\{\mathcal{L}(\x,\w_n,\Alpha_n) \}$ on $\mathbb{R}$, according to the completeness of $\mathbb{R}$, the $\mathcal{L}_n$ is convergent on $\mathbb{R}$, we denote the limit by $c$, \ie we have $\lim_{n\to\infty}\mathcal{L}_n(\w_n,\Alpha_n)=c$ and 
\begin{align}
    \x_j(\w,\Alpha)&=\sum_{i<j}\Alpha_{ij}\cdot \o_{ij}(\x_i(\w,\Alpha),\w_{ij})\\
    &=\sum_{i<j}\sum_{l=1}^{|\mathcal{O}|}\Alpha_{ijl}o_{ijl}(\x_i,\w_{ijl}),
\end{align}
Where $o_{ijl}$ denotes the $l$-th operation in operation candidate $\mathcal{O}$ from node $i$ to node $j$ and $w_{ijl}$ represents the network parameters in our model. The difference between the operations is the number of nodes or layers of MLP. Then we denote a one-layer neural network $\mathcal{NN}_l$ with an activation function $\sigma(x)$ by $O_{ijl}$, 
then each operation $ o_{ijl}$ is the composition of $O_{ijl}$, which can be written as:
\begin{equation}
    O_{ijl}(\x_i,\w_{ijl})=\sigma(\w_{ijl}\cdot\x_i).
\end{equation}

Without loss of generality, we take the induced norm as the matrix norm, which is defined by 
 
 \begin{equation}
    \norm{ \w}={\rm sup}\frac{\norm{\w\x}}{\norm{\x}}.
 \end{equation}

We have:

\begin{align}
    &\norm{O_{ijl}(\x_i,\w_{ijl,1})-O_{ijl}(\x_i,\w_{ijl,2})}\\
    &=\norm{\sigma(\w_{ijl,1}\cdot\x_i)-\sigma(\w_{ijl,2}\cdot\x_i)}\\
    &=\norm{\sigma((\w_{ijl,1}-\w_{ijl,2})\cdot\x_i)}\\
    &\leqslant \norm{(\w_{ijl,1}-\w_{ijl,2})\cdot\x_i}\\
    &\leqslant \norm{\w_{ijl,1}-\w_{ijl,2}} \, \norm{\x_i},
\end{align}
where $\sigma$ is the ReLU function.
In mathematical analysis, given that two metric spaces $(X, d_X)$ and $(Y, d_Y)$, where $d_X$ denotes the metric on the set $X$ and $d_Y$ is the metric on set $Y$, a function $f:X\rightarrow Y$ is called \textbf{Lipschitz continuous} if there exists a real constant $K\geqslant0$ such that, for all $x_1, x_2$ in $X$, 

\begin{equation}
    d_Y(f(x_1, x_2)) \leqslant Kd_X(x_1, x_2).
\end{equation}

According to the above definition, the\textbf{ Lipschitz} functions form a vector space/linear space, because $O_{ijl} $ is\textbf{ Lipschitz},
$\x_j(\w,\Alpha)$ is also\textbf{ Lipschitz}, therefore, $f(\x,\w,\Alpha)$ is\textbf{ Lipschitz}  with respect to $\w$. As for $\Alpha$, we have:

\begin{align}
    &\frac{1}{N_d}\norm{f(\x,\w,\Alpha_1)-f(\x,\w,\Alpha_2)}\\
    &=\frac{1}{N_d}\norm{\sum_{i<j}^{j\leqslant N_d}\Alpha_{ij,1}\cdot o_{ij}(\x_i,\w_{ij})-\sum_{i<j}^{j\leqslant N_d}\Alpha_{ij,2}\cdot o_{ij}(\x_i,\w_{ij})}\\
    &=\frac{1}{N_d}\norm{\sum_{i<j}^{j\leqslant N_d}(\Alpha_{ij,1}-\Alpha_{ij,2})\cdot o_{ij}(\x_i,\w_{ij})}\\
    &\leqslant \frac{1}{N_d}\sum_{i<j}^{j\leqslant N_d}\norm{\Alpha_1-\Alpha_2}\norm{o_{ij}(\x_i,\w_{ij})}, \quad (\norm{\Alpha_{ij,1}-\Alpha_{ij,2}}\leqslant  \norm{\Alpha_1-\Alpha_2}, \forall i,j \leqslant N_d),
\end{align}
where $N_d$ is the number of total nodes. Notice that $o_{ij}(\x_i,\w_{ij})$ is a latent variable. thus it's
bounded. Therefore, $f(\x,\w,\Alpha)$ is\textbf{ Lipschitz continuous} with respect to $\Alpha$, and we have:
\begin{align}
    &|f(\x,\w_1,\Alpha)-f(\x,\w_2,\Alpha)|\leqslant C_1\norm{\w_1-\w_2}\\
    &|f(\x,\w,\Alpha_1)-f(\x,\w,\Alpha_2)|\leqslant C_2\norm{\Alpha_1-\Alpha_2}.
    \label{eq7}
\end{align}

In the main paper, our algorithm can be concluded as:
\begin{align}
\w_{n+1}=\w_n-r_1\nabla_{\w}\mathcal{L}(\w_n,\Alpha_n),\qquad &
\Alpha_{n+1}=\Alpha_n-r_2\nabla_{\Alpha}\mathcal{L}(\w_n,\Alpha_n).\qquad\\
\w_{n+1}^{(DMF)}=\w_n-r_1\nabla_{\w}\mathcal{L}(\w_n,\Alpha_n),\qquad
 &\Alpha_{n+1}^{(DMF)}=\Alpha_n-r_2\nabla_{\Alpha}\mathcal{L}(\w_{n+1},\Alpha_n).
\end{align}

Without loss of generality, we only consider a simplified situation which output $\y$ is a scalar, denoted by $y$. This way, we have:

\begin{align}
    &\norm{\Alpha_{n+1}-\Alpha_{n+1}^{(DMF)}}\\
    &=r_2\norm{\nabla_{\Alpha}\mathcal{L}(\w_n,\Alpha_n)-\nabla_{\Alpha}\mathcal{L}(\w_{n+1},\Alpha_n)}\\
    &=r_2\norm{\frac{1}{\mathcal{L}(\w_n,\Alpha_n)}(f(\x,\w_n,\Alpha_n)-y)\nabla_{\Alpha}f(\x,\w_n,\Alpha_n)\\
    &\quad-\frac{1}{\mathcal{L}(\w_{n+1},\Alpha_n)}(f(\x,\w_{n+1},\Alpha_n)-y)\nabla_{\Alpha}f(\x,\w_{n+1},\Alpha_n)}\notag
    \label{eq4}
\end{align}

According to the convergence of $\mathcal{L}(\w_n,\Alpha_n)$, we approximate \ref{eq4} by:

\begin{align}
    &\norm{\Alpha_{n+1}-\Alpha_{n+1}^{(DMF)}}\\
    &\approx \frac{r_2}{\mathcal{L}(\w_n,\Alpha_n)}\norm{f(\x,\w_n,\Alpha_n)-y\nabla_{\Alpha}f(\x,\w_n,\Alpha_n)-f(\x,\w_{n+1},\Alpha_n)+y\nabla_{\Alpha}f(\x,\w_{n+1},\Alpha_n)}\\
    &=\frac{r_2}{\mathcal{L}(\w_n,\Alpha_n)}\norm{f(\x,\w_n,\Alpha_n)\nabla_{\Alpha}f(\x,\w_n,\Alpha_n)-f(\x,\w_{n+1},\Alpha_n)\nabla_{\Alpha}f(\x,\w_{n+1},\Alpha_n)\\
    &\quad -y\nabla_{\Alpha}(f(\x,\w_n,\Alpha_n)-f(\x,\w_{n+1},\Alpha_n))}\notag\\
    &=\frac{r_2}{2\mathcal{L}(\w_n,\Alpha_n)}\norm{{\nabla_{\Alpha}(f^2(\x,\w_{n},\Alpha_n)-f^2(\x,\w_{n+1},\Alpha_n))}\\
    &\quad-2y \nabla_{\Alpha}(f(\x,\w_n,\Alpha_n)-f(\x,\w_{n+1},\Alpha_n))}\notag\\
    &=\frac{r_2}{2\mathcal{L}(\w_n,\Alpha_n)}\norm{\nabla_{\Alpha}(f^2(\x,\w_n,\Alpha_n)-f^2(\x,\w_{n+1},\Alpha_n)\\
    &\quad-2y(f(\x,\w_n,\Alpha_n)-f(\x,\w_{n+1},\Alpha_n)))}\notag\\
    &=\frac{r_2}{2\mathcal{L}(\w_n,\Alpha_n)}\norm{\nabla_{\Alpha}f(\x,\w_n,\Alpha_n)+f(\x,\w_{n+1},\Alpha_n)\\
    &\quad-2y)(f(\x,\w_n,\Alpha_n)-f(\x,\w_{n+1},\Alpha_n))}\notag\\
    &\approx \frac{r_2}{2\mathcal{L}(\w_n,\Alpha_n)}\norm{\nabla_{\Alpha}[r_1(f(\x,\w_n,\Alpha_n)+f(\x,\w_{n+1},\Alpha_n)\\
    &\quad -2y)(\nabla^T_{\w}f(\x,\w_n,\Alpha_n)\nabla_{\w}\mathcal{L}(\w_n,\Alpha_n))]}\notag\\
    & = \frac{r_1r_2}{2\mathcal{L}(\w_n,\Alpha_n)}\norm{\nabla_{\Alpha}(f(\x,\w_n,\Alpha_n)+f(\x,\w_{n+1},\Alpha_n)\\
    &\quad -2y)\nabla^T_{\w}f(\x,\w_n,\Alpha_n)\nabla_{\w}\mathcal{L}(\w_n,\Alpha_n)+(f(\x,\w_n,\Alpha_n)\notag\\
    &\quad+f(\x,\w_{n+1},\Alpha_n)-2y)\nabla_{\Alpha}(\nabla^T_{\w}f(\x,\w_n,\Alpha_n)\nabla_{\w}\mathcal{L}(\w_n,\Alpha_n))}\notag
\end{align}

Again, due to the convergence of $\mathcal{L}(\w_n,\Alpha_n)$, the $f(\x,\w_n,\Alpha_n)-y$ is also convergent on $\mathbb{R}$, we denote the limit by $c$, let $n\rightarrow\infty $, $c\rightarrow 0$ then, we can further simplify Equation $\ref{eq4}$ into:

\begin{align}
    &\norm{\Alpha_{n+1}-\Alpha_{n+1}^{(DMF)}}\\
    &= \frac{r_1r_2}{2\mathcal{L}(\w_n,\Alpha_n)}\norm{\nabla_{\Alpha}(f(\x,\w_n,\Alpha_n)+f(\x,\w_{n+1},\Alpha_n)\\
    &\quad -2y)\nabla^T_{\w}f(\x,\w_n,\Alpha_n)\nabla_{\w}\mathcal{L}(\w_n,\Alpha_n)\notag}\\
    &\leqslant \frac{r_1r_2}{2\mathcal{L}(\w_n,\Alpha_n)}\norm{\nabla_{\Alpha}(f(\x,\w_n,\Alpha_n)+f(\x,\w_{n+1},\Alpha_n)\\
    &\quad -2y)}\norm{\nabla_{\w}f(\x,\w_n,\Alpha_n)}\norm{\nabla_{\w}\mathcal{L}(\w_n,\Alpha_n)\notag}\\
    &=\frac{r_1r_2}{\mathcal{L}(\w_n,\Alpha_n)}\norm{\nabla_{\Alpha}f(\x,\w_n,\Alpha_n)}\norm{\nabla_{\w}f(\x,\w_n,\Alpha_n)}\norm{\nabla_{\w}\mathcal{L}(\w_n,\Alpha_n)}\\
\end{align}

According to \ref{eq7}, $f(\x,\w_n,\Alpha_n)$ is\textbf{ Lipschitz}, and $f(\x,\w_n,\Alpha_n)$ have bounded first derivatives, which shows that $\nabla_{\w}f(\x,\w_n,\Alpha_n)$ and $\nabla_{\Alpha}f(\x,\w_n,\Alpha_n)$ have upper bounds. Therefore, we have:

\begin{equation}
    \norm{\Alpha_{n+1}-\Alpha_{n+1}^{(DMF)}}\leqslant M\frac{r_1r_2}{\mathcal{L}(\w_n,\Alpha_n)}.
\end{equation}

When $r_1$ and $r_2$ are small enough, and we have $ \w_{n+1} = \w_{n+1}^{(DMF)} $. Therefore, the two training methods are equivalent.

\end{document}